\documentclass[twoside,11pt]{article}
\usepackage{jair, theapa, rawfonts}
\usepackage{graphicx}
\usepackage{mathtools}
\usepackage{multirow}
\usepackage{amssymb}
\usepackage{todonotes}
\usepackage{footmisc}
\usepackage{pifont}
\usepackage{float}
\usepackage{fixltx2e}
\usepackage{subcaption}
\usepackage{caption}
\usepackage{color, colortbl}
\usepackage{xcolor}
\usepackage{booktabs}
\usepackage{subfiles}
\usepackage{url}
\usepackage{pdfpages}  % include external PDF

\usepackage{wrapfig}

\usepackage{rotating}
\usepackage[T1]{fontenc}
\newcommand{\cmark}{\ding{51}}%
\newcommand{\xmark}{\ding{55}}%
\jairheading{71}{2021}{1183 - 1317}{09/2019}{08/2021} % title header, JAIR template
\usepackage[toc,page]{appendix}
\usepackage{nomencl}
\makenomenclature
\usepackage{etoolbox}
\renewcommand\nomgroup[1]{%
  \item[\Large\bfseries
  \ifstrequal{#1}{N}{Nomenclature}{%
  \ifstrequal{#1}{A}{Appendix}{}}%
]\vspace{10pt}} % this is to add vertical space between the groups.
\ShortHeadings{Trends in Integration of Vision and Language Research}
{Mogadala, Kalimuthu, \& Klakow}

\begin{document}

\title{Trends in Integration of Vision and Language Research: \\ A Survey of Tasks, Datasets, and Methods}

\author{\name Aditya Mogadala \email amogadala@lsv.uni-saarland.de \\
       \name Marimuthu Kalimuthu \email mkalimuthu@lsv.uni-saarland.de \\
       \name Dietrich Klakow \email dietrich.klakow@lsv.uni-saarland.de \\
       \addr Spoken Language Systems (LSV) \\ Saarland Informatics Campus\\ 
       Saarland University\\ 
       66123 Saarbr{\"u}cken, Germany}

\maketitle

\begin{abstract}
Interest in Artificial Intelligence (AI) and its applications has seen unprecedented growth in the last few years. This success can be partly attributed to the advancements made in the sub-fields of AI such as machine learning, computer vision, and natural language processing. Much of the growth in these fields has been made possible with deep learning, a sub-area of machine learning that uses artificial neural networks. This has created significant interest in the integration of vision and language. In this survey, we focus on ten prominent tasks that integrate language and vision by discussing their problem formulation, methods, existing datasets, evaluation measures, and compare the results obtained with corresponding state-of-the-art methods. Our efforts go beyond earlier surveys which are either task-specific or concentrate only on one type of visual content, i.e., image or video. Furthermore, we also provide some potential future directions in this field of research with an anticipation that this survey stimulates innovative thoughts and ideas to address the existing challenges and build new applications. 
\end{abstract}

\section{Introduction}
\label{sec:intro}

Recent wave of unprecedented progress in deep learning methods has advanced the fields of Computer Vision (CV) and Natural Language Processing (NLP) to an extent that they are now making significant progress across several challenging tasks. Independent of NLP, computer vision has achieved prominent improvements in tasks such as visual content classification~\shortcite{he:2016}, object detection~\shortcite{redmon:2017}, semantic segmentation~\shortcite{he:2017}, etc., using large annotated datasets or by employing self-supervision~\shortcite{jing:2019} on large-scale unlabeled data. Similarly, independent from computer vision, NLP has seen a surge of interest in solving multiple tasks at once with unsupervised pretraining of language models~\shortcite{devlin:2018,radford:2019,lample:2019,gpt3-brown:2020} using large unlabeled corpora. However, there is a growing interest in solving challenges that combine linguistic and visual information from these traditionally independent fields. The methods which address the challenge of integration should provide a complete understanding of visual and/or textual content, and are expected to (1) generate comprehensible but concise and grammatically well-formed descriptions of the visual content, or vice versa by generating the visual content for a given textual description in a natural language of choice, (2) identify objects in the visual content and infer their relationships to reason about, or answer arbitrary questions about them, (3) navigate through an environment by leveraging input from both vision and natural language instructions, (4) translate textual content from one language to another while leveraging the visual content for sense disambiguation, (5) generate stories about the visual content, and so on. Designing methods which can process and relate information from multiple modalities (i.e., linguistic and visual information) is usually considered to be a sub-part of multimodal learning models~\shortcite{mogadala:2015}. 

Efficiently solving the above mentioned and other related challenges can result in many potential real-world applications. For example, visually impaired individuals can be assisted by visual scene understanding, where they can get information about a scene from generated descriptions and by being able to ask questions about it. Other applications include automatic surveillance~\shortcite{baumann:2008}, autonomous driving~\shortcite{kim:2018}, human-computer interaction~\shortcite{rickert:2007}, city navigation~\shortcite{dewalk:2018}, and so on. Also, solving such challenges can serve as an excellent test bed for computer vision and NLP systems, one that is much more intelligent and comprehensive than independent computer vision and NLP evaluations.

Given such a broad scope for fundamental and applied research, there has been several surveys in recent years aiming to provide a comprehensive overview of the integration of vision and language tasks. These surveys, however, have restricted themselves on covering specific vision and language integration tasks such as image description~\shortcite{ber:2016,bai:2018,hossain:2019} or video description generation~\shortcite{aafaq:2018}, visual question answering~\shortcite{kafle:2017,wuvqa:2017}, action recognition~\shortcite{gella:2017} and visual semantics~\shortcite{liusi:2019}. The surveys which went beyond these specific tasks have summarized dataset statistics~\shortcite{ferraro:2015}, provided a comprehensive overview of only NLP tasks such as natural language generation (NLG)~\shortcite{gatt:2018,nlg-survey-garbacea:2020} and commonsense reasoning~\shortcite{storks:2019}. However, there was also an attempt to cover multiple modalities (including sound)~\shortcite{baltruvsaitis:2019}, but it was structured in a bottom-up manner giving more importance to underlying fusion technologies than the task itself. Also, there was some interest in understanding the limitations of the integration of vision and language research~\shortcite{kafle:2019}. However, it is limited to the task of language-grounded image understanding. Furthermore, there were ideas to develop theories on the complementarity of language and visual data in the human-machine communication from a theoretical point of view~\shortcite{moens:2019}.

With our efforts in this survey, we go beyond these and present a comprehensive overview of ten different tasks that are prominent in the current integration of vision and language research. We first begin with a background on the traditional tasks in computer vision and NLP separately, and then show how they facilitate in designing the prominent ten tasks for the integration of vision and language modalities in Section~\ref{sec:background}. Following that, we provide an in-depth exploration of each of the ten tasks and present details about the datasets, methods, results, and open challenges in separate sections beginning from Section~\ref{sec:vdgandstory} and ending at Section~\ref{sec:vlntask}. In Section~\ref{sec:vlpretrain}, we provide details about the joint pretraining of vision and language, which is gaining momentum in recent years, that aims to solve multiple tasks at once using learned representations. It is then followed in Section~\ref{sec:future} by potential future research directions. Finally, in Section~\ref{sec:conc}, we conclude our survey and offer some insights.

\section{Background}
\label{sec:background}

In this section, we first briefly introduce some of the standard tasks that are studied in computer vision and NLP separately. We then present how the tasks are modified such that they facilitate in designing ten prominent tasks for the integration of vision and language.

\subsection{Computer Vision (CV) Tasks}
\label{ssec:visiontask}
An array of different tasks are studied in computer vision. Keeping in mind the underlying goal of computer vision is to describe and explain visual information, we divide these tasks based on where the visual data arises. In this survey, we mainly focus on images and videos as the visual information, although RGB-D and point cloud data are becoming prevalent.

\subsubsection{Image as Visual Information }
\label{sssec:visiontaskimage}
We describe two different aspects of the use of images in computer vision: (1) the tasks where images are used as input, and (2) the representation of images. In the following, we discuss various computer vision tasks that use images as input and present the recent progress and improvements made for representing image data.

\paragraph{Tasks.} The following tasks use images as input: (1) Image Classification (2) Object Localization (3) Object Detection (4) Object Segmentation (5) Object Identification (6) Instance Segmentation and (7) Panoptic Segmentation. 

The fundamental difference between aforementioned tasks is that majority of them focus on carving out the exact position of visual object in an image, while rest of them provide a predefined class label for an image. There are also advanced tasks that use images as visual information and assist in the integration of computer vision and NLP. These tasks include (1) Image Style Transfer (2) Image Colorization and (3) Image Synthesis.

\paragraph{Representation.}The advent of deep learning~\shortcite{lecun:2015,dl4ai-bengio:2021} has tremendously changed the field of computer vision. Convolutional Neural Networks (CNNs)~\shortcite{lecun:1995} have become the de facto standard for generating representations of images using end-to-end trainable models.

There are several variations of CNNs that learn image features with supervised or self-supervised techniques~\shortcite{jing:2019}. Most of these techniques are designed to learn transferable general image features by leveraging tasks presented earlier.

Commonly, transferable global image representations are learned with deep CNN architectures such as AlexNet~\shortcite{krizhevsky:2012}, VGGNet~\shortcite{simonyan:2014}, GoogLeNet~\shortcite{szegedy:2015}, Inception-v3~\shortcite{szegedy:2016}, Residual Networks (ResNets)~\shortcite{he:2016}, DenseNets~\shortcite{huang:2017}, and Efficient Net~\cite{tanefficientnet:2019} using large datasets, viz. ImageNet\footnote{\url{https://www.image-net.org}\label{fnote: imagenet-dataset-url}} \shortcite{deng:2009}, MSCOCO\footnote{\url{http://cocodataset.org/#home} \label{fnote: mscoco-dataset-url}} \shortcite{lin:2014}, and Visual Genome\footnote{\url{https://visualgenome.org}\label{fnote: vg-dataset-url}} \shortcite{krishnagenome:2017}. However, for some vision and language integration tasks, it is preferred to learn global image features during task-specific training as opposed to independently learning generic, pretrained representations.

For learning local features of objects that are typically represented with bounding boxes in images, the preferred choice is to utilize some region specific CNN architectures such as Region-based CNNs (R-CNN)~\shortcite{renrcnn:2015}. More recently, there is an interest in using self-attention based approaches, namely Transformers~\shortcite{vaswani:2017} for achieving end-to-end object detection~\shortcite{carionend:2020}.

\subsubsection{Video as Visual Information}
\label{sssec:visiontaskvideo}
Similar to images, when a video is used as visual data, we need to consider two crucial aspects: (1) knowing the tasks where videos are used as inputs, and (2) the representation of a video. In the following, we discuss different tasks in computer vision that use video as input and further present the recent progress made in video representations.

\paragraph{Tasks.}  Recently, the tasks on videos are also gaining importance, such as (1) Object Tracking (2) Action Classification (3) Emotion Detection (4) Scene Detection and (5) Automated Editing. The core difference between earlier tasks is that majority of them focus on tracking a visual object present in a scene of a video, while rest of them identify the task happening in a video such a action etc. 

\paragraph{Representation.} To account for the temporal nature of videos, RGB images are stacked as frames to form a 4D representation (i.e., video). Usually, visual data observed in videos is extracted in the form of screenshots that are amenable to the same techniques for image local and global representation. However, in addition, spatio-temporal features are also developed with general video analysis such as C3D~\shortcite{tran:2014}, or from action recognition datasets i.e., Kinetics action recognition~\shortcite{kay:2017} to build RGB-D or Inflated 3D ConvNet (I3D) features~\shortcite{carreira:2017} using different CNN architectures.

\subsection{NLP Tasks}
\label{ssec:languagerep}
Like in Section~\ref{ssec:visiontask}, the fundamental goals of most NLP tasks are to comprehend or generate language. In this section, we describe a few of the popular tasks that drive NLP research. We also discuss current approaches used to represent language.

\paragraph{Tasks.} The aim of NLP tasks is two-fold: i) understanding language, ii) generating language. Some of the classical NLP tasks, that are used to comprehend language, are shallow parsing, syntax parsing, semantic role labeling, named entity recognition, entity linking, co-reference resolution, etc. Tasks to generate language in a conditional or unconditional manner are machine translation, text summarization, etc.

\paragraph{Representation.} In deep learning based approaches, language is usually represented either as a bag-of-words or as distributed representations. For words in a sentence, initializations are commonly done with pretrained word embeddings~\shortcite{mikolov:2013,pennington:2014}. Additionally, to represent variable-length text inputs, sequence learning techniques such as recurrent neural network variations like unidirectional Long Short-Term Memory (LSTM)~\shortcite{hochreiter:1997}, or bidirectional LSTM (BiLSTM) and unidirectional Gated Recurrent Units (GRUs)~\shortcite{chung:2014}, or bidirectional GRUs (BiGRUs) are applied. Recently, to provide parallelization in sequential training, self-attention based approaches, viz. Transformers~\shortcite{vaswani:2017}, have been employed to build architectures such as BERT~\cite{devlin:2018} and its variations.

\subsection{CV and NLP Integration Tasks}
\label{ssec:backintvisionandlang}
Over the past few years, significant progress has been made in the research concerning the integration of language and vision. Several tasks exist which combine language, observed at different levels (i.e., words, phrases, sentences, paragraphs, and documents), with visual information, typically represented as images or videos. Initially, most works concentrated on combining low-level linguistic units, such as words with images or videos for building visual-semantic embeddings~\shortcite{visual-semantic-barnard:2003,frome:2013,kirosuni:2014,liuvid:2015,cao:2016,tsai:2017,guo:2018,mogadaladiscovering:2018,wangitm:2019,kimmule:2020}, which are beneficial for downstream applications, as well as understanding adversarial attacks~\shortcite{wuunified:2019} to improve model robustness.

\begin{figure}[!htb]
    \centering
        \includegraphics[width=\textwidth]{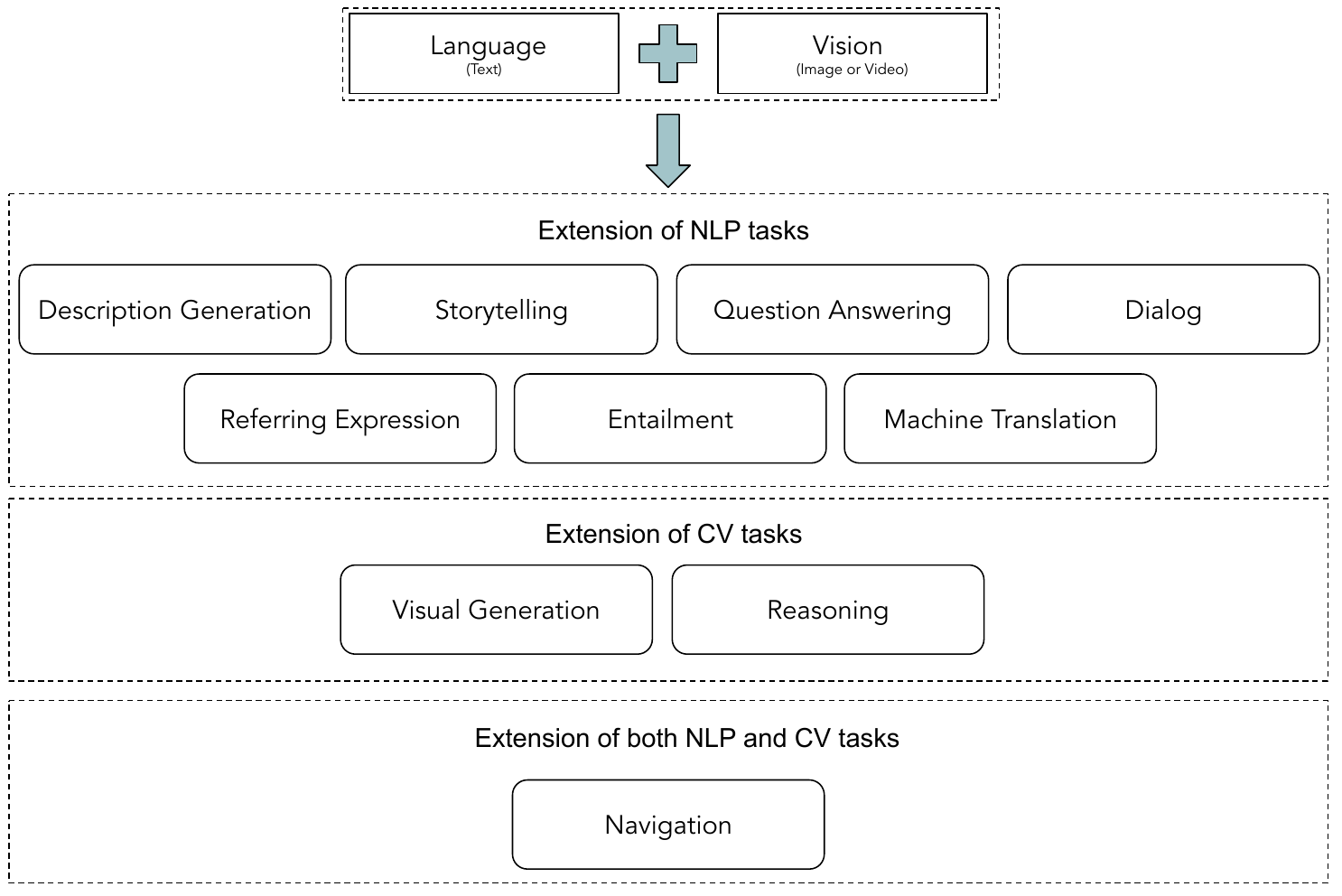}
    \caption{Ten different Language and Vision integration tasks.}\label{fig:tasks}
\end{figure}

However, it will be appealing to look into those tasks that go beyond words and consider variable-length texts larger than words as language input. Most of these tasks can be seen as an extension to either CV, NLP, or both. Figure~\ref{fig:tasks} provides an illustration of different tasks and their groupings.

To get a grasp on how these tasks are perceived as a natural extension of tasks in computer vision, NLP, or both, we briefly describe their relation with similar tasks addressed in their research.

\paragraph{Extension of NLP Tasks}

\begin{itemize}
    \item \textbf{Visual Description Generation} is closely related to conditional language modeling~\shortcite{de:2015} or the Natural Language Generation (NLG)~\shortcite{reiter:2000} tasks in NLP. Given non-linguistic information (e.g., image or video), the goal is to generate a human-readable text snippet that describes the input.
    \item The task of \textbf{Visual Storytelling} solves a similar problem to visual description generation. However, instead of dealing with a single visual input, a sequence of visual inputs is used to generate a narrative summary based on the text aligned with them. It can be seen that the task is closely aligned to text summarization~\shortcite{nallapati:2016,liu:2018}, mostly generating abstractive summaries.
    \item \textbf{Visual Question Answering} draws its inspiration from text-based question answering~\shortcite{harabagiu:2000,strzalkowski:2006}, which is one of the long-standing topics in NLP research. Here, answering questions about a visual input is perceived as its natural extension.
    \item The task of \textbf{Visual Dialog} aims at creating a meaningful dialog in a natural and conversational language about a visual content. It is perceived as the visual analogue of the text-based dialog and conversation system~\shortcite{weizenbaum:1966,dodge:2015,li:2016} that has been explored in NLP over many years.
    \item \textbf{Visual Referring Expression} is an extension of referring expression~\shortcite{krahmer:2012} in natural language generation systems. Also, the sub-problem in visual referring expression (i.e., comprehension) is perceived as an analogy of pragmatics in linguistics~\shortcite{thomas:2014} due to its usage of context.
    \item \textbf{Visual Entailment} is an inference task for predicting whether the image semantically entails the text. The task has been proposed as a natural extension to natural language inference~\shortcite{condoravdi:2003,bowman:2015}, where the premise is text, instead of a visual content.
    \item The goal in \textbf{Multimodal Machine Translation} is to achieve translation from source language(s) to target language(s) by leveraging the visual information as auxiliary modality along with the natural language text in source language(s). It is influenced by the well-known NLP task of machine translation that aims to automatically translate textual contents between any two natural languages~\shortcite{brown:1990,bahdanau:2014}.
\end{itemize}

\paragraph{Extension of CV Tasks}
\begin{itemize}
    \item \textbf{Visual Generation} deals with the generation of visual content by conditioning on input text from a chosen natural language. It can be perceived as a multimodal extension of the popular computer vision tasks of image-to-image translation~\shortcite{isola:2017} and neural style transfer~\shortcite{gatys:2016}.
    
    \item The task of \textbf{Visual Reasoning} is a direct extension of visual perception where standard computer vision tasks such as image classification~\shortcite{krizhevsky:2012}, object detection~\shortcite{reni:2015}, or semantic segmentation~\shortcite{long:2015} are performed. Instead of providing only class labels (in case of classification), bounding boxes (in case of detection), or segments (in case of segmentation), visual reasoning is expected to output a relationship between detected objects by generating an entire visual scene graph. Furthermore, the scene graph is leveraged to reason and answer questions about visual content. It can also be used to reason about whether a natural language statement is true or not regarding a visual input~\shortcite{suhr-nlvr:2017}. 
\end{itemize}

\paragraph{Extension of both NLP and CV Tasks}
\begin{itemize}
\item \textbf{Vision-and-Language Navigation} is one task that can be seen as a transition from standard vision-based navigation using only visual input~\shortcite{sinopoli:2001,blosch:2010} or natural language instruction based navigation~\shortcite{macmahon:2006,vogel:2010}. The expectation here is that the natural language navigation instruction should be interpreted based on visual input. Hence, it combines both vision and language.
\end{itemize}

\paragraph{Representation.}
In earlier sections, we discussed different architectures used to represent both vision and language separately. Combining representations of language and vision is essential to address vision and language integration tasks in an effective manner. There are various models that have been proposed for each task to build representations that integrate vision and language. We discuss these in greater detail in forthcoming sections where each of the tasks are introduced.

\subsection{Summary}
\label{ssec:summrep}
In background section, we have reviewed a variety of tasks that integrate vision and NLP. Additionally, we explored diverse methods that are used for the \textit{representation} of vision and language modalities. Furthermore, we understood the training procedure used by different methods that use supervised learning. For example, models built using those methods leverage first-order optimization algorithms such as Stochastic Gradient Descent (SGD)~\shortcite{bottou:2010}, ADAM~\shortcite{kingma:2014} or RMSProp~\shortcite{tieleman:2012}. While, some methods also utilize Reinforcement Learning (RL)~\shortcite{suttonintroduction:1998} in contrast with only supervised learning.

We will see that many of the models developed for these tasks use similar architectures for the representation of vision and language modalities and depend on standard gradient-based optimization algorithms for training. This shows that, although the aims of each task are different, the underlying principles to extract meaning from unstructured data remain the same.

%\end{document}

\section{Visual Description Generation and Storytelling}
\label{sec:vdgandstory}
In this section, we explore two different tasks, \textit{Visual Description Generation} and \textit{Visual Storytelling}. Although the goals of these tasks do not perfectly line up, they share the common intention of generating a textual description when conditioned on visual input. In the following, we present more details about each of these tasks separately. 
\subsection{Visual Description Generation}
\label{ssec:vc}
The aim in description generation is to generate either a global description or dense captions for a given visual input. Depending on the type of visual input, i.e., either an image or a video, there are various ways to explore the problem.

\subsubsection{Image Description Generation - Introduction}
\label{sssec:imagecaptiongenintro}
There are many subareas of image description generation where the underlying goal of generating global or dense descriptions remains the same, but the way those descriptions appear is different. In the following section, we explore some of the popular categories observed in image description generation.
\paragraph{Standard Image Description Generation.}

\begin{figure}[!htb]
    \centering
        \includegraphics[width=\textwidth]{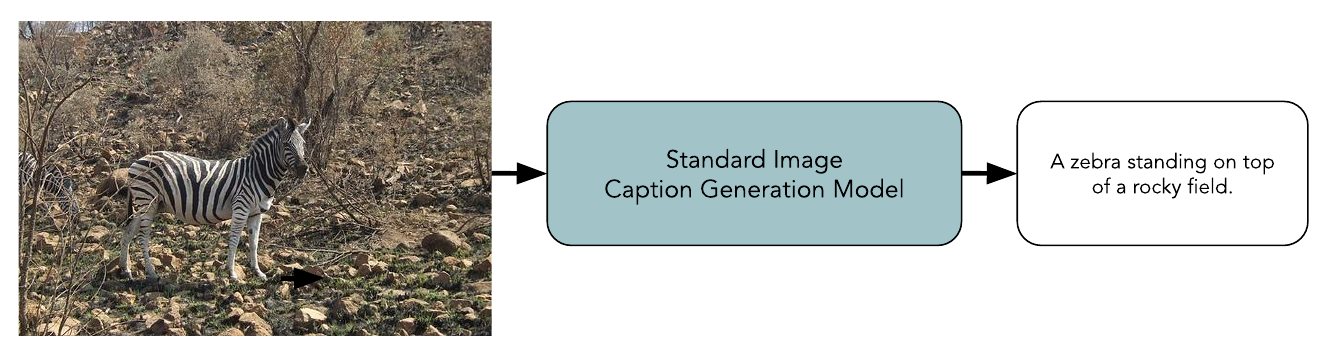}
    \caption{Given an \textit{image}, the Standard Image Caption Generation Model generates a single global textual description of the scene.}\label{fig:standardic}
\end{figure}

The goal in standard image description generation is to generate a sentence-level description of the scene in a given image. Here, methods leverage only vocabulary of the dataset to generate the best description that depicts the scene in the image. Figure~\ref{fig:standardic} provides a conceptual representation of the task.

Initially, several methods were developed based on templates, n-grams, and dependency parsing~\shortcite{farhadi:2010,yangic:2011,licomp:2011,mitchellmidge:2012,kulkarni:2013,elliott:2013,fang:2015}. Recently, however, image description generation models based on the encoder-decoder a.k.a. Sequence-to-Sequence (Seq2Seq) frameworks~\shortcite{cho:2014,vinyals:2015} have become popular. Moreover, the above said frameworks have been extended with \textit{attention} mechanisms~\shortcite{bahdanau:2014} to support the selection of local image features that are useful for the generation of words at each time step~\shortcite{xu:2015}.  Table~\ref{caption-summs} summarizes different setups for generating image descriptions using neural network based non-attention, attention, and reinforcement learning techniques. Other variations include cross-lingual image captioning~\shortcite{miyazaki:2016} and multi-language image description generation~\shortcite{elliott:2015}. 

In the following, we explore some of the related ideas that expand the scope of image description generation.
\begin{table*}[!ht]
  \small
\begin{center}
\begin{tabular}{lcc}
\hline      %\toprule
\rowcolor{teal!35}
%\multicolumn{4}{c}{Image description Generation Setups} \\
%\hline
Approach & Attention & RL \\ 
\hline \addlinespace[0.3em]     %\midrule
MLBL~\shortcite{kiros:2014} & \xmark & \xmark \\
m-RNN~\shortcite{mao:2014}  & \xmark & \xmark \\
Minds Eye~\shortcite{xchen:2015}& \xmark & \xmark \\
BRNN~\shortcite{karpathy:2015}& \xmark & \xmark \\
NIC~\shortcite{vinyals:2015} & \xmark & \xmark \\
LRCN~\shortcite{donahue:2015} & \xmark & \xmark \\
Guided LSTM~\shortcite{jia:2015} & \xmark & \xmark \\
Deep Bidirectional LSTM~\shortcite{wang:2016}& \xmark & \xmark \\
Regional Visual Attributes~\shortcite{wu:2017} & \xmark & \xmark \\
Language CNN~\shortcite{gu:2017} & \xmark & \xmark \\
ConceptNet-NIC~\shortcite{zhou:2019} & \xmark & \xmark \\
\cmidrule(lr){1-3}
Visual Attention~\shortcite{xu:2015} & \cmark & \xmark \\
Region-based Attention~\shortcite{jin:2015} & \cmark & \xmark \\
Attribute Attention~\shortcite{you:2016}& \cmark & \xmark \\
Review Attention~\shortcite{yang:2016} & \cmark & \xmark \\
Adaptive Attention~\shortcite{lu:2016} & \cmark & \xmark \\
Areas of Attention~\shortcite{pedersoli:2016} & \cmark & \xmark \\
Contrastive Adaptive Attention~\shortcite{dai:2017} & \cmark & \xmark \\
Neural Baby Talk w/ Attention~\shortcite{lu:2018} & \cmark & \xmark \\
Convolutional Attention~\shortcite{aneja:2018}& \cmark & \xmark \\
Reflective Decoding Network~\shortcite{rdn-caption-ke:2019}& \cmark & \xmark \\
\cmidrule(lr){1-3}
Self-Critical Attention~\shortcite{rennie:2016} & \cmark & \cmark \\
Policy Gradient~\shortcite{liu:2017} & \cmark & \cmark \\
Up-Down~\shortcite{anderson:2017} & \cmark & \cmark \\
Multi-task Captioning~\shortcite{zhao:2018} & \cmark & \cmark \\
Stack Captioning~\shortcite{gu:2018} & \cmark & \cmark \\
Attention on Attention~\shortcite{aoa-caption-huang:2019} & \cmark & \cmark \\
Meshed-Memory Transformer~\shortcite{m2-transformer-cornia:2020}& \cmark & \cmark \\
\bottomrule
\end{tabular}
\end{center}
\caption{\label{caption-summs}Summary of methods for generating a global description of an image. Approaches are segregated based on their usage of no-attention, attention, and RL techniques.}
\end{table*}

\paragraph{Dense Image Description Generation.} Dense image description generation task aims to create descriptions at the local object-level in a given image. It is referred to as dense captions since the commonly used image datasets have images containing multiple objects.
Several approaches~\shortcite{plummer:2015,johnson:2016,rohrbach:2016,hu:2017} have been proposed to generate dense captions in images. Usually, they use representations of phrases and their relationships to generate descriptions~\shortcite{kim:2019}.

\paragraph{Image Paragraph Generation.} The aim in image paragraph generation is to create paragraphs instead of generating a single simple description, or dense descriptions for an image. Generated paragraphs are expected to be coherent and contain fine-grained natural language descriptions~\shortcite{krause:2017,liang:2017,chatterjee:2018}.

\paragraph{Spoken Language Image Description Generation.} Spoken language image description generation expands the description generation task to work with spoken language, instead of limiting to only the written forms of language. Investigations such as visually grounded speech signals~\shortcite{chrupala:2017} address the standard image description generation task from the perspective of a spoken language.

\paragraph{Stylistic Image Description Generation.} Stylistic image description generation adds styles to the standard image description generation, where the generated descriptions adhere to a specific style. For example,~\shortciteA{mathews:2016} generated captions which capture the \textit{sentiments} of an image, while~\shortciteA{gan:2017} attempted at generating humorous and romantic captions. In addition, this task has been extended by leveraging unpaired textual corpora~\shortcite{mathews:2018} to generate story-like captions. Furthermore, to make the generated captions more human-like, personality traits have been used to generate captions~\shortcite{shuster:2019}. Recently, multi-style image description generation~\shortcite{guomscap:2019} has been explored, in which a single model using unpaired data is built to generate different stylized captions.

\paragraph{Unseen Objects Image Description Generation.} Unseen objects image description generation leverages images which lack paired descriptions. Most of the paired image-description datasets have few visual objects to represent. Hence, methods such as Deep Compositional Captioning (DCC)~\shortcite{anne:2016}, Novel Object Captioner (NOC)~\shortcite{venu:2017}, 
Constrained Beam Search (CBS)~\shortcite{andersonb:2017}, and LSTM-C~\shortcite{yao:2017} address the challenge of generating descriptions for these images. They generate descriptions for visual object categories that are previously unseen in image-description corpora, either by transferring information between seen and unseen objects before inference (i.e., before test time), or by keeping constraints on the generation of description words during inference (i.e., during test time). A few approaches~\shortcite{mogadala:2018,lu:2018} have transferred information both before and during inference. Recently, pointing LSTM was designed to point to the novel objects~\shortcite{lipoint:2019} by balancing generation and copying of words. Nevertheless, earlier approaches work only with a limited set of objects. To address this issue, a large-scale nocaps dataset~\shortcite{agrawal:2018} was created.

\paragraph{Diverse Image Description Generation.} Diverse image description generation task aims to incorporate variety and diversity in the generated captions. A few approaches~\shortcite{dai:2017b,shetty:2017} have leveraged 
adversarial training, while~\shortciteA{vijayakumar:2016} used diverse beam search to decode diverse image captions in English. Approaches have also been proposed to describe cross-domain images~\shortcite{chen:2017}.

\paragraph{Controllable Image Description Generation.} Controllable image description generation task focuses on selecting specific objects in an image, defined by a control signal, to generate descriptions. Initially, \shortciteA{yinobj:2017} generated layouts from images, while~\shortciteA{wangobj:2018} counted image objects to produce multiple captions for a given image. Additionally, a control signal has been used to make the image captioning process more controllable, and also to generate diverse captions.~\shortciteA{cornia:2018} used either a sequence or a set of image regions. Also, chunks of the generated sentences were explicitly grounded on regions. Moreover, instead of making captions only diverse, there were also attempts to make the generated descriptions more accurate~\shortcite{deshpande:2018}.

\paragraph{Image Caption Emendation as Generation.}
Caption emendation task is a variant of caption generation where the aim is to build a model to emend (a.k.a. edit or correct) both syntactic and semantic errors in the captions. There has been a lot of interest in recent years on this emerging topic of research.~\shortciteA{show-tell-polish-guo:2020} proposed \textit{Show, Tell, and Polish} framework to better mimic humans in sentence constructions. That is, coming up with a first version and then keep \textit{polishing} it until it feels right. The core idea of this architecture is to perform a two-pass decoding, instead of the typical single-pass decoding. Thus, the model contains two decoder modules, viz., \textit{base decoder} and \textit{ruminant decoder}, whereby the base decoder generates a first version of caption which then feeds into the ruminant decoder for refinement (a.k.a. polishing). Along the same lines,~\shortciteA{capemend-kalimuthu:2020} introduced \textit{fusion models for caption emendation}, which is a generic fusion model framework containing a standard encoder-decoder format image captioning model, a pretrained auxiliary language model (AuxLM - BERT MLM), and a fusion module component that fuses language-only representations of AuxLM and visual-linguistic representations of decoder using different fusion techniques. The intuition behind introducing an external language model trained on a large-scale language corpora is to capture world knowledge and rich linguistic features, which are both scarce in annotated captions data, in an attempt to generate fluent and accurate descriptions. In both of the above approaches, emendation is achieved by generating a caption while utilizing the baseline caption as a reference. That is, the model is trained to correct any errors and incongruencies in the baseline caption. Likewise, \shortciteA{edit-tell-sammani:2020} propose \textit{Show, Edit, and Tell} framework as an iterative adaptive refinement approach that utilizes attention LSTMs and denoising autoencoders for correcting captions.

\subsubsection{Image Description Generation - Datasets}
\label{sssec:icdatasets}

A wide range of datasets are available for conducting research in integration of vision and language. In fact, they are one of the main driving forces behind recent accelerated advancements that we are witnessing in this field~\shortcite{dl4ai-bengio:2021}. Visual information associated with textual content in these datasets differ from each other in many aspects such as size, quality, and the way in which they are collected. In our survey, we summarize the characteristics of these datasets and provide basic statistics about them. However, we do not furnish a deeper analysis of them, as this was already done by~\shortciteA{ferraro:2015}.

An array of diverse datasets, both of small and large-scale, were created and made available publicly in the past decade to address the challenge of image description generation.  Some of the early large-scale datasets focus on image captions, while the others are only of small- or medium-scale. In the following sections, we cover only those datasets that are extensively used in the image captioning literature.

\paragraph{SBU Captioned Photo Dataset (SBU1M).}\label{para:sbu1m-data}SBU1M\footnote{\url{http://vision.cs.stonybrook.edu/~vicente/sbucaptions}\label{fnote: sbu-1m-url}} \shortcite{vincente-sbu-1m:2011} is an automatically collected image description dataset that uses query terms to retrieve images and associated text from Flickr\footnote{\url{https://www.flickr.com}\label{fnote: flickr-url}}. This web-scale dataset is distributed as a single plain text file containing 1 million URLs of Flickr images and their corresponding captions. Although one of the older datasets in image description research, it has been rarely used in recent years. Table~\ref{table:sbu1m-dataset} provides basic statistics about this dataset.

\begin{table}[!ht]
\small
    \centering
    \begin{tabular}{l c c c}
    \hline %\toprule
        \rowcolor{teal!35}
         Total Images          & Captions per Image      & Total Captions     & Object Categories\\
        \hline \addlinespace[0.3em] % \midrule
         1,000,000      & 1             & 1,000,000 & 89  \\
    \bottomrule
    \end{tabular}
    \caption{\label{table:sbu1m-dataset} Basic statistics of the SBU1M image description dataset.}
\end{table}

\paragraph{Flickr8k.}\label{para:flickr8k-data} As with SBU1M, images in the Flickr8k\footnote{\url{http://hockenmaier.cs.illinois.edu/8k-pictures.html}\label{fnote: flickr8k-dataset-url}} \shortcite{hodosh:2013} dataset are also retrieved from Flickr$^{\ref{fnote: flickr-url}}$. However, unlike the automated way of collection of SBU1M, the images in Flickr8k are selected through user queries for specific objects and actions using the Amazon Mechanical Turk (AMT) platform. The images are then captioned by annotators on AMT such that each image contains five captions that are independently created.  Table~\ref{table:flickr8k-dataset} presents the so-called \textit{karpathy split}\footnote{\url{https://cs.stanford.edu/people/karpathy/deepimagesent} \label{fnote: karpathy-split-url}} of the dataset.

\begin{table}[!ht]
\small
    \centering
    \begin{tabular}{l | c c c}
    \hline %\toprule
        \rowcolor{teal!35}
        Split               & Images   & Captions per Image    & Total Captions\\
        \hline \addlinespace[0.3em] % \midrule
        Training            &  6,000    & 5                     & 30,000  \\
        Validation          &  1,000    & 5                     & 5,000  \\
        Test                &  1,000    & 5                     & 5,000 \\ 
        \midrule
        Total               & 8,000     & 5                     & 40,000 \\ 
    \bottomrule
    \end{tabular}
    \caption{\label{table:flickr8k-dataset} Splits of the Flickr8k image description dataset.}
\end{table}

\paragraph{Flickr30k.}\label{para:flickr30k-data}Flickr30k\footnote{\url{http://hockenmaier.cs.illinois.edu/Denotation.html} \label{fnote: flickr30k-dataset-url}} \shortcite{young:2014} is an extended version of the previously published Flickr8k dataset, containing images collected from Flickr$^{\ref{fnote: flickr-url}}$ and captions obtained via crowdsourcing using AMT platform, following the same strategies employed in Flickr8k. Table~\ref{table:flickr30k-dataset} presents the previously-mentioned \textit{karpathy split}$^{\ref{fnote: karpathy-split-url}}$ of the dataset.
%{$^\ref{fnote: flickr-url}$}

\begin{table}[!ht]
\small
    \centering
    \begin{tabular}{l | c c c}
    \hline  %\toprule
        \rowcolor{teal!35}
         Split      & Images        & Captions per Image   & Total Captions \\
        \hline \addlinespace[0.3em] %\midrule
        Training   &29,000          & 5                    &  145,000  \\
        Validation &1,014           & 5                    &  5,070  \\
        Test       &1,000           & 5                    & 5,000 \\ 
        \midrule
        Total      & 31,014         & 5                    &  155,070 \\ 
    \bottomrule
    \end{tabular}
    %\vspace{-1ex}
    \caption{\label{table:flickr30k-dataset} Splits of the Flickr30k image description dataset.}
\end{table}

\paragraph{Flickr30k-Entities.}\label{para:flickr30kent-data} Flickr30k-Entities\footnote{\url{http://bryanplummer.com/Flickr30kEntities} \label{fnote: flickr30k-entities-dataset-url}} \shortcite{plummer:2015} extends Flickr30k with manually annotated bounding boxes for images and entity mentions in the captions in order to accomplish the task of language grounding in images, viz. \textit{phrase localization}, while performing captioning. Specifically, there are 275,775 bounding boxes for the images of Flickr30k and 513,644 entity mentions in the 158k captions of Flickr30k. One peculiarity of this dataset is that it comes with 244k co-reference chains, in which each chain is a link between the mentions of the same entities across the five different captions of a given image. Some statistics and \textit{karpathy split}$^{\ref{fnote: karpathy-split-url}}$ of this dataset is presented in Table~\ref{table:flickr30kentities-dataset}.

\begin{table}[!ht]
\small
    \centering
    \begin{tabular}{l | c c c c c c}
    \hline  % \toprule
        \rowcolor{teal!35}
                                & Num. of   & Object        & Objects       & Objects    & Captions      & Total\\
        \rowcolor{teal!35}
        \multirow{-2}{*}{Split} & Images    & Categories    & per Category  & per Image & per Image     &Captions   \\
        \hline \addlinespace[0.3em] %\midrule
        Training                & 29,783   & -              & -             & -         & 5             & 148,915   \\
        Validation              & 1,000    & -              & -             & -         & 5             & 5,000 \\
        Test                    & 1,000    & -              & -             & -         & 5             & 5,000 \\ 
        \midrule
        Total                   & 31,783   & 44,518         & 6.2           & 8.7       & 5             & 158,915  \\ 
    \bottomrule
    \end{tabular}
    \caption{\label{table:flickr30kentities-dataset} Splits and statistics of the Flickr30k-Entities image description dataset.}
\end{table}

\paragraph{MSCOCO.}\label{para:mscoco-data}MSCOCO$^{\ref{fnote: mscoco-dataset-url}}$ \shortcite{lin:2014} is a widely-used and considerably larger-scale dataset than the image captioning datasets discussed so far. It contains natural images that are collected from Flickr$^{\ref{fnote: flickr-url}}$. The AMT platform is then used to curate and collect descriptions for the images. This dataset does not have an official split, hence the \textit{karpathy split}$^{\ref{fnote: karpathy-split-url}}$ from the above datasets is commonly used in the vision and language research community. The statistics and splits of the dataset can be found in Table~\ref{table:mscococap-dataset}.

\begin{table}[!ht]
\small
    \centering
    \begin{tabular}{l | c c c c}
    \hline  %\toprule
    \rowcolor{teal!35}
        Split                    & Images   &Captions per Image        & Total Captions     & Object Categories\\
        \hline \addlinespace[0.3em] %\midrule
        Training                & 113,287  & 5                   & 566,435  & -\\
        Validation              & 5,000    & 5                   & 25,000   & -\\
        Test                    & 5,000    & 5                   & 25,000   & -\\
        \midrule
        Total                   & 123,287  & 5                   & 616,435  & 80\\ 
    \bottomrule
    \end{tabular}
    \caption{\label{table:mscococap-dataset} Splits of the MSCOCO image description dataset.}
\end{table}

\paragraph{MSCOCO-Entities.}\label{para:mscocoent-data} MSCOCO-Entities\footnote{\url{https://github.com/aimagelab/show-control-and-tell} \label{fnote: ms-coco-entities-dataset-url}} ~\shortcite{cornia:2018} is a recently-introduced dataset based on the original MSCOCO~\shortcite{lin:2014} dataset, with the goal of achieving the twin challenges of grounding and controllability in generated image captions. Unlike Flickr30k-Entities, the grounding annotations in this dataset are obtained in a semi-automated way. Table~\ref{table:mscocoentities-dataset} presents some statistics about the dataset as well as its split.

\begin{table}[!ht]
\small
    \centering
    \begin{tabular}{l | c c c c c c}
    \hline  %\toprule
        \rowcolor{teal!35}
         Split      & Images    & Total Captions & Noun chunks    &  Noun chunks per caption & Unique Classes \\
        \hline \addlinespace[0.3em] %\midrule
        Training   & 113,287     & 545,202       & 1,518,667     & 2.79                      & 1,330\\
        Validation & 5,000       & 7,818         & 20,787        & 2.66                      &  725\\
        Test       & 5,000       & 7,797         & 20,596        & 2.64                      & 730\\
    \bottomrule
    \end{tabular}
    \caption{\label{table:mscocoentities-dataset} Splits and statistics of the MSCOCO-Entities image description dataset.}
\end{table}

\paragraph{STAIR Captions.}\label{para:stair-data} STAIR Captions\footnote{\url{http://captions.stair.center} \label{fnote: stair-captions-dataset-url}} \shortcite{yoshikawa:2017} is a large-scale Japanese image captioning dataset that provides Japanese language descriptions for the 164,062 images of MSCOCO, while retaining the same dataset splits, viz. \textit{karpathy split}$^{\ref{fnote: karpathy-split-url}}$, as with MSCOCO (see Table~\ref{table:mscococap-dataset}). The annotation of captions is done manually using crowdsourcing. Original statistics from the authors of the dataset is provided in Table~\ref{table:stairdataset}.

\begin{table}[!ht]
\small
    \centering
    \begin{tabular}{l c c c c}
    \hline  %\toprule
        \rowcolor{teal!35}
        Total Num.           & Captions    & Total Num.         & Vocabulary        & Avg. Number   \\
        \rowcolor{teal!35}
        of Images          & per Image     & of Captions        & Size              & of Chars      \\
        \hline \addlinespace[0.3em] %\midrule
        164,062 (123,287)   & 5          & 820,310 (616,435)     & 35,642 (31,938)   & 23.79 (23.80) \\
    \bottomrule
    \end{tabular}
    \caption{\label{table:stairdataset} Statistics of the STAIR Captions image description dataset (Japanese). Details on the public part of the dataset is indicated in brackets.}
\end{table}

\paragraph{Multi30k-CLID.}\label{para:multi30k-data} The Multi30k-CLID\footnote{\url{https://www.statmt.org/wmt16/multimodal-task.html} \label{fnote: multi30k-clid-2016-dataset-url}} \shortcite{elliott:2016a} dataset was designed for the task of Cross-Lingual Image Description (CLID) generation with an ultimate goal of pushing existing vision and language research towards \textit{multilingual multimodal language processing}. In the first edition of the task in 2016, the Flickr30k-Entities$^{\ref{fnote: flickr30k-entities-dataset-url}}$ dataset \shortcite{plummer:2015} was extended to the German language by crowdsourcing the descriptions independently from their English language counterparts with the help of professional translators. As with original Flickr30k, each image comes with five descriptions in German. Hence, the English-German pairs are considered as comparable, though not parallel, corpora. The splits of this dataset for English and German languages can be found in Table~\ref{table:multi30kclid-dataset-2016}.

\begin{table*}[!htbp]
\newcommand{\midruleDVSScripts}{\cmidrule(lr){1-4} } %\cmidrule(lr){3-4}
\center
\begin{tabular}{l| c | c c}
        \hline  %\toprule
            \rowcolor{teal!35}
                        &                                           &\multicolumn{2}{c}{Language of the Captions} \\\cline{3-4}
             \rowcolor{teal!35}
             \multirow{-2}{*}{Split}      & \multirow{-2}{*}{Images}            &English    &German \\
        \hline \addlinespace[0.3em] %\midruleDVSScripts
            Training                        &29,000                             &145,000   &145,000    \\
            Validation                      &1,014                              &5,070     &5,070    \\
            Testing                         &1,000                              &5,000     &5,000    \\
\bottomrule
\end{tabular}
\caption{\label{table:multi30kclid-dataset-2016} Splits and statistics of the Multi30k-CLID (2016) dataset.}
\end{table*}
%\end{wraptable}

In the second version\footnote{\url{https://www.statmt.org/wmt17/multimodal-task.html} \label{fnote: multi30k-clid-2017-dataset-url} } of the task in 2017, the Flickr30k-Entities$^{\ref{fnote: flickr30k-entities-dataset-url}}$ dataset was further extended to support French language captions~\shortcite{elliott:2017}. The annotations were again obtained via crowdsourcing following the same principles as with the previous version. Table~\ref{table:multi30kclid-dataset-2017} presents the number of instances in each language and the splits of the dataset.

\begin{table*}[!htb]
\newcommand{\midruleDVSScripts}{\cmidrule(lr){1-5} } % \cmidrule(lr){2-2} \cmidrule(lr){3-5}
\center
\begin{tabular}{l | c | c c c}
        \hline  %\toprule
            \rowcolor{teal!35}
                 &                                                  &\multicolumn{3}{c}{Language of the Captions} \\\cline{3-5}
            \rowcolor{teal!35}
             \multirow{-2}{*}{Split}      & \multirow{-2}{*}{Images}        &English    &French     &German \\
        \hline \addlinespace[0.3em] %\midruleDVSScripts
            Training                            &29,000                     &145,000   &145,000    &145,000    \\
            Validation                          &1,014                      &5,070     &5,070      &5,070    \\
            Testing                             &1,000                      &5,000     &5,000      &5,000    \\
\bottomrule
\end{tabular}
\caption{\label{table:multi30kclid-dataset-2017} Splits and statistics of the Multi30k-CLID (2017) dataset.}
\end{table*}

Similar to the earlier editions of the task, in the 2018 version\footnote{\url{http://www.statmt.org/wmt18/multimodal-task.html} \label{fnote: multi30k-clid-2018-dataset-url}} Czech language translations of the captions were added \shortcite{barrault:2018}. Following the same strategy of the prior versions of this dataset for obtaining annotations, human translators were employed to produce Czech translations for the captions of Flickr30k-Entities$^{\ref{fnote: flickr30k-entities-dataset-url}}$. Table~\ref{table:multi30kclid-dataset-2018} presents splits and statistics of all four languages of the dataset.

\begin{table*}[!ht]
\newcommand{\midruleDVSScripts}{\cmidrule(lr){1-6} } % \cmidrule(lr){3-6}
\center
\begin{tabular}{l | c | c c c c}
        \hline  %\toprule
            \rowcolor{teal!35}
                 &                                                  &\multicolumn{4}{c}{Language of the Captions} \\\cline{3-6}
            \rowcolor{teal!35}
             \multirow{-2}{*}{Split}      & \multirow{-2}{*}{Images}        &Czech     &English    &French     &German \\
        \hline \addlinespace[0.3em] % \midruleDVSScripts
            Training                        &29,000                         &145,000   &145,000   &145,000    &145,000    \\
            Validation                      &1,014                          &5,070     &5,070     &5,070      &5,070    \\
            Testing                         &1,071                          &5,355     &5,355     &5,355      &5,355    \\
\bottomrule
\end{tabular}

\caption{\label{table:multi30kclid-dataset-2018}Splits and statistics of the Multi30k-CLID (2018) dataset.}
\end{table*}

\paragraph{Conceptual Captions (CC).}\label{para:cc-data} Conceptual Captions\footnote{\url{https://ai.google.com/research/ConceptualCaptions/download} \label{fnote: conceptual-captions-dataset-url}} \shortcite{sharma:2018} is a recently introduced web-scale dataset containing more than 3.3M images paired with English language captions. The dataset was harvested from the web in an automatic manner in which the captions were extracted from the alt text of retrieved HTML webpages. As a consequence, contrary to other curated image captioning datasets in which each image is paired with five captions, the images in CC have only one description, a fact that is evident in Table~\ref{table:conceptual-captions-dataset} which also presents the dataset splits.

\begin{table}[!htbp]
%\begin{wraptable}{R}{6cm}
\small
    \centering
    \begin{tabular}{l| c c}
        \hline   %\toprule
        \rowcolor{teal!35}
         Split      &Images         & Captions  \\
        \hline \addlinespace[0.3em] %\midrule
        Training    &3,318,333      & 3,318,333  \\
        Validation  &15,840         & 15,840  \\
        Test        &22,530         & 22,530 \\
    \bottomrule
    \end{tabular}
    \caption{\label{table:conceptual-captions-dataset}Splits of the Conceptual Captions dataset.}
\end{table}
%\end{wraptable}

Although it is a large-scale dataset with a wide variety and style in captions, continued availability of the dataset for downloading by future users is a major issue, primarily due to the fact that the dataset has been distributed as a CSV file containing URLs of images. Thus, it inherently suffers from the problem of URLs becoming stale (for instance due to contents being removed, unresponsive HTTP requests, etc.), which puts the dataset at a disadvantage.

\paragraph{Personality Captions (PC).}\label{para:personality-caps-data} Personality Captions\footnote{\url{https://parl.ai/projects/personality_captions} \label{fnote: personality-captions-dataset-url}} \shortcite{shuster:2019} is a large scale image caption dataset that comes with so-called \textit{personality traits} that are useful for controllable and style-based image captioning. Thus, the samples in the PC dataset are provided as triplets (\textit{image}, \textit{personality trait}, \textit{caption}). Basic statistics such as vocabulary size, including the dataset splits, is provided in Table~\ref{table:personality-captions-dataset}.

\begin{table}[!ht]
\small
    \centering
    \begin{tabular}{l| c c c c c c}
        \hline   %\toprule
        \rowcolor{teal!35}
                                    &Num. of    & Captions    & Num. of     & Personality   &Vocabulary     &Avg. Tokens\\
         \rowcolor{teal!35}
         \multirow{-2}{*}{Split}    & Images    & per Image   & Captions    & Types         & Size          & per Caption \\
        \hline \addlinespace[0.3em] %\midrule
        Training                    &186,858    & 1           & 186,858     & 215           &33,641         &11.2 \\
        Validation                  &5,000      & 1           & 5,000       & 215           &5,460          &10.9 \\
        Test                        &10,000     & 5           & 50,000      & 215           &16,655         &11.1 \\
    \bottomrule
    \end{tabular}
    \caption{\label{table:personality-captions-dataset} Splits and statistics of the Personality Captions dataset.}
\end{table}

\subsubsection{Image Description Generation - Evaluation Measures, Models, and Results}
\label{sssec:idgall}

In this section, we describe only the evaluation measures which are used for the task of \textit{Image Description Generation}, as \textbf{Models}, \textbf{Results}, and some \textbf{Discussion} have been broadly presented in recent surveys~\shortcite{hossain:2019}.

\paragraph{Evaluation Measures.} We divide the evaluation measures into three different categories. The first set of measures is ``Language Metrics'', the second category is about ``Retrieval Metrics'', and the third category denotes ``Human Evaluation''. 
\newline
\newline
``Language Metrics'' evaluate machine-generated text based on reference text by computing similarity scores using simple n-gram statistics and word overlaps.
\begin{itemize}
\item \textbf{Bilingual Evaluation Understudy (BLEU)}~\shortcite{papineni:2002} was originally developed for machine translation to compare machine generated output with human Ground Truth (GT). BLEU calculates the overlap between predicted unigrams (BLEU-1 (B-1)), or, more generally, n-grams (BLEU-2 (B-2) with bigrams, BLEU-3 (B-3) with trigrams, BLEU-4 (B-4) with quadrigrams, and so on.) from the set of candidate and reference sentences. To achieve a high BLEU score, generated descriptions should match the human GT words as well as their order.  Maximum achievable BLEU score is 1.0 (or sometimes, equivalently 100), which is obtained when an exact match occurs between generated and reference sentence.

\item \textbf{Metric for Evaluation of Translation with Explicit Ordering}, popularly known as \textbf{METEOR}~\shortcite{banerjee:2005} has overcome some issues of BLEU, such as the need for exact word matching. Instead of a literal token matching, METEOR rather performs semantic matching by leveraging WordNet to match words at various levels, using synonymy and paraphrase matching. The METEOR score is then computed using the alignment between the machine generated output and the corresponding reference sentences. To be more specific, initially, the set of unigrams from the generated and reference sentences is used to perform an alignment. If multiple options are available for alignments between the generated and reference sentence, the alignment setting with least comparisons is preferred. After finalizing the alignment process, the METEOR score is calculated.

\item \textbf{Recall Oriented Understudy for Gisting Evaluation (ROUGE)}~\shortcite{linrouge:2004} was designed to evaluate textual summaries.  As opposed to BLEU, which concentrates on n-gram precision, ROUGE instead calculates the recall score of the generated sentences corresponding to the reference sentences. The most prominent ROUGE variant used is ROUGE-L, which is based on the longest common subsequence. Other variants include ROUGE-W (Weighted Longest Common Sub-sequence) and ROUGE-S (Skip-Bigram Co-Occurrences Statistics). One advantage of ROUGE-L over BLEU and METEOR is that it checks for subsequences within a sentence. Moreover, specifying the n-gram length (as required in BLEU) is not necessary as it is automatically incorporated.

\item \textbf{Consensus-based Image Description Evaluation (CIDEr)}~\shortcite{vedantam:2015} evaluates the consensus between a generated sentence and a set of reference sentences by performing different language pruning techniques, such as stemming and building a set of n-grams. N-grams that are common among the reference sentences of all visual data are given lower weight, as they are less informative about the visual content, and biased towards the textual content of the sentences. The weight for each n-gram is computed using Term Frequency (TF) - Inverse Document Frequency (IDF) (TF-IDF), where TF puts higher weight on frequently occurring n-grams in the reference sentence of the visual content, whereas IDF puts lower weight on commonly appearing n-grams across the whole dataset. To remove the mismatch between human evaluation and CIDEr scores, a variant of CIDEr, CIDEr-D, is used. It adds small variations, such as not performing stemming and ensuring that the words with high confidence are not repeated in a sentence by introducing a Gaussian penalty over length differences between the generated and reference sentences. As in the case of vanilla CIDEr, it produces high scores even if the sentences do not make sense.

\item \textbf{Semantic Propositional Image Captioning Evaluation (SPICE)}~\shortcite{ander:2016} measures the similarity between the scene graph tuples parsed from generated sentences and human created GT sentences. The scene graph encodes objects and their relationships through dependency parsing. Hence, it makes SPICE heavily dependent on parsing, which can be prone to errors. Similar to METEOR, SPICE uses WordNet to find and treat synonyms as positive matches when computing the F1 score between the tuples of generated sentences and the ground truth.
\end{itemize}

``Retrieval Metrics'' evaluate the machine generated text based on standard information retrieval measures~\shortcite{manning:2010} and are presented in the following paragraphs.
\begin{itemize}
    \item \textbf{Recall@k (R@k)}'s goal is to evaluate the number of relevant ground truth sentences retrieved in the Top-k (e.g., Top-1, Top-5 etc.) candidates. A higher R@k indicates better performance.
    \item \textbf{Median Rank (MedRank)} finds the median rank value of the retrieved ground truth. A lower MedRank value indicates better performance.
    \item \textbf{Mean Reciprocal Rank (MRR)} is a binary measure, where the rank of the highest ranking relevant document for a query is used to calculate the reciprocal rank averaged over all queries. A higher MRR indicates better performance.
    \item \textbf{Mean Rank (Mean)} refers to the mean rank achieved in retrieving the relevant sentence. A lower Mean value is better.
    \item \textbf{Normalized Discounted Cumulative Gain (NDCG)} is a variant of Discounted Cumulative Gain (DCG)~\shortcite{jarvelin:2000}. NDCG is a cumulative, multilevel measure of ranking quality that is usually truncated at a particular rank level.
\end{itemize}
``Human Evaluation'' employs crowd-workers to evaluate the quality of the generated content and is described in the following paragraph.
\begin{itemize}
    \item \textbf{Human Evaluation} The earlier discussed metrics provide only quantitative measures for evaluating different tasks. Due to the lack of high correlation between machine-generated textual or visual data with the human provided GT, most of the tasks, however, require human evaluations to judge the quality of the generated content. Therefore, based on the task, various kinds of instructions are given to humans who act as an evaluator in the evaluation study. In most tasks, we are interested only in finding relevance of the output to input.
\end{itemize}
\subsubsection{Video Description Generation - Introduction}
\label{sssec:videocaptiongenintro}

Going beyond images, the goal in video captioning is to comprehend the spatio-temporal information in a video for the purpose of generating either a single or multiple textual descriptions. As with image description generation (Section~\ref{sssec:imagecaptiongenintro}), we explore some of the popular types and categories of video description generation tasks in the following.

\begin{figure}[!htb]
    \centering
        \includegraphics[width=\textwidth]{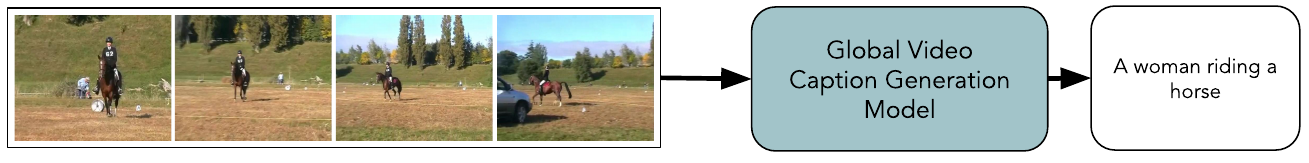}
    \caption{Given a \textit{video} (represented as sequence of frames), the Video Caption Generation Model generates a single global description.}\label{fig:standardvc}
\end{figure}

\paragraph{Global Video Description Generation.} Global video description generation approaches \shortcite{motwani:2012,regneri:2013} initially started by grounding sentences that describe actions in the visual information extracted from videos. It was further expanded into generating global natural language descriptions for videos with various approaches, for example, leveraging latent topics~\shortcite{das:2013}, corpora knowledge~\shortcite{krishnamoorthy:2013}, graphical models~\shortcite{rohrbach:2013}, and sequence-to-sequence learning~\shortcite{venugopalan:2014,venugopalan:2015,donahue:2015,srivastava:2015,xu:2016,ramanishka:2016,jin:2016}.  Figure~\ref{fig:standardvc} depicts the description generation task for a complete video.
The aforementioned approaches leverage only those training datasets with a limited set of visual objects. However, the recognition and description of entities and activities in real-world videos is more difficult. Nevertheless, generating natural language descriptions for such videos is addressed with a factor graph by combining visual detection with language statistics~\shortcite{thomason:2014}.

Additionally, sequence-to-sequence (seq2seq) based approaches have been improved with external corpora~\shortcite{venugopalan:2016} and also using \textit{attention} with various techniques such as soft-attention~\shortcite{yao:2015}, multimodal fusion~\shortcite{hori:2017}, temporal attention~\shortcite{song:2017}, semantic consistency~\shortcite{gao:2017}, and residual connections~\shortcite{li:2018}. Apart from attention-based methods, novel architectures have also been explored, such as incorporation of semantic attributes learned from videos~\shortcite{pan:2017}, ensemble-based description generator networks~\shortcite{shetty:2018} and encoder-decoder-reconstructors which leverage both the forward and backward flows, i.e., video-to-description and description-to-video, for video captioning~\shortcite{wangrec:2018}. Multi-faceted attention has also been used to select the most salient visual features or semantic attributes, with which an overall sentence is generated~\shortcite{long:2018}.

Apart from architecture improvements, different machine learning approaches have also been explored. Video captioning has been tackled using a multi-task learning scenario by sharing knowledge between two related tasks (such as temporal- and context-aware video) combined with entailment generation task~\shortcite{pasunurua:2017}. Other approaches have leveraged reinforcement learning, either by providing entailment rewards~\shortcite{pasunurue:2017} , or to address the description generation for multiple fine-grained actions~\shortcite{wang:2018}. Further,~\shortciteA{mazaheri:2018} proposed a deep network designed to detect inaccuracies in a sentence, and fix them by replacing the inaccurate word(s) with the help of a Visual Text Correction system. Recently, Zhang et al.~\shortciteA{zhang-vatex:2020} introduced an object relational graph (ORG) based encoder which encapsulates the relation among visual objects to build richer representation and a decoder the integrates the external language model to capture abundant linguistic knowledge for efficient video description generation.

In the following, we discuss some related ideas which expand the scope of video description generation.

\paragraph{Dense Video Description Generation.} The aim of dense video description generation is to achieve fine-grained video understanding by addressing two sub-problems: (1) localizing events in a video, and (2) generating captions for these localized events~\shortcite{zhou:2018,xu:2019}. Further, extending earlier research, some approaches~\shortcite{zhouvc:2018} have explicitly linked the sentence to a corresponding bounding box in one of the frames of a video by annotating each of the noun phrases observed in the sentence. Incorporating background knowledge for video description generation is also another line of research~\shortcite{whitehead:2018}. However, the core challenge, namely the automatic evaluation of video captioning, is still unsolved. It is currently being studied from the perspective of direct assessment with the help of human assessors~\shortcite{graham:2018}.

\paragraph{Movie Description Generation.} Movie description generation perceives the video description generation task from a different perspective, in which movie clips are used as inputs. Initially, aligning books to movies~\shortcite{tapaswi:2015,zhu:2015} was used to generate story-like explanations. Later, movie descriptions~\shortcite{rohrbach:2015} were directly created by transcribing audio descriptions by concentrating on precisely describing what is shown in the movie scenes.

\subsubsection{Video Description Generation - Datasets}
\label{ssec:vddatasets}
Similar to the \textit{image} description generation task, several datasets have been created to address the task of \textit{video} description generation. In the following, we cover those datasets that are popular and extensively used. For the sake of brevity, we denote \textit{hours} $\rightarrow$ \textit{h}, \textit{minutes} $\rightarrow$ \textit{m}, and \textit{seconds} $\rightarrow$ \textit{s}.

\paragraph{Microsoft Video Description (MSVD).}\label{para:msvd-data}MSVD\footnote{\url{https://www.cs.utexas.edu/users/ml/clamp/videoDescription} \label{fnote: msvd-dataset-url}}~\shortcite{chen:2011} is an open domain dataset collected from YouTube clips and annotated using AMT. The dataset is multilingual and contains human generated descriptions in languages such as German, English, Chinese, etc. On average, there are forty-one single sentence descriptions per clip. More statistics about the dataset are presented in Table~\ref{table:msvd-dataset} whereas Table~\ref{table:msvd-dataset-split} presents its split.
\begin{table*}[!ht]
\center
\begin{tabular}{l c c c c c c c c}
    \hline  %\toprule
    \rowcolor{teal!35}
        \multicolumn{1}{c}{Total} &\multicolumn{1}{c}{Total} &\multicolumn{1}{c}{Total}  &\multicolumn{1}{c}{Avg.} &\multicolumn{1}{c}{Total} &\multicolumn{1}{c}{Total} &\multicolumn{1}{c}{Total}  &\multicolumn{1}{c}{Vocabulary} \\
    \rowcolor{teal!35}
        Videos     &Classes         &Length       &Length   &Clips    &Sentences      &Words  & Size \\
    \hline \addlinespace[0.3em]  %\midrule
        1,970      &218     &5.3 h     &10 s   &1,970     &70,028        &607,339    &13,010    \\
\bottomrule
\end{tabular}
\caption{\label{table:msvd-dataset} Statistics of the MSVD dataset.}
\end{table*}

\begin{table}[!ht]
\small
    \centering
    \begin{tabular}{l | c c}
    \hline  %\toprule
    \rowcolor{teal!35}
        Split       & Frames    & Videos \\
    \hline \addlinespace[0.3em]    %\midrule
        Training     &33,682    &  1,200  \\
        Validation   &3,275     & 100   \\
        Test         &20,528    & 670    \\
    \midrule
        Total        &57,485    & 1970  \\
    \bottomrule
    \end{tabular}
    \caption{\label{table:msvd-dataset-split} Splits of the MSVD dataset.}
\end{table}

\paragraph{MPII Cooking Activities.}\label{para:mpiicook-data} The MPII Cooking\footnote{\url{https://www.mpi-inf.mpg.de/departments/computer-vision-and-machine-learning/research/human-activity-recognition/mpii-cooking-activities-dataset} \label{fnote: mpii-cooking-dataset-url}}~\shortcite{rohrbach:2012} dataset consists of 65 different cooking activities such as ``wash hands'', ``put in bowl'', etc., when participants are preparing one of 14 dishes such as \textit{fruit salad}, \textit{casserole}, etc. The dish preparation time ranges between 3 and 41 minutes. The videos are recorded in high resolution (1624x1224), following which the \textit{activity} annotations are manually created by 6 people. Table~\ref{table:mpii-cooking-dataset} presents more statistics about the dataset whereas the splits of it can be found in Table~\ref{table:mpii-cooking-dataset-split}.

\begin{table}[!ht]
\small
    \centering
    \begin{tabular}{l c c c c c c c c}
    \hline  %\toprule
    \rowcolor{teal!35}
        Num. of      &Total &Total  &Total     &Video       &Total     &Num. of      &Total     & Activity  \\
    \rowcolor{teal!35}
        Subjects    &Clips &Videos &Frames     &Length      &Length    &Activities   &Dishes    &Annotations    \\
    \hline \addlinespace[0.3em]  %\midrule
        12          &5,609  &44     &881,755   &3 to 41 m    &8.0 h     &65           &14        &5,609  \\
    \bottomrule
    \end{tabular}
    \caption{\label{table:mpii-cooking-dataset} Statistics of the MPII Cooking Activities dataset.}
\end{table}

\begin{table}[!ht]
\small
    \centering
    \begin{tabular}{l | c c}
    \hline  %\toprule
    \rowcolor{teal!35}
        Split        & Frames     & Subjects \\
    \hline \addlinespace[0.3em] %\midrule
        Training     & 1,071      & 10   \\
        Validation   & -          & -   \\
        Test         & 1,277      & 7    \\
    \bottomrule
    \end{tabular}
    \caption{\label{table:mpii-cooking-dataset-split} Splits of the MPII Cooking dataset.}
\end{table}

\paragraph{YouCook.}\label{para:youcook-data} YouCook\footnote{\url{http://web.eecs.umich.edu/~jjcorso/r/youcook} \label{fnote: youcook-dataset-url}}~\shortcite{das:2013} is a more complex real-world cooking dataset when compared to MPII Cooking in which the complexity arises because of dynamic scene and camera changes. The videos are all downloaded from YouTube and are broadly categorized into 6 different cooking styles, viz. baking, grilling, etc. Video descriptions are obtained via crowdsourcing using AMT. On average, eight descriptions are collected per video. Frames are annotated with objects belonging to \textit{categories} (such as bowls, utensils, etc.) and \textit{actions}. More details and splits of the dataset can be found in Table~\ref{table:youcook-dataset} and Table~\ref{table:youcook-dataset-split} respectively.
\begin{table}[!ht]
\small
    \centering
    \begin{tabular}{l c c c c c c}
    \hline  %\toprule
      \rowcolor{teal!35}
        Cooking    &Object   &Total   &Total   &Num. of      &Num. of     &Vocabulary    \\
      \rowcolor{teal!35}
        Styles     &Classes  &Videos  &Length  &Sentences    &Words       &Size  \\
        \hline \addlinespace[0.3em] %\midrule
        6          &10       &88      &2.3 h   &2,688        &42,457      &2,711  \\
    \bottomrule
    \end{tabular}
    \caption{\label{table:youcook-dataset} Statistics of the YouCook dataset.}
\end{table}

\begin{table}[!ht]
\small
    \centering
    \begin{tabular}{l | c }
    \hline  %\toprule
        \rowcolor{teal!35}
        Split       &Videos  \\
        \hline \addlinespace[0.3em] %\midrule
        Training     &49   \\
        Validation   & -    \\
        Test         &39    \\
    \bottomrule
    \end{tabular}
    \caption{ \label{table:youcook-dataset-split} Splits of the YouCook dataset.}
\end{table}

\paragraph{YouCook II.}\label{para:youcookii-data}Similar to the YouCook dataset, YouCook II\footnote{\url{http://youcook2.eecs.umich.edu} \label{fnote: youcook-ii-dataset-url}}~\shortcite{zhou2:2018} also consists of instructional cooking videos that are all collected from YouTube. The videos include 89 cooking recipes from four regions: South Asia, East Asia, Europe/Middle East, and America. One unique aspect of this dataset when compared to previously discussed video description datasets is that that the videos are annotated with \textit{procedure segments} that contain rich semantic information. Table~\ref{table:youcook-2-dataset} presents the statistics about the dataset.
\begin{table}[!ht]
\small
    \centering
    \begin{tabular}{l c c c c c c c}
    \hline  %\toprule
        \rowcolor{teal!35}
        Cooking   &Total    &Total Video   &Avg. Video  &Procedure     &Total      &Num. of     &Vocab.    \\
        \rowcolor{teal!35}
        Recipes   &Videos   &Length        &Length      &Seg. per Video          &Clips        &Sentences   &Size  \\
        \hline \addlinespace[0.3em] %\midrule
         89        &2,000   & 175.6 h      &316 s       & 3-16              &15,400       &15,400      &2,600  \\
    \bottomrule
    \end{tabular}
    \caption{ \label{table:youcook-2-dataset} Statistics of the YouCook II dataset.}
\end{table}

For each recipe, the videos are randomly split into training, validation, and testing in ratios of 67\%, 23\%, and 10\% respectively. The actual numbers are presented in Table~\ref{table:youcook-2-dataset-split}.

\begin{table}[!ht]
\small
    \centering
    \begin{tabular}{l | c}
    \hline  %\toprule
    \rowcolor{teal!35}
        Split       & Videos  \\
        \hline \addlinespace[0.3em] %\midrule
        Training     & 1,340   \\
        Validation   & 460   \\
        Test         & 200    \\
    \bottomrule
    \end{tabular}
    \caption{\label{table:youcook-2-dataset-split} Splits of the YouCook II dataset.}
\end{table}

\paragraph{Textually Annotated Cooking Scenes (TACoS).}\label{para:tacos-data} The TACoS\footnote{\url{https://www.coli.uni-saarland.de/projects/smile/page.php?id=tacos} \label{fnote: tacos-dataset-url}}~\shortcite{regneri:2013} dataset is an extended version of a subset of MPII Composites~\shortcite{rohrbachscript:2012} which contains cooking videos that are each annotated with multiple textual descriptions. It contains only those videos that include activities such as manipulation of cooking ingredients. Around 26 cooking activities are collected with 127 videos. More statistics on the dataset is presented in Table~\ref{table:tacos-dataset-1} and Table~\ref{table:tacos-dataset-2}. For building and evaluating models, the dataset is split into 50\% for training, 25\% for validation, and 25\% for testing.
\begin{table*}[!ht]
\small
\centering
\begin{tabular}{l c c c c c c c}
\hline  %\toprule
    \rowcolor{teal!35}
        \multicolumn{1}{l}{Total}  &\multicolumn{1}{c}{Total}    &\multicolumn{1}{c}{Descriptions} &\multicolumn{1}{c}{Annotation} &\multicolumn{1}{c}{Annotations} &\multicolumn{1}{c}{Cooking} &\multicolumn{1}{c}{Action}   \\
    \rowcolor{teal!35}
        Videos        & Clips       &per Video  &Assignments        & after filtering       &Tasks/Dishes      &Descriptions  \\
    \hline \addlinespace[0.3em] %\midrule
        127           & 7,206      &20          &2,540              &2,206                  &26                &17,334 (tokens)  \\
\bottomrule
\end{tabular}
\caption{\label{table:tacos-dataset-1} The TACoS dataset statistics - I.}
\end{table*}
\begin{table*}[!ht]
\small
\centering
\begin{tabular}{l c c c c c c}
    \hline  %\toprule
    \rowcolor{teal!35}
        \multicolumn{1}{l}{Sentence}   &\multicolumn{1}{l}{Total}  &\multicolumn{1}{c}{Content Words}  &\multicolumn{1}{c}{Num. of} &\multicolumn{1}{c}{Num. of} \\
    \rowcolor{teal!35}
        Types                          &Words                     &(viz. nouns, verbs, adjectives)     &Verbs (tokens)               & Verbs (lemmas) \\
    \hline \addlinespace[0.3em] %\midrule
        11,796                         &146,771                    &75,210	                           &28,292                    &435  \\
\bottomrule
\end{tabular}
\caption{\label{table:tacos-dataset-2}The TACoS dataset statistics - II.}
\end{table*}

\paragraph{TACoS-MultiLevel.}\label{para:tacosmlevel-data}The above discussed TACoS dataset was extended into TACoS-MultiLevel\footnote{\url{https://www.mpi-inf.mpg.de/departments/computer-vision-and-machine-learning/research/vision-and-language/tacos-multi-level-corpus} \label{fnote: tacos-multilevel-dataset-url}}~\shortcite{rohrbachcoherent:2014} by collecting three levels of descriptions constituting (i) 15 detailed descriptions per video, (ii) 3-5 short descriptions, and (iii) a single sentence description, using AMT platform. Overall, the dataset comes with 2,600 triplets of descriptions. Further statistics on the dataset can be found in Table~\ref{table:tacos-mlevel-dataset}.

\begin{table*}[!ht]
\small
\centering
\begin{tabular}{l c c c c c c}
    \hline  %\toprule
    \rowcolor{teal!35}
        \multicolumn{1}{l}{Total}   &\multicolumn{1}{l}{Total}    &\multicolumn{1}{l}{Total Video}  &\multicolumn{1}{c}{Avg.}  &\multicolumn{1}{c}{Number of} &\multicolumn{1}{c}{Total}  \\
    \rowcolor{teal!35}
        Videos      & Clips        &Length       &Length     &Sentences      &Words  \\
    \hline \addlinespace[0.3em] %\midrule
        185         & 14,105      &27.1 h     &360 s   &52,593        &2,000    \\
\bottomrule
\end{tabular}
\caption{\label{table:tacos-mlevel-dataset} Statistics of the TACoS-MultiLevel dataset.}
\end{table*}

\paragraph{MPII Movie Description (MPII-MD).}\label{para:mpiimd-data}MPII-MD\footnote{\url{https://www.mpi-inf.mpg.de/departments/computer-vision-and-machine-learning/research/vision-and-language/mpii-movie-description-dataset} \label{fnote: mpii-md-dataset-url}} \shortcite{rohrbach:2015} dataset contains clips extracted from Hollywood movies and their transcribed audio descriptions. In addition, each clip is paired with a single sentence that is extracted from the script of the movie. Furthermore, transcribed audio is associated with spoken sentences by using timestamps. Misalignment between the audio and visual content is handled by leveraging manual annotation. Additional statistics on the dataset is presented in Table~\ref{table:mpii-md-dataset}.

\begin{table*}[!ht]
\small
\centering
\begin{tabular}{l | c | c  | c c c c c}
        \hline  %\toprule
        \rowcolor{teal!35}
                            &Unique    & Before alignment  & \multicolumn{5}{c}{After alignment} \\\cline{3-8}
    %\hline %\midrule
        \rowcolor{teal!35}
\multirow{-2}{*}{}  & Movies    & Words  & Words     & Sentences     & Clips     & Avg. Length   & Total \\
    \hline \addlinespace[0.3em] %\midrule
Audio Desc.     & 55	    & 346,557           &332,846    &37,272         &37,266     &4.1 s          & 42.5 h \\
Movie script    & 50	    & 398,072           &320,621    &31,103         &31,071     &3.6 s          & 31.1 h \\
\midrule
Total           & 94	    & 744,629           &653,467    &68,375         &68,337     &3.9 s          & 73.6 h \\
\bottomrule
\end{tabular}
\caption{\label{table:mpii-md-dataset} Statistics of the MPII-MD dataset.}
\end{table*}

For the task of video description, the MPII-MD dataset is split as follows: 11 movies with associated scripts and audio descriptions (in total 22 alignments, 2 per movie) are used as validation (8) and test sets (14). The remaining 83 movies are used for training purposes.

\paragraph{Montreal Video Annotation Dataset (M-VAD).}\label{para:mvad-data}M-VAD\footnote{\url{https://mila.quebec/en/publications-archive/public-datasets/m-vad/} \label{fnote: m-vad-dataset-url}} \shortcite{torabi:2015} is a large Descriptive Video Service (DVS)-derived video dataset that is created using 92 Movies, covering a wide variety of genres. It is collected in a semi-automatic manner with minimal human intervention. The words in the descriptions are annotated with Part-Of-Speech (POS) tags using the Stanford POS tagger. Around 500 proper names are removed from the corpus, since learning proper names is not interesting for a video description model.

\begin{table}[!ht]
\small
    \centering
    \begin{tabular}{l c c c c c c}
    \hline  %\toprule
    \rowcolor{teal!35}
        Type        & Movies    & Words     & Paragraphs    & Sentences     & Avg. Length   & Total \\
    \hline \addlinespace[0.3em]    %\midrule
        Un-filtered &92         &531,778    &52,683         &59,415         &6.3 s          &91 h \\
        Filtered    &92         &510,933    &48,986         &55,904         &6.2 s          &84.6 h \\
    \bottomrule
    \end{tabular}
    \caption{\label{table:m-vad-dataset} Statistics of the M-VAD dataset.}
\end{table}

Table~\ref{table:m-vad-dataset} presents some statistics about the dataset, while Table~\ref{table:m-vad-dataset-split} presents the official dataset split that balances the genre within each split.

\begin{table}[!ht]
\small
    \centering
    \begin{tabular}{l | c}
    \hline  %\toprule
    \rowcolor{teal!35}
        Split       & Video Clips  \\
        \hline \addlinespace[0.3em] %\midrule
        Training     & 38,949   \\
        Validation   & 4,888   \\
        Test         & 5,149    \\
    \bottomrule
    \end{tabular}
    \caption{\label{table:m-vad-dataset-split} Splits of the M-VAD dataset.}
\end{table}

\paragraph{MSR Video to Text (MSR-VTT).}\label{para:msrvtt-data}MSR-VTT\footnote{\url{http://ms-multimedia-challenge.com/2017/dataset} \label{fnote: msr-vtt-dataset-url}} \shortcite{xu:2016}, also known as MSR-VTT-10k, is a large-scale video dataset containing automatically crawled videos belonging to 20 categories for the task of video description generation. The sentence annotations are obtained via crowdsourcing using AMT. In addition to the video content, the dataset also contains audio information. Table~\ref{table:msr-vtt-dataset} presents more statistics about the dataset.

\begin{table}[!ht]
\small
    \centering
    \begin{tabular}{l c c c c c c c c}
    \hline  %\toprule
    \rowcolor{teal!35}
        Categories &Videos  & Clips     & Sentences per Clip    & Sentences & Words         & Vocab.        & Duration \\
        \hline \addlinespace[0.3em]  %\midrule
        20         &7,180   & 10,000    & 20                    & 200,000   & 1,856,523     &29,316         &41.2 h\\
    \bottomrule
    \end{tabular}
    \caption{ \label{table:msr-vtt-dataset} Statistics of the MSR-VTT dataset.}
\end{table}

Out of 7.2k videos, 30k video clips have been created. However, only a random subset of 10k clips has been released. The dataset is split in the ratio of 65\%:30\%:5\% for training, validation, and testing. Specific numbers are presented in Table~\ref{table:msr-vtt-dataset-split}.

\begin{table}[!ht]
\small
    \centering
    \begin{tabular}{l | c}
    \hline  %\toprule
    \rowcolor{teal!35}
        Split & Video Clips \\
        \hline \addlinespace[0.3em] %\midrule
        Training    & 6,513  \\
        Validation  & 497   \\
        Test        & 2,990   \\
    \bottomrule
    \end{tabular}
    \caption{ \label{table:msr-vtt-dataset-split} Splits of the MSR-VTT dataset.}
\end{table}

\paragraph{Videos Titles in the Wild (VTW).}\label{para:vtw-data}VTW\footnote{\url{http://aliensunmin.github.io/project/video-language/index.html#VTW} \label{fnote: vtw-dataset-url}} \shortcite{zeng:2016} is a large-scale dataset of automatically crawled user-generated YouTube videos paired with titles and descriptions. The video clips are on average 90 seconds in duration and are described with one sentence per clip to enable video title generation. It also comes with augmented sentences that contain information that may not be present in the video clip. More statistics of the dataset can be found in Table~\ref{table:videoinwild-dataset}.

\begin{table}[!ht]
\small
    \centering
    \begin{tabular}{l c c c c c c c c}
    \hline
        \rowcolor{teal!35}
        Dataset     & Sentences     & Vocab.     & Sentences/Word & Nouns    & Verbs     & Adjective     & Adverb \\
        \hline \addlinespace[0.3em]  %\midrule
        VTW-title    &18,100        & 8,874     & 2.0           & 5,850     & 2,187     &1,187          &224 \\
        VTW-full     &44,603        & 23,059    & 1.9           & 13,606    & 6,223     &3,967          &846 \\
    \bottomrule
    \end{tabular}
    \caption{ \label{table:videoinwild-dataset} Statistics of the VTW dataset.}
\end{table}

Similar to M-VAD, the dataset is randomly split into 80\% for training and 10\% each for validation and testing. Specific numbers are presented in Table~\ref{table:videoinwild-dataset-split}.

\begin{table}[!ht]
\small
    \centering
    \begin{tabular}{ l | c c }
    \hline
        \rowcolor{teal!35}
        Split           & Videos    & Sentences/Titles \\
        \hline \addlinespace[0.3em]  %\midrule
        Training        &14,100     &14,100  \\
        Validation      &2,000      & 2,000  \\
        Test            &2,000      & 2,000 \\
    \bottomrule
    \end{tabular}
    \caption{\label{table:videoinwild-dataset-split} Splits of the VTW dataset.}
\end{table}

\paragraph{ActivityNet Captions (ANetCap).}\label{para:anetcap-data}ANetCap\footnote{\url{http://activity-net.org/challenges/2017/captioning.html} \label{fnote: activitynet-dataset-url}} \shortcite{krishnaanet:2017} is a large-scale video dataset\footnote{\url{https://cs.stanford.edu/people/ranjaykrishna/densevid} \label{fnote: anetcap-dataset-url}} that extends a subset of videos from ActivityNet with dense descriptions. There are multiple descriptions for every video and the videos contain multiple events occurring at the same time. Another notable aspect of this dataset is that the descriptions focus more on actions happening in videos. As a result, this dataset falls under the category of being more action-centric than object-centric.

\begin{table}[!ht]
\small
    \centering
    \begin{tabular}{l c c c c c}
    \hline
        \rowcolor{teal!35}
        Videos      & Total Video Hours        & Avg. Video Length   & Sentences     & Avg. Sentence Length\\
        \hline \addlinespace[0.3em]  %\midrule
        20,000      & 849        & 180 s         & 100,000       & 13.48 (words)\\
    \bottomrule
    \end{tabular}
    \caption{\label{table:activitynet-dataset} Statistics of the ANetCap dataset.}
\end{table}

 Table~\ref{table:activitynet-dataset} presents more statistics on the dataset, while Table~\ref{table:activitynet-dataset-split} presents its split.

\begin{table}[!ht]
\small
    \centering
    \begin{tabular}{ l | c }
    \hline
        \rowcolor{teal!35}
        Split           & Videos \\
        \hline \addlinespace[0.3em]  %\midrule
        Training        & 10,024  \\
        Validation      & 4,926   \\
        Test            & 5,044   \\
    \bottomrule
    \end{tabular}
    \caption{\label{table:activitynet-dataset-split} Splits of the ANetCap dataset.}
\end{table}

\paragraph{ActivityNet Entities (ANetEntities).}\label{para:anetentities-data} The ANetEntities\footnote{\url{https://github.com/facebookresearch/ActivityNet-Entities} \label{fnote: activitynet-entities-dataset-url}} \shortcite{zhouvc:2018} dataset augments ANetCap \shortcite{krishnaanet:2017} with manually annotated bounding boxes, and was created for the task of grounding language in videos while generating descriptions. It adds around 158k bounding box annotations on ANetCap, each grounded to a Noun Phrase (NP) in the sentence description. More statistics and the dataset splits can be found in Table~\ref{table:activitynet-entities-dataset-split}.

\begin{table}[!ht]
\small
    \centering
    \begin{tabular}{l | c c c c}
    \hline  %\toprule
    \rowcolor{teal!35}
        Split           & Videos    & Sentences      & Objects       & Bounding Boxes  \\
        \hline \addlinespace[0.3em] %\midrule
        Training        & 10,000    & 35,000        & 432           & 105,000   \\
        Validation      & 2,500     &  8,600        & 427           & 26,500    \\
        Test            & 2,500     &  8,500        & 421           & 26,100    \\
        \hline \addlinespace[0.3em] %\midrule
        Total           & 15,000    & 52,100        & 432           & 157,600   \\
       \bottomrule
    \end{tabular}
    \caption{\label{table:activitynet-entities-dataset-split} Statistics and splits of the ANetEntities dataset.}
\end{table}

\paragraph{COmprehensive INstructional video analysis (COIN).}\label{para:coin-data}COIN\footnote{\url{https://coin-dataset.github.io} \label{fnote:coin-dataset-url}} \shortcite{tang:2019} is a large-scale dataset of instructional YouTube videos from 12 domains such as vehicles, gadgets, sports, etc., that are common in our daily lives. It is aimed at overcoming two limitations of current instructional video datasets, namely \textit{diversity} and \textit{scale}. It covers over 180 tasks in 12k videos.

\begin{table}[!ht]
\small
    \centering
    \begin{tabular}{l c c c c c c}
    \hline  %\toprule
    \rowcolor{teal!35}
    \multicolumn{1}{l}{Num. of}  &\multicolumn{1}{c}{Num. of}	&\multicolumn{1}{c}{Total}    &\multicolumn{1}{c}{Total} &\multicolumn{1}{c}{Total} &\multicolumn{1}{c}{Avg. Video} &\multicolumn{1}{c}{Avg. Segment}  \\
        \rowcolor{teal!35}
        Domains             &Tasks   &Videos  &Segments   &Duration       &Length    &Length \\
        \hline \addlinespace[0.3em] %\midrule
         12                 &180     &11,827  &46,354     &476 h, 38 m    &2.36 m    &14.91 s          \\
    \bottomrule
    \end{tabular}
    \caption{ \label{table:coin-dataset} Statistics of the COIN dataset.}
\end{table}

One unique aspect of this dataset is that it introduces a three-level hierarchy, viz. \textit{domain}, \textit{task}, and \textit{step}, for organizing videos. Table~\ref{table:coin-dataset} shows some statistics of the dataset whereas Table~\ref{table:coin-dataset-split} presents training and validation splits of COIN.

\begin{table}[!ht]
\small
    \centering
    \begin{tabular}{l | c}
    \hline  %\toprule
    \rowcolor{teal!35}
        Split       & Videos  \\
        \hline \addlinespace[0.3em] %\midrule
        Training     & 9,030   \\
        Validation   & -  \\
        Test         & 2,797    \\
    \bottomrule
    \end{tabular}
    \caption{\label{table:coin-dataset-split} Splits of the COIN dataset.}
\end{table}

\paragraph{HowTo100M.}\label{para:howto100m-data}HowTo100M\footnote{\url{https://www.di.ens.fr/willow/research/howto100m} \label{fnote:howto100m-dataset-url}} \shortcite{miech-howto100m:2019} is a large-scale dataset of narrated videos with emphasis on instructional YouTube videos where the video creators teach complex tasks with an explicit intention of explaining the visual content on screen. The dataset includes a wide variety of 23k activities from the domains such as gardening, personal care, fitness, hand crafting, cooking, etc. and is three orders of magnitude than the previously discussed video description datasets. Table~\ref{table:howto100m-dataset} presents more statistics about the dataset.

\begin{table}[!ht]
\small
    \centering
    \begin{tabular}{l c c c c c c c}
    \hline  %\toprule
    \rowcolor{teal!35}
    \multicolumn{1}{l}{Num. of}  &\multicolumn{1}{c}{Num. of}	&\multicolumn{1}{c}{Total}    &\multicolumn{1}{c}{Total}   &\multicolumn{1}{c}{Total}  &\multicolumn{1}{c}{Total} &\multicolumn{1}{c}{Avg. Video}  &\multicolumn{1}{c}{Avg. Clip-Caption}\\
        \rowcolor{teal!35}
        Domains     &Tasks      &Videos  &Clips   &Duration     &Captions   &Length     &Pairs per Video\\
        \hline \addlinespace[0.3em] %\midrule
         12         &23,611    &1.221M  &136M    &134,472 h     &136M       &6.5 m      &110 \\
    \bottomrule
    \end{tabular}
    \caption{ \label{table:howto100m-dataset} Statistics of the HowTo100M dataset.}
\end{table}

This dataset has not yet been used for the task of video description generation. Hence, an official dataset split is not available for evaluation purposes.

\subsubsection{Video Description Generation - Evaluation Measures, Models, and Results}
\label{sssec:vdgall}

In this section, we describe only the evaluation measures which are used for the task of \textit{Image Description Generation} as \textbf{Models}, \textbf{Results}, and some \textbf{Discussion} have been broadly discussed in recent surveys~\shortcite{aafaq:2018}.

\paragraph{Evaluation Measures.} The measures used for \textit{Video Description Generation} are the same as the Language metrics and Retrieval metrics used in \textit{Image Description Generation} and are presented in the Section~\ref{sssec:idgall}.

\subsection{Visual Storytelling}
\label{ssec:vst}
The task of visual storytelling aims to encode a sequence of images or frames (in the video) to generate a paragraph which is story-like. This is usually considered more beneficial than generating a paragraph from a single image or video.

\subsubsection{Image Storytelling - Introduction}
\label{sssec:imstintro}
\begin{figure}[!htb]
    \centering
        \includegraphics[width=\textwidth]{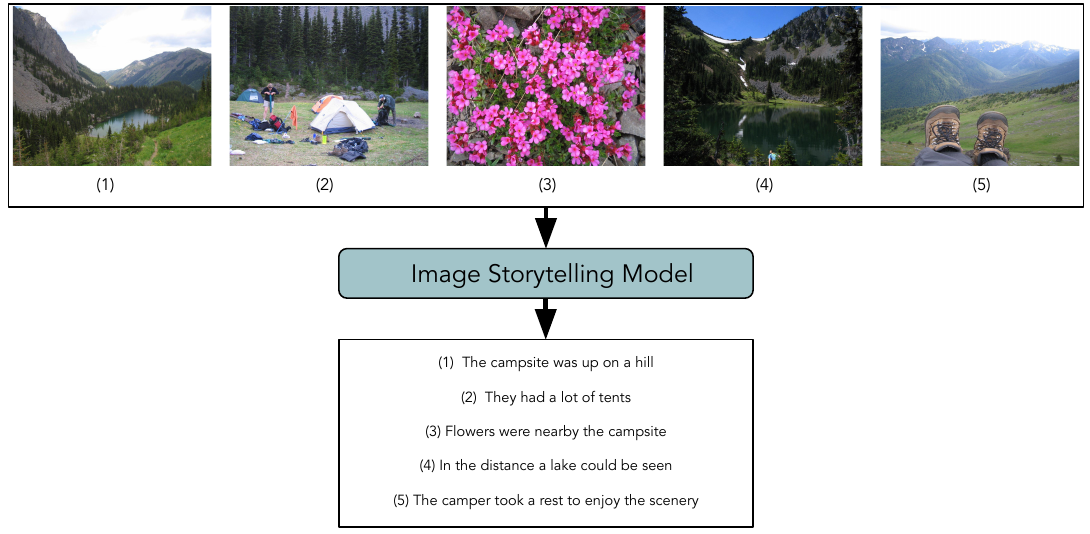}
    \caption{Given a \textit{sequence of images}, an Image Storytelling Model generates a textual story in sequence.}\label{fig:imagestory}
\end{figure}

The aim of image storytelling is to generate stories from a sequence of images. Although sequence of images can be perceived as a video, consecutive images in the streams can have sharp changes of visual content, which can cause an abrupt discontinuity between consecutive sentences~\shortcite{park:2015}. Hence, it is seen as a sequential vision-to-language task~\shortcite{huang:2016} where images are not considered in isolation. Figure~\ref{fig:imagestory} shows a schematic representation of image storytelling where a story in a sequence is generated.

Initially, semantic coherence in a photo stream is captured by reducing the visual variance. Further, the semantic space is acquired by jointly embedding each photo with its corresponding contextual sentence such that their correlations are discovered~\shortcite{liupt:2017}. It was then improved by exploiting hierarchical architecture~\shortcite{yualbum:2017} and further optimized by incorporating reinforcement learning with rewards~\shortcite{wangsrt:2018} for generating relevant and expressive narrative paragraphs. Instead of \textit{flat deep reinforcement learning}, a hierarchically structured reinforced training has also been studied~\shortcite{huang:2018} and has been shown to achieve significantly better performance than with a flat structure. Similarly, ~\shortciteA{wangno:2018} used adversarial reward learning to learn an implicit reward function from human demonstrations to optimize policy search with the learned reward function.

Nevertheless, the standard form of narration suffers from repetitiveness, with the same objects or events serving to undermine a good story structure. Hence, inter-sentence diversity was explored with diverse beam search to generate more expressive stories~\shortcite{hsu:2018}. The task has also been approached from a different perspective, in which, given a jumbled set of aligned image-description pairs that belong to a story, the task is to sort them such that the output sequence forms a coherent story~\shortcite{agrawal:2016}. 

While earlier research addresses only natural images, some approaches~\shortcite{limed:2019} also incorporated medical domain knowledge to generate realistic and accurate descriptions for medical images.

\subsubsection{Image Storytelling - Datasets}
\label{sssec:imstorydata}
There are not many datasets created to address the creative task of image storytelling. In the following, we cover all datasets that have been used to advance this artistically interesting and challenging problem.

\paragraph{New York City Storytelling (NYC-Storytelling).}The \label{para:nycdata}NYC-Storytelling\footnote{\url{https://github.com/cesc-park/CRCN} \label{fnote: nyc-storytelling-dataset-url}} \shortcite{park:2015} dataset was created from blogs in which users post their travelogues. The dataset is collected in a semi-automatic manner: automatic crawling followed by manual selection of travelogues and finally preprocessing using the NLTK\footnote{\url{https://www.nltk.org}\label{fnote: nltk-url}} library. For evaluation purposes, the dataset is split in a ratio of 8:1:1 for training, validation, and testing respectively. Table~\ref{table:nyc-storytelling-dataset} presents minimal statistics of the dataset.

\begin{table}[!ht]
\small
    \centering
    \begin{tabular}{l c}
    \hline  %\toprule
    \rowcolor{teal!35}
        Images      & Blog posts \\
        \hline \addlinespace[0.3em] %\midrule
        78,467      &11,863 \\
       \bottomrule
    \end{tabular}
    \caption{\label{table:nyc-storytelling-dataset} Statistics of the NYC-Storytelling dataset.}
\end{table}

\paragraph{Disneyland Storytelling.}\label{para:disneydata}Similar to NYC-Storytelling, Disneyland Storytelling is also based on blogs documenting travelogues but specifically about \textit{Disneyland Park}. This dataset was originally created by \shortcite{kim:2015} but has been reused for visual storytelling tasks. The same ratio of data splits as with the NYC-Storytelling dataset is used for evaluation purposes. The minimal statistics of the dataset can be found in Table~\ref{table:disneyland-dataset}.

\begin{table}[!ht]
\small
    \centering
    \begin{tabular}{l c}
    \hline  %\toprule
    \rowcolor{teal!35}
        Images      & Blog posts \\
        \hline \addlinespace[0.3em] %\midrule
        60,545      &7,717 \\
       \bottomrule
    \end{tabular}
    \caption{\label{table:disneyland-dataset} Statistics of the Disneyland-Storytelling dataset.}
\end{table}

\paragraph{Sequential Image Narrative Dataset (SIND).}\label{para:sinddata}SIND \shortcite{huang:2016} is the first large-scale dataset created for the task of image storytelling. Natural language descriptions of the dataset are divided into three types: (i) Descriptions of Images-in-Isolation (DII), (ii) Descriptions of Images-in-Sequence (DIS), and (iii) Stories for Images-in-Sequence (SIS). The stories are collected via crowdsourcing using AMT.  Similar to other image storytelling datasets, this dataset is split into 80\%, 10\%, and 10\% for training, validation, and testing purposes respectively. Table~\ref{table:sind-storytelling-dataset} presents the statistics of the dataset.

 \begin{table}[!ht]
\small
    \centering
    \begin{tabular}{l c c c c}
    \hline  %\toprule
    \rowcolor{teal!35}
                  & Images     & Flickr Albums       & (Text, Image)     & Vocab   \\
    \hline \addlinespace[0.3em] %\midrule
        DII           & -          & -              & 151,800           & 13,800    \\
        DIS           & -          & -              & 151,800           & 5,000     \\
        SIS           & -          & -              & 252,900           & 18,200    \\
    \hline
        Total         & 210,819    & 10,117         &  -                &  -       \\
       \bottomrule
    \end{tabular}
    \caption{\label{table:sind-storytelling-dataset} Statistics of the SIND dataset.}
\end{table}

\paragraph{Visual Storytelling Dataset (VIST).}\label{para:vistdata} VIST\footnote{\url{http://visionandlanguage.net/VIST} \label{fnote: vist-dataset-url}} is the second version (v.2) of SIND (see Section~\ref{para:sinddata}) and is aimed at modeling the social language of humans for evolving AI to be more human-like in understanding.  Basic statistics of the dataset are shown in Table~\ref{table:vist-storytelling-dataset} while the splits of it can be found in Table~\ref{table:vist-storytelling-dataset-split}. 

\begin{table}[!ht]
\small
    \centering
    \begin{tabular}{l c}
    \hline  %\toprule
    \rowcolor{teal!35}
         Images     & Text Sequences \\
        \hline \addlinespace[0.3em] %\midrule
        81,743     & 10,117  \\
       \bottomrule
    \end{tabular}
    \caption{\label{table:vist-storytelling-dataset} Statistics of the VIST (SIND v.2) dataset.}
\end{table}

\begin{table}[!ht]
\small
    \centering
    \begin{tabular}{l | c c}
    \hline  %\toprule
    \rowcolor{teal!35}
        Split       & Stories    & Sentences \\
    \hline \addlinespace[0.3em]    %\midrule
        Training     & 40,155   & 200,775 \\
        Validation   & 4,990     & 24,950   \\
        Test         & 5,055     & 25,275    \\
    \bottomrule
    \end{tabular}
    \caption{\label{table:vist-storytelling-dataset-split} Splits of the VIST dataset.}
\end{table}

\subsubsection{Image Storytelling - Evaluation Measures, Models, and Results}
\label{sssec:imagestorytellall}

In this section, we review the measures used to evaluate different \textit{Image Storytelling} models and the results obtained by them.

\paragraph{Evaluation Measures.} To evaluate \textit{Image Storytelling} models, the Language metrics and Retrieval metrics presented in Section~\ref{sssec:idgall} are used.

\paragraph{Models.} Many models have been created in attempts to solve the \textit{Image Storytelling} task. In Table~\ref{arcimagestory}, we present some exemplar architectures (refer to \textit{Combined} column) created to address the task by integrating both image and language inputs. We also include a column that showcases the optimization techniques used to train those models.

\begin{table*}[!ht]
\small
  \centering
  \begin{tabular}{lccccc}
    \hline  %\toprule
    \rowcolor{teal!35}
    {Approach} & Image & Language & Combined & Optimizer & RL\\
    \hline \addlinespace[0.3em]     %\midrule
    ~\shortcite{kiros:2014} & AlexNet & LM & MLBL & - & \xmark\\
    ~\shortcite{karpathy:2015} &  VGG & RNN & NeuralTalk & RMSprop & \xmark \\
    ~\shortcite{vinyals:2015} & GoogLeNet & LSTM & NIC  & SGD & \xmark\\ 
    ~\shortcite{park:2015} & VGG & RNN & CRCN & RMSprop & \xmark\\
    ~\shortcite{huang:2016} & VGG & GRU & Story-Flat & - & \xmark\\
    ~\shortcite{krause:2017}& VGG & LSTM & HierarchicalRNN & ADAM & \xmark\\
    ~\shortcite{liupt:2017} & VGG & LSTM  & BARNN & - & \xmark \\
    ~\shortcite{wangsrt:2018} & VGG & LSTM & GAN & ADAM & \cmark \\
    ~\shortcite{wangno:2018} & ResNet-152 & GRU & AREL & ADAM & \cmark \\
    \bottomrule
  \end{tabular}
  \caption{\label{arcimagestory} Exemplar \textit{Image Storytelling} architectures.}
\end{table*}

\paragraph{Results.} In Table~\ref{resimsnyc}, Table~\ref{resimdisney}, Table~\ref{resimsind}, and Table~\ref{resimvist} we present the results obtained with a subset of models which use the datasets presented earlier in Section~\ref{sssec:imstorydata}.

\begin{table*}[!ht]
\small
  \centering
  \begin{tabular}{lccccccc}
    \hline    %\toprule
    \rowcolor{teal!35}
    {Model} & B-4 & CIDEr & METEOR & R@1 & R@5 & MedRank \\
    \hline \addlinespace[0.3em]     %\midrule
    MLBL~\shortcite{kiros:2014} & 0.01 & 2.6 & 5.29 & 1.19 & 4.52 & 100.5 \\
    NeuralTalk~\shortcite{karpathy:2015} & 0.00 & 0.5 & 1.34 & 0.48 & 2.86 & 120.5 \\
    NIC~\shortcite{vinyals:2015} & 0.10 & 9.1 & 5.73 & 0.95 & 7.38 & 88.5 \\ 
    CRCN~\shortcite{park:2015} & \textbf{2.08} & 30.9 & 7.69 & 11.67 & 31.19 & 14.00 \\
    Story-Flat~\shortcite{huang:2016} & - & - & 7.37 & - & - & - \\
    HierarchialRNN~\shortcite{krause:2017}& - & - & 6.07 & - & - & - \\
    BARNN~\shortcite{liupt:2017} & - & \textbf{41.6}  & - & \textbf{29.37} & \textbf{45.43} & \textbf{8} \\
    AREL~\shortcite{wangsrt:2018} & - & - & \textbf{8.39} & - & - & - \\
    \bottomrule
  \end{tabular}
  \caption{\label{resimsnyc}Results of different models on the NYC-Storytelling dataset.}
\end{table*}

\begin{table*}[!ht]
\small
  \centering
  \begin{tabular}{lccccccc}
    \hline  %\toprule
    \rowcolor{teal!35}
    {Model} & B-4 & CIDEr & METEOR & R@1 & R@5 & MedRank \\
    \hline \addlinespace[0.3em]     %\midrule
    MLBL~\shortcite{kiros:2014} & 0.01 & 3.4 & 4.99 & 1.02 & 4.08 & 62 \\
    NeuralTalk~\shortcite{karpathy:2015} & 0.00 & 0.4 & 1.34 & 1.02 & 3.40 & 88 \\
    NIC~\shortcite{vinyals:2015} & 0.07 & 10.0 & 4.51 & 2.83 & 10.38 & 61.5 \\ 
    CRCN~\shortcite{park:2015} & 3.49 & 52.7 & 8.78 & 14.29 & 31.29 & 16 \\
    Story-Flat~\shortcite{huang:2016} & - & - & 7.61 & - & - & - \\
    HierarchialRNN~\shortcite{krause:2017}& - & - & 7.72 & - & - & - \\
    BARNN~\shortcite{liupt:2017} & - & \textbf{54.1} & - & \textbf{35.01} & \textbf{49.07} & \textbf{6} \\
    AREL~\shortcite{wangsrt:2018} & - & - & \textbf{9.90} & - & - & - \\
    \bottomrule
  \end{tabular}
  \caption{\label{resimdisney} Results of various models on the Disneyland-Storytelling dataset.}
\end{table*}

\begin{table*}[!ht]
\small
  \centering
  \begin{tabular}{lccccccc}
    \hline  %\toprule
    \rowcolor{teal!35}
    {Model} & B-4 & CIDEr & METEOR & R@1 & R@5 & MedRank \\
    \hline \addlinespace[0.3em]     %\midrule
    CRCN~\shortcite{park:2015} & - & - & -  & 9.87 & 28.74 & 21 \\
    Story-Flat~\shortcite{huang:2016} & 3.50 & 6.84 & 10.25 & - & - & - \\
    HierarchialRNN~\shortcite{krause:2017}& 3.7 & 6.51 & 9.97 & - & - & - \\
    AREL~\shortcite{wangsrt:2018} & \textbf{5.16} & \textbf{11.35} & \textbf{12.32} & - & - & - \\
    \bottomrule
  \end{tabular}
  \caption{\label{resimsind} Results of different models on the SIND dataset.}
\end{table*}

\begin{table*}[!ht]
\small
  \centering
  \begin{tabular}{lccccccc}
    \hline  %\toprule
    \rowcolor{teal!35}
    {Model} & B-4 & CIDEr & METEOR & R@1 & R@5 & MedRank \\
    \hline \addlinespace[0.3em]     %\midrule
    enc-attn-dec~\shortcite{xu:2015} & - & 4.96 & 32.98 & - & - & - \\
    h-attn-rank~\shortcite{yualbum:2017} & - & 7.38 & 33.94 & - & - & - \\
    BARNN~\shortcite{liupt:2017} & - & - & 33.32 & \textbf{24.07} & \textbf{44.29} & \textbf{9} \\
    AREL-t-100~\shortcite{wangno:2018} & 14.1 & \textbf{9.4} & \textbf{35.0} & - & - & - \\
    \bottomrule
  \end{tabular}
  \caption{\label{resimvist} Results of various models on the VIST dataset.}
\end{table*}

\subsubsection{Image Storytelling - Discussion}
\label{sssec:imagestoryopenques}
We observe that for \textit{Image Storytelling}, the adversarial approach, i.e., Adversarial REward Learning (AREL) proposed by \shortciteA{wangno:2018}, achieves best results on both retrieval and language metrics for different datasets. This attests to AREL's ability to clone expert behaviors while still generating more human-like stories.

\subsubsection{Video Storytelling - Introduction}
\label{sssec:vistintro}
\begin{figure}[!htb]
    \centering
        \includegraphics[width=\textwidth]{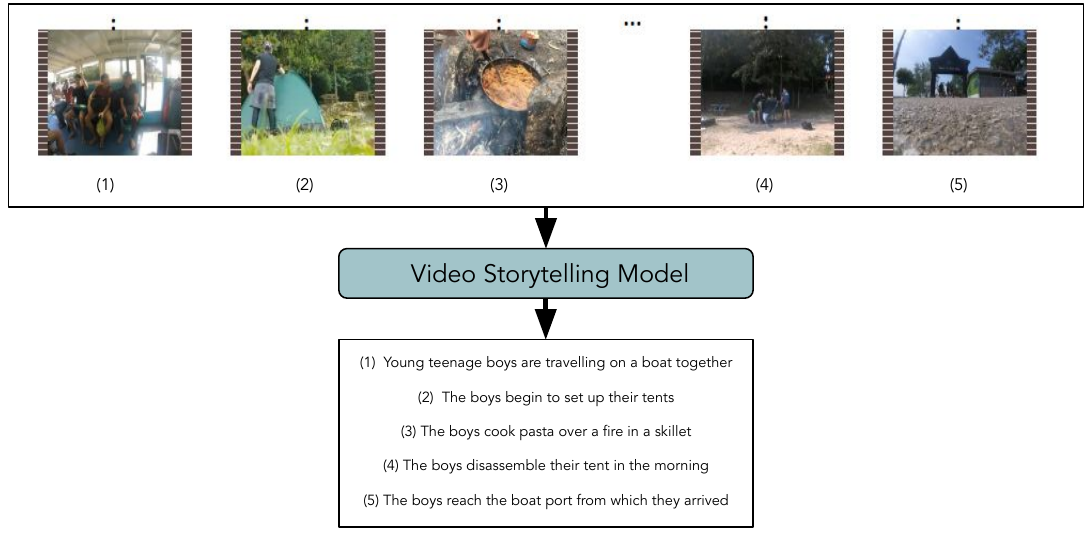}
    \caption{Given \textit{video frames} (adopted from ~\shortcite{liv:2018}) as input, a Video Storytelling Model generates a textual story in sequence.}\label{fig:videostory}
\end{figure}

In comparison to image storytelling, which only deals with a small sequence of images, the aim of video storytelling is to generate coherent and succinct stories for long videos. However, video storytelling is less explored. The video storytelling task was pioneered by~\shortciteA{liv:2018} to address challenges such as diversity in the story and the inherent complexity of video. They introduced residual Bidirectional RNNs (BiRNNs) for leveraging context and a narrator model with reinforcement learning. Further,~\shortciteA{gelladata:2018} created a multi-sentence video description dataset (VideoStory) to resemble stories from social media videos. The goal of social media-specific video description generation was to offer support to people with visual disabilities or other technical issues such as internet bandwidth limitations. Figure~\ref{fig:videostory} illustrates the task of video storytelling where a story in a sequence is generated based on a video as the sole input.

It is worth noting that this task bears close resemblance to the well-researched area of video summarization using only videos~\shortcite{ma:2002}.

\subsubsection{Video Storytelling - Datasets}
\label{sssec:videostorydata}
Similar to \textit{image storytelling} datasets, currently two different datasets are available to address the task of video storytelling. In the following, we elaborate on these two datasets.

\paragraph{VideoStory.}\label{para:videostorydataset} VideoStory \shortcite{gelladata:2018} is a multi-sentence description dataset created from social media videos that are selected to be highly diverse and engaging. Table~\ref{table:videostory-dataset} shows more statistics on the dataset.

\begin{table}[!htb]
\small
    \centering
    \begin{tabular}{l c c c c c}
    \hline  %\toprule
    \rowcolor{teal!35}
         Total      & Total         &Total      & Avg. Video     & Total         &Sentences\\
    \rowcolor{teal!35}
        Videos      & Length        &Clips      & Duration      & Sentences     & per Video \\
    \hline \addlinespace[0.3em]    %\midrule
        20,000      & 396 h         &123,000    & 70s           & 123,000       & 4.67 \\
       \bottomrule
    \end{tabular}
    \caption{\label{table:videostory-dataset} Statistics of the VideoStory dataset.}
\end{table}

Models can be evaluated locally on the earmarked \textit{test} set whereas \textit{test (blind)} is reserved for online evaluation purposes. However, the dataset including annotations has not been made public yet. Table~\ref{table:videostory-dataset-split} presents actual number of videos, clips, and sentence annotations for each of the splits. 

\begin{table}[!ht]
\small
    \centering
    \begin{tabular}{l | c c c c c}
    \hline  %\toprule
    \rowcolor{teal!35}
        Split           & Videos     & Clips    &Paragraphs/video  & Paragraphs    & Words/paragraph   \\
    \hline \addlinespace[0.3em]    %\midrule
        Training        & 17,098     & 80,598   &1                      & 17,098       & 61.76 \\
        Validation      & 999        & 13,796   &3                      & 2,997        & 59.88  \\
        Testing         & 1,011      & 14,093   &3                      & 3,033        & 59.77  \\
        Test (Blind)    & 1,039      & 14,139   &3                      & 3,117        & 69.45  \\
    \hline \addlinespace[0.3em]    %\midrule
        Total           & 20,147     & 122,626  &-                      & 26,245        & 62.23 \\
       \bottomrule
    \end{tabular}
    \caption{\label{table:videostory-dataset-split} Splits of the VideoStory dataset.}
\end{table}

\paragraph{VideoStory-NUS.}The \label{para:videostorynusdataset}VideoStory-NUS\footnote{\url{https://zenodo.org/record/2383739} \label{fnote: videostory-NUS-dataset-url}} \shortcite{liv:2018} dataset contains social event videos that were collected from YouTube by querying for common and complex events, namely \textit{Birthday}, \textit{Camping}, \textit{Christmas}, and \textit{Wedding}. Specifically, it comes with 105 manually chosen videos with sufficient inter-event and intra-event variations which are annotated with descriptive stories obtained through AMT.  Each video is annotated by at least 5 different AMT workers, thus resulting in 529 stories in total. More statistics of the dataset can be found in Table~\ref{table:videostory-nus-dataset}.
\begin{table}[!ht]
\small
    \centering
    \begin{tabular}{l c c c c c}
    \hline  %\toprule
        \rowcolor{teal!35}
                                    &                           & Avg. Video     &  Avg. Story     & Avg. Sentence      & Vocab.\\
        \rowcolor{teal!35}
        \multirow{-2}{*}{Domain}    & \multirow{-2}{*}{Videos}  & Length         & Length          & Length             & Size   \\
    \hline \addlinespace[0.3em]    %\midrule
        Open                        &105                        & 12 m 35 s      &162.6            & 12.1               & 4,045 \\
      \bottomrule
    \end{tabular}
    \caption{\label{table:videostory-nus-dataset} Statistics of the VideoStory-NUS dataset.}
\end{table}

 For experimental purposes, the dataset is randomly split in a ratio of 14:3:3 for training, validation, and testing respectively. Actual numbers are presented in Table~\ref{table:videostory-nus-dataset-split}. 

\begin{table}[!ht]
\small
    \centering
    \begin{tabular}{l | c c}
    \hline  %\toprule
    \rowcolor{teal!35}
        Split           & Percentage (\%)   & Videos  \\
    \hline \addlinespace[0.3em]    %\midrule
        Training        & 70                & 73    \\
        Validation      & 15                & 16    \\
        Test            & 15                & 16    \\
      \bottomrule
    \end{tabular}
    \caption{\label{table:videostory-nus-dataset-split} Splits of the VideoStory-NUS dataset.}
\end{table}

\subsubsection{Video Storytelling - Evaluation Measures, Models, and Results}
\label{sssec:videostorytellall}

In this section, we review the measures used to evaluate different \textit{Video Storytelling} models and the results obtained by them.

\paragraph{Evaluation Measures.} To evaluate \textit{Video Storytelling} models, the Language metrics and Retrieval metrics presented in Section~\ref{sssec:vdgall} are used.

\paragraph{Models.} There are a number of different models available for the task of \textit{Video Storytelling}. These models combine representations of video and language in an efficient manner to address the task. In Table~\ref{arcvideostory}, we present some exemplar architectures (refer to \textit{Combined} column) created to accomplish the task by integrating both video and language inputs. To understand the optimization techniques used, we also include a column that showcases the optimization method used to train the models.

\begin{table*}[!ht]
\small
  \centering
  \begin{tabular}{lcccccc}
    \hline  %\toprule
    \rowcolor{teal!35}
    {Approach} & Video & Frame & Language & Combined & Optimizer & RL\\
    \hline \addlinespace[0.3em]     %\midrule
    ~\shortcite{yuvideo:2016} & C3D & VGG & GRU & H-RNN & RMSProp & \xmark \\
    ~\shortcite{gelladata:2018} & R3D & ResNet-101 & GRU & seq-seq+context & ADAM & \xmark \\
    ~\shortcite{liv:2018} & - & ResNet-101 & GRU & ResBRNN & ADAM & \cmark \\
    \bottomrule
  \end{tabular}
  \caption{\label{arcvideostory} Exemplar \textit{Video Storytelling} architectures.}
\end{table*}

\paragraph{Results.} The \textit{Video Storytelling} results showcases the efficacy of the proposed models. In Table~\ref{resvidstory} and Table~\ref{resvideostorynus} we present results obtained with a subset of models built using the datasets presented earlier in Section~\ref{sssec:videostorydata}.

\begin{table}[!ht]
\small
  \centering
  \begin{tabular}{lccccccc}
    \hline  %\toprule
    \rowcolor{teal!35}
    {Model} & B-4 & CIDEr & METEOR & R@1 & R@5 & MedRank \\
    \hline \addlinespace[0.3em]     %\midrule
    seq-seq+context~\shortcite{gelladata:2018} & 1.20 & 9.37 & 33.88 & - & - & - \\
    \bottomrule
  \end{tabular}
  \caption{\label{resvidstory} Results obtained with different models on the VideoStory dataset.}
\end{table}
\begin{table*}[!ht]
\small
  \centering
  \begin{tabular}{lccccccc}
    \hline  %\toprule
    \rowcolor{teal!35}
    {Model} & B-4 & CIDEr & METEOR & R@1 & R@5 & MedRank \\
    \hline \addlinespace[0.3em]     %\midrule
    mRNN~\shortcite{mao:2014} & 11.8 & 81.3 & 18.0 & 5.34 & 21.23 & 29 \\
    Deep Video-Text~\shortcite{xujoint:2015} & 11.5 & 79.5 & 17.7 & 4.72 & 19.85 & 31 \\
    H-RNN~\shortcite{yuvideo:2016} & \textbf{16.1} & 64.6 & 15.5 & - & - & - \\
    ResBRNN~\shortcite{liv:2018} & 14.7 & 94.3 & 19.6 & \textbf{7.44} & \textbf{25.77} & \textbf{22} \\
    ResBRNN-kNN~\shortcite{liv:2018} & 15.6 & \textbf{103.6} & \textbf{20.1} & - & - & - \\
    \bottomrule
  \end{tabular}
  \caption{\label{resvideostorynus} Results obtained with different models on the VideoStory-NUS dataset.}
\end{table*}

\subsubsection{Video Storytelling - Discussion}
\label{sssec:videostoryopenques}
For \textit{Video Storytelling}, a different set of methods are used for comparing two datasets. In Table~\ref{resvidstory}, we observe that only one method utilizing the sequence-to-sequence paradigm with contextual information (i.e., seq2seq+context) is evaluated on the ``VideoStory'' dataset. Nevertheless, another set of methods used for comparison for the ``VideoStory-NUS'' dataset is in Table~\ref{resvideostorynus}. It shows that the approach proposed by~\shortciteA{liv:2018} using Residual BRNN with k-Nearest Neighbours (i.e., ResBRNN-kNN) outperforms most of the baseline methods.

\section{Visual Referring Expression Comprehension and Generation}
\label{sec:vreandpg}
In this section, we explore the task of \textit{Visual Referring Expression Comprehension and Generation}. The objective of the task is to ground a natural language expression (e.g. a noun phrase or a longer piece of text) to objects in a visual input. 
\subsection{Image Referring Expression Comprehension and Generation}
\label{sssec:iretask}
In the following, we provide a detailed description of the \textit{Visual Referring Expression Comprehension and Generation} by using an image as the visual input.
\subsubsection{Image Referring Expression Comprehension and Generation - Intro}
\label{sssec:iretaskintro}
In a natural environment, people use referring expressions to unambiguously identify, indicate, or point to particular objects. This is usually done with a simple phrase or within a larger context (e.g. a sentence). Having a larger context provides better scope for avoiding ambiguity and allows the referential expression to easily map to the target object. However, there can also be other possibilities in which people are asked to describe a target object based on its surrounding objects.

\begin{figure}[!htb]
    \centering
        \includegraphics[width=\textwidth]{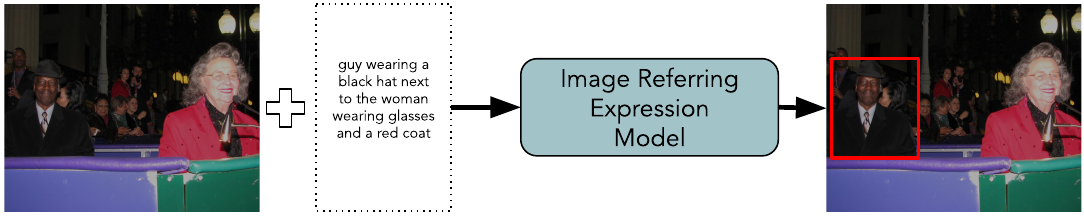}
    \caption{Given an \textit{image} and a \textit{referring expression}, an Image Referring Expression Comprehension Model identifies it in the image using bounding boxes.}\label{fig:imagerefer}
\end{figure}

This provides us with two different possibilities for the visual referring expression task. In the first scenario, referring expressions deal with \textit{generation}, in which an algorithm generates a referring expression for a given target object that is present in a visual scene. In the second scenario, the referring expression is used to perform comprehension, in which an algorithm locates in an image the object described by a given referring expression. Figure~\ref{fig:imagerefer} shows an example for the task of referring expression comprehension. 

Given these tasks, different approaches have been proposed for referring expression generation~\shortcite{golland:2010,mitchell:2013}, comprehension~\shortcite{kazemzadeh:2014}, and both combined~\shortcite{maorefer:2016,yumodel:2016}. Note that there is a difference between referring expression tasks and grounding of free-form textual phrases~\shortcite{rohrbachground:2016} in an image. 

\paragraph{Image Referring Expression Generation.} An initial approach~\shortcite{fitzgerald:2013} tackled the problem from the perspective of density estimation, in which the goal was to learn distributions over logical expressions identifying sets of objects in the world. Other research designed a comprehension-guided referring expression generator~\shortcite{luo:2017} by using a comprehension module trained on human-generated expressions to generate referring expressions.

\paragraph{Image Referring Expression Comprehension.} ~\shortciteA{nagaraja:2016} investigated referring expression comprehension to integrate contexts between objects. Later on, techniques such as Multiple Instance Learning (MIL) were used to explore context regions and max-margin based MIL objective functions for training. Further, ~\shortciteA{huobjectret:2016} leveraged a natural language query of the object to localize a target object using a Spatial Context Recurrent Convnet (SCRC) model. It operates as a scoring function on candidate boxes for object retrieval, integrating spatial configurations and global scene-level contextual information. This explicit modeling of the referent and context region pairs has proven useful. Approaches such as compositional modular networks~\shortcite{hucmn:2017} analyzed referential expressions by identifying entities and relationships mentioned in the input expression and grounding them all in the scene. Such an approach has been shown to effectively inspect local regions and pairwise interactions between them. A modular approach was also explored where three modular components related to subject appearance, location, and relationship to other objects was used to model with Modular Attention Network~\shortcite{yumatnet:2018}. It has proven effective at focusing on the subjects and their relationships. Approaches such has GroundNet~\shortcite{cirik:2018} have leveraged syntactic analysis of the input referring expression to build a dynamic computation graph of neural modules that definesan  architecture for performing localization. Variational models have also been used for referential expression comprehension where variational Bayesian methods called \textit{variational context}~\shortcite{zhangref:2018} were used to solve the problem of complex context modeling. These methods have proven capable of exploiting the relation between the referent and context, thereby reducing the search space of context.  Furthermore, an accumulated attention mechanism~\shortcite{dengacc:2018} has been proposed to accumulate the attention for useful information in image, query, and objects. It has demonstrated the ability to reduce the redundancy and noise issues that were in other approaches.

Recently, a Cross-Modal Relationship Extractor (CMRE) and a Gated Graph Convolutional Network (GGCN) were combined into a cross-modal relationship inference network~\shortcite{yangcmin:2019}. CMRE has been shown to highlight objects and relationships which have connections with a given referring expression, while GGCN computes multimodal semantic contexts by fusing information from different modes and propagating multimodal information through the structured relation graph. Coming from a perspective of natural language understanding, a Recursive Grounding Tree~\shortcite{hong:2019} sought to automatically compose a binary tree structure by parsing the referring expression, in order to perform visual reasoning along the tree in a bottom-up fashion. It has been shown to allow gradients from continuous score functions with a discrete tree construction. There has also been interest in combining visual reasoning with referential expressions through the creation of new dataset~\shortcite{liuclevr:2019}. Most of the above approaches use bounding box localization, but additionally object segmentation~\shortcite{liusegment:2017} has also been explored for referring expression comprehension.

\paragraph{Image Referring Expression Generation and Comprehension.} Few approaches have performed both generation and comprehension tasks. Visual context~\shortcite{maorefer:2016,yumodel:2016} was initially used in referring expression models to find visual comparison to other objects within an image. It has shown significant improvements. Further, a unified framework~\shortcite{yuspeaker:2017} was designed using a speaker, a listener, and a reinforcer.  The \textit{speaker} generates referring expressions, the \textit{listener} comprehends referring expressions, and the \textit{reinforcer} introduces a reward function to guide sampling of more discriminative expressions. Feedback from the discriminative reinforcer has proven capable of benefiting the tasks. The role of attributes~\shortcite{liuattr:2017} was also studied to show that they help in disambiguation when referring to a particular object.

\subsubsection{Image Referring Expression Comprehension and Generation - Datasets}
\label{sssec:imreferdata}

For the task of image referring expression, both \textit{real} and \textit{synthetic} image datasets have been designed. In the following, we present the details of the datasets in separate sections.

\paragraph{Real Images.}\label{para:realref}In the real and natural images category, the ImageCLEF\footnote{\url{https://www.imageclef.org/SIAPRdata} \label{fnote: imageclef-iapr-dataset-url}} and MSCOCO$^{\ref{fnote: mscoco-dataset-url}}$ (see Section \ref{para:mscoco-data})  datasets are commonly used for creating referring expression annotations. From a subset of ImageCLEF's IAPR dataset\footref{fnote: imageclef-iapr-dataset-url}, referring expressions are collected in a game-based setting, namely ReferItGame\footref{fnote: referitgame-url} \shortcite{kazemzadeh:2014}. The resulting dataset is called as RefCLEF\footref{fnote: refcoco-dataset-github-url} and its statistics can be found in Table~\ref{table:referit-dataset}.

\begin{table}[!ht]
\small
    \centering
    \begin{tabular}{l c c c}
    \hline  %\toprule
    \rowcolor{teal!35}
        Real        & Distinct    & Referring       & Train/Test \\
    \rowcolor{teal!35}
        Images      & Objects     & Expressions     & Splits    \\
    \hline \addlinespace[0.3em]    %\midrule
        19,894      &96,654     &130,525            & Per-Image split \\
    \bottomrule
    \end{tabular}
    \caption{\label{table:referit-dataset} Statistics of the RefCLEF dataset.}  %\colorbox{brown!45}{RefClef}
\end{table}

The RefCOCO\footnote{\url{https://github.com/lichengunc/refer} \label{fnote: refcoco-dataset-github-url} }, RefCOCO+\footref{fnote: refcoco-dataset-github-url} \shortcite{yumodel:2016}, and RefCOCOg~\shortcite{maorefer:2016} datasets were all created using MSCOCO images. For RefCOCO and RefCOCO+, the ``People vs. Object'' split evaluates images containing multiple people (Test A) and images containing multiple instances of all other objects (Test B). Both RefCOCO and RefCOCO+ were collected in the same interactive setting as above, ReferItGame\footnote{\url{http://tamaraberg.com/referitgame} \label{fnote: referitgame-url}} \shortcite{kazemzadeh:2014}. Table~\ref{table:refcoco-dataset} presents the statistics of the RefCOCO dataset whereas Table~\ref{table:refcoco+-dataset} shows the statistics of the RefCOCO+ dataset.

\begin{table}[!ht]
\small
    \centering
    \begin{tabular}{lccc}
    \hline  %\toprule
    \rowcolor{teal!35}
                                 & Total       & Referring       & Train/Test \\
    \rowcolor{teal!35}
    \multirow{-2}{*}{Images}     & Objects     & Expressions     & Splits \\
    \hline \addlinespace[0.3em]    %\midrule
                    19,994       & 50,000      & 142,209         & People vs. Object\\
    \bottomrule
    \end{tabular}
    \caption{\label{table:refcoco-dataset} Statistics of the RefCOCO dataset.}  %\colorbox{brown!55}{RefCOCO}
\end{table}

One important distinction between the RefCOCO and RefCOCO+ datasets is that the latter was collected in a comparatively restrictive setting when compared to the former. Specifically, the usage of location words was not permitted in the referring expressions in case of RefCOCO+ whereas there was no such restriction on the language for RefCOCO.

\begin{table}[!ht]
\small
    \centering
    \begin{tabular}{l c c c c}
    \hline  %\toprule
    \rowcolor{teal!35}
                                    & Total       & Referring       & Train/Test \\
    \rowcolor{teal!35}
    \multirow{-2}{*}{Images}        & Objects     & Expressions     & Splits \\
    \hline \addlinespace[0.3em]    %\midrule
                    19,992          & 49,856      & 141,564         & People vs. Object \\
    \bottomrule
    \end{tabular}
    \caption{\label{table:refcoco+-dataset} Statistics of the RefCOCO+ dataset.} %\colorbox{brown!55}{RefCOCO+}
\end{table}

To overcome some of the limitations of RefCLEF, a dataset based on based on MSCOCO\footref{fnote: mscoco-dataset-url} was created. This dataset, known as RefCOCOg\footnote{\url{https://github.com/mjhucla/Google_Refexp_toolbox} \label{fnote: refcocog-dataset-url}} \shortcite{maorefer:2016}, contains much longer sentences and was collected in a non-interactive setting using AMT, in contrast to the interactive setting used with RefCLEF, RefCOCO, and RefCOCO+. The statistics of this dataset is presented in Table~\ref{table:refcocog-dataset}.

\begin{table}[!ht]
\small
    \centering
    \begin{tabular}{l c c c}
    \hline  %\toprule
    \rowcolor{teal!35}
                                    & Total       & Referring       & Train/Test \\
    \rowcolor{teal!35}
    \multirow{-2}{*}{Images}        & Objects     & Expressions     & Splits \\
    \hline \addlinespace[0.3em]    %\midrule
                    26,711          & 54,822      & 85,474          & Per-Object \\
    \bottomrule
    \end{tabular}
    \caption{\label{table:refcocog-dataset} Statistics of the RefCOCOg dataset.}  %\colorbox{brown!55}{RefCOCOg}
\end{table}

Earlier mentioned referring expression datasets use single sentences for image referring expression. In contrast, the GuessWhat\footnote{\url{https://github.com/GuessWhatGame/guesswhat} \label{fnote: guesswhat-dataset-url}}~\shortcite{vries:2017} dataset was created with a cooperative two-player guessing game, the goal of which was to locate an unknown object in an image (collected from MSCOCO) by asking a sequence of questions. Hence, it creates multiple sentences (i.e., a dialog) for a given image in order to perform referring expression. Another notable aspect of this dataset is that only images containing a number of objects in the range of 3 to 20 are chosen from MSCOCO. The dialogue collection was achieved via crowdsourcing using AMT. For evaluation, the dataset is randomly split into 70\% for training, 15\% for validation, and 15\% for testing. Table~\ref{table: guesswhat-dataset-stats} presents more details about the dataset.

\begin{table}[!ht]
\small
    \centering
    \begin{tabular}{l | c c c c c c}
    \hline  %\toprule
    \rowcolor{teal!35}
        Dataset Type    & Images      & Objects     & Dialogues     & Questions     & Words         & Vocab. Size   \\
    \hline \addlinespace[0.3em]    %\midrule
        Full            & 66,537      & 134,073     & 155,280       & 821,889       & 3,986,192     & 11,465    \\
        Finished        & 65,112      & 125,349     & 144,434       & 732,081       & 3,540,497     & 10,985     \\
        Success         & 62,954      & 114,271     & 131,394       & 648,493       & 3,125,219     & 10,469      \\
    \bottomrule
    \end{tabular}
    \caption{ \label{table: guesswhat-dataset-stats} Statistics of ``GuessWhat'' dataset. The row `Full' means all the dialogues are included, `Finished' means all finished dialogues (successful and unsuccessful) are included, and `Success' means only successful dialogues are included.}
\end{table}

\paragraph{Synthetic Images.}\label{para:synref}In the synthetic category, the CLEVR-Ref+\footnote{\url{https://cs.jhu.edu/~cxliu/2019/clevr-ref+.html} \label{fnote: clevr-ref+-dataset-url}}~\shortcite{liuclevr:2019} dataset was introduced to address issues such as bias in datasets with real images, since it has been recently been shown that referring expression models suffer from unintended biases \shortcite{cirikref+:2018}. CLEVR-Ref+ reuses the images from the CLEVR dataset (see Section \ref{para:clevrdata}), while replacing the questions in CLEVR with referring expressions and answers with referred objects. The main purpose of CLEVR-Ref+ is to diagnose image reasoning with referring expressions by exercising the desired control over the nature of samples. Table~\ref{table:clevr-ref+-dataset} present splits of the dataset.
\begin{table}[!ht]
\small
    \centering
    \begin{tabular}{l | c c}
    \hline  %\toprule
    \rowcolor{teal!35}
        Split           &Images             & Referring Expressions \\
    \hline \addlinespace[0.3em]    %\midrule
        Training        &70,000             & 700,000 \\
        Validation      &15,000	            & 150,000 \\
        Test  	        &15,000             & 150,000 \\
    \bottomrule
    \end{tabular}
    \caption{\label{table:clevr-ref+-dataset} Splits of the CLEVR-Ref+ dataset.}
\end{table}

\subsubsection{Image Referring Expression Comprehension and Generation - Evaluation Measures, Models, and Results}
\label{sssec:imagereferall}

In this section, we review the measures used to evaluate different \textit{Image Referring Expression} models and the results achieved by them.

\paragraph{Evaluation Measures.} The measure that is usually used for the evaluation of \textit{Image Referring Expression} models is Precision@1, i.e., precision calculated with the Intersection over Union (IoU) ratio between the true and predicted bounding box.

\paragraph{Models.} The models designed to approach the task of \textit{Image Referring Expression} provide an effective way to optimize the Precision@1 measure by identifying the right object in a visual input which matches the textual phrase. In Table~\ref{arcimagerefer}, we present some exemplar architectures (refer to \textit{Combined} column) created to address the task by integrating both image and language inputs. We also include a column that showcases the optimization techniques used to train those models.

\begin{table*}[!ht]
\small
  \centering
  \begin{tabular}{lccccc}
    \hline  %\toprule
    \rowcolor{teal!35}
    Approach & Image & Language & Combined & Optimizer & RL\\
    \hline \addlinespace[0.3em]     %\midrule
   ~\shortcite{maorefer:2016} & VGG & LSTM & MMI & SGD & \xmark \\
   ~\shortcite{nagaraja:2016} & VGG & LSTM & Neg. Bag & SGD & \xmark \\
   ~\shortcite{yumodel:2016}  & VGG & LSTM & Context & - & \xmark\\
   ~\shortcite{luo:2017}  & VGG & BiLSTM & CG & ADAM & \xmark \\
   ~\shortcite{liuattr:2017}  & VGG & LSTM & Combined & ADAM & \xmark\\
   ~\shortcite{hucmn:2017}  & VGG & LSTM & CMN & - & \xmark \\
   ~\shortcite{yuspeaker:2017}  & VGG & LSTM & Reinforcer & ADAM & \cmark \\
   ~\shortcite{zhangref:2018}  & VGG & BiLSTM & VarContext & SGD & \cmark \\
   ~\shortcite{dengacc:2018} & VGG & LSTM & AccumulateAtt & SGD & \xmark \\
   ~\shortcite{zhuangparallel:2018}  & VGG & LSTM & ParallelAtt & ADAM & \xmark\\
   ~\shortcite{yumatnet:2018}  & ResNet-101 & BiLSTM & MAttNet & - & \xmark \\
   ~\shortcite{hong:2019}  & ResNet-101 & BiLSTM & RVG-Tree & ADAM & \xmark \\
   ~\shortcite{yangcmin:2019}  & ResNet-101 & BiLSTM & CMRIN & ADAM & \xmark \\
    \bottomrule
  \end{tabular}
  \caption{\label{arcimagerefer} Exemplar \textit{Image Referring Expression and Comprehension} architectures.}
\end{table*}

\paragraph{Results.} Several models and datasets have been created to address the task of \textit{Image Referring Expression}. These datasets provide variety in the content so that they enhance the generalization ability of the models. In this section, we cover the results obtained by the models on some representative datasets. Table~\ref{resimagerefercoco} and Table~\ref{resimagerefercocoplus} presents results obtained with a subset of models built using the datasets such as RefCOCO, RefCOCO+, and RefCOCOg presented in Section~\ref{sssec:imreferdata}.

\begin{table*}
\begin{center}
\begin{tabular}{lccc}
\hline  %\toprule
\rowcolor{teal!35}
&\multicolumn{3}{c}{RefCOCO}  \\ %\cline{2-4}
%\hline     %\midrule
\rowcolor{teal!35}
\multirow{-2}{*}{Model} & val & testA & testB  \\ 
\hline \addlinespace[0.3em] %\midrule
MMI~\shortcite{maorefer:2016}  & - & 63.15 & 64.21  \\
Neg. Bag~\shortcite{nagaraja:2016}  & 76.90 & 75.60 & 78.00 \\
Context~\shortcite{yumodel:2016} & 76.18 & 74.39 & 77.30  \\
CG~\shortcite{luo:2017}  & - & 74.04 & 73.43  \\
Attributes~\shortcite{liuattr:2017}  & - & 78.85 & 78.07 \\
CMN~\shortcite{hucmn:2017}  & - & 75.94 & 79.57  \\
Reinforcer~\shortcite{yuspeaker:2017}  & 79.56 & 78.95 & 80.22   \\
VarContext~\shortcite{zhangref:2018}  & - & 78.98 & 82.39  \\
AccumulateAtt~\shortcite{dengacc:2018} & 81.27 & 81.17 & 80.01 \\
ParallelAtt~\shortcite{zhuangparallel:2018}  & 81.67 & 80.81 & 81.32  \\
MAttNet+ResNet-101~\shortcite{yumatnet:2018}  & 85.65 & 85.26 & 84.57   \\
RVG-Tree+ResNet-101~\shortcite{hong:2019}  & 83.48 & 82.52 & 82.90  \\ 
CMRIN+ResNet-101~\shortcite{yangcmin:2019}  & \textbf{86.99} & \textbf{87.63} & \textbf{84.73}  \\

\bottomrule
\end{tabular}
\end{center}
\caption{\label{resimagerefercoco} Comparison of Precision@1 (\%) scores of different methods on RefCOCO.}
\end{table*}

\begin{table*}
\setlength{\aboverulesep}{0pt}
\setlength{\belowrulesep}{0pt}
\begin{center}
\begin{tabular}{lccccc}
\hline  %\toprule
\rowcolor{teal!35}
      & \multicolumn{3}{c}{RefCOCO+}  & \multicolumn{2}{c}{RefCOCOg}  \\ %\cline{2-4} \cline{5-6}
%\hline     %\midrule
\rowcolor{teal!35}
\multirow{-2}{*}{Model}                & val & testA & testB & val & test \\
\hline \addlinespace[0.3em]     %\midrule
MMI~\shortcite{maorefer:2016}  & - & 48.73 & 42.13 & - & - \\
Neg Bag~\shortcite{nagaraja:2016}  & - & - & - & - & 68.40 \\
Context~\shortcite{yumodel:2016} & 58.94 & 61.29 & 56.24 & - & - \\
CG~\shortcite{luo:2017}  & - & 60.26 & 55.03 & - & - \\
Attributes~\shortcite{liuattr:2017}  & - & 61.47 & 57.22 & - & - \\
CMN~\shortcite{hucmn:2017}  & - & 59.29 & 59.34 & - & - \\
Reinforcer~\shortcite{yuspeaker:2017}  & 62.26 & 64.60 & 59.62 & 71.65 & 71.92  \\
VariationalContext~\shortcite{zhangref:2018}  & - & 62.56 & 62.90 & - & -  \\
AccumulateAttn~\shortcite{dengacc:2018} & 65.56 & 68.76 & 60.63 & - & -  \\
ParallelAttn~\shortcite{zhuangparallel:2018}  & 64.18 & 66.31 & 61.46 & - & -  \\
MAttNet+ResNet-101~\shortcite{yumatnet:2018} & 71.01 & 75.13 & 66.17 & 78.10 & 78.12  \\
RVG-Tree+ResNet-101~\shortcite{hong:2019} & 68.86 & 70.21 & 65.49 & 76.82 & 75.20  \\
CMRIN+ResNet-101~\shortcite{yangcmin:2019}  & \textbf{75.52} & \textbf{80.93} & \textbf{68.99} & \textbf{80.45} & \textbf{80.66}  \\ 
\bottomrule
\end{tabular}
\end{center}
\caption{\label{resimagerefercocoplus} Comparison of Precision@1 (\%) scores of different methods on the RefCOCO+ and RefCOCOg datasets.}
\end{table*}

\subsubsection{Image Referring Expression Comprehension and Generation - Discussion}
\label{sssec:imagereferopenques}
For \textit{Image Referring Expression}, on all MSCOCO based datasets (i.e., RefCOCO, RefCOCO+, and RefCOCOg) the technique proposed by~\shortciteA{yangcmin:2019} outperforms existing baselines. This approach builds a Cross-Modal Relationship Extractor (CMRE) to highlight objects and their relationships. Furthermore, a Gated Graph Convolutional Network (GGCN) is used to compute multimodal semantic contexts by fusing information from different modes and propagating multimodal information. This Cross-Modal Relationship Inference Network (CMRIN) along with ResNet-101 visual features have been shown to achieve the best results.

\subsection{Video Referring Expression Comprehension and Generation}
\label{ssec:videoretask}
In the following, we describe the setting of \textit{Visual Referring Expression Comprehension and Generation} task when a video is used as the visual input.

\subsubsection{Video Referring Expression Comprehension and Generation - Intro}
\label{sssec:videoretaskintro}
\begin{figure}[!htb]
    \centering
        \includegraphics[width=\textwidth]{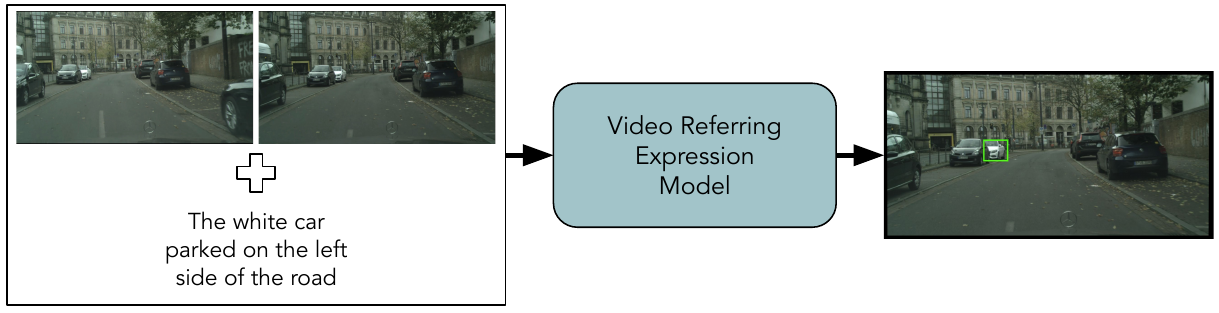}
    \caption{Given a \textit{video} (represented as a sequence of frames from ~\shortciteA{balajee:2018}) and a \textit{referring expression}, a Referring Expression Comprehension Model identifies it in the video using bounding boxes.}\label{fig:videorefer}
\end{figure}

When compared to image referring expression comprehension and generation, the task of video referring expression comprehension and generation is less explored at the time of publication of this survey. Although, there has been a surge in interest in tackling the spatio-temporal contexts and motion features that are inherent to videos, most of the works thus far, however, have concentrated on only one variant of image referring expression, namely comprehension.~\shortciteA{balajee:2018} used stereo videos to exploit richer and more realistic temporal-spatial contextual information along with gaze cues for referring expression comprehension. Figure~\ref{fig:videorefer} shows an example of the video referring expression comprehension. 
Another approach by~\shortciteA{khoreva:2018} explored Language Referring Expressions to point to the objects in the video to achieve object segmentation. Slightly different from the described task,~\shortciteA{wangtemporally:2020} proposed an end-to-end boundary-aware model for video grounding. The model uses a lightweight branch to predict semantic boundaries corresponding to the given linguistic information. It aggregates contextual information by explicitly modeling the relationship between the current element and its neighbours.

\subsubsection{Video Referring Expression Comprehension and Generation - Datasets}
\label{sssec:videoreferdata}

In this section, we present the datasets used to evaluate the task of \textit{Video Referring Expression Comprehension.}

\paragraph{Object Referring in videos with Gaze (ORGaze).} For performing Video Referring Expression, the Cityscapes\footnote{\url{https://www.cityscapes-dataset.com}} dataset containing a diverse set of stereo video sequences recorded in street scenes has been modified to have gaze information. Therefore, the ORGaze\footnote{\url{https://people.ee.ethz.ch/~arunv/ORGaze.html} \label{fnote: orgaze-dataset-url}}~\shortcite{balajee:2018} dataset contains object referring in videos with language and human gaze. More details of the dataset is presented in Table~\ref{table:orgaze-dataset}. 

\begin{table}[!ht]
\small
    \centering
    \begin{tabular}{l c c c c}
    \hline  %\toprule
    \rowcolor{teal!35}
        Videos  & Objects    & Condition  & Lighting   & Annotations            \\
    \hline \addlinespace[0.3em]    %\midrule
                &           &           &            &Bounding Boxes            \\
        5,000   &30,000     &Urban      &Daytime     &Gaze Recordings           \\
                &           &           &            &Language Expression       \\
    \bottomrule
    \end{tabular}
    \caption{ \label{table:orgaze-dataset} Statistics of the ORGaze dataset.}
\end{table}

The authors split the cities in the training set of Cityscapes for training and validation while using all the cities in validation set of Cityscapes for testing purposes. More concretely, the validation set is constructed by selecting one city (e.g., Z{\"u}rich) from Cityscapes training set while leaving the rest of the cities as part of the training set. For constructing the test set, the videos from all the cities in Cityscapes validation set (e.g., Frankfurt, Lindau, M{\"u}nster) of Cityscapes are used. Of the total 30,000 annotated objects, 80\% has been used for \textit{training} and the remaining 20\% was reserved for model evaluation of the task.

\subsubsection{Video Referring Expression Comprehension and Generation - Evaluation Measures, Models, and Results}
\label{sssec:videoreferall}

In this section, we review the evaluation measures used to benchmark different \textit{Video Referring Expression Comprehension} models and the results achieved by them.

\paragraph{Evaluation Measures.} The measure that is used for the evaluation of \textit{Video Referring Expression Comprehension} model is ``Top-1 Accuracy'' and also object proposal accuracy referred with Language-based Object Proposals (LOP), Faster R-CNN (FRCNN), and EdgeBox~\shortcite{zitnick:2014}.

\paragraph{Models.}  Many models have been created to solve the task of \textit{Video Referring Expression Comprehension}. In Table~\ref{arcvideorefer}, we present some exemplar architectures (refer to \textit{Combined} column) created to address the task by integrating both video and language. We also include a column that showcases the optimization techniques used to train those models. 

\begin{table*}[!ht]
\small
  \centering
  \begin{tabular}{lcccccc}
    \hline  %\toprule
    \rowcolor{teal!35}
    {Approach} & Video & Frame & Language & Combined & Optimizer & RL \\ 
    \hline \addlinespace[0.3em]     %\midrule
    ~\shortcite{balajee:2018} & - & VGG & LSTM & WithGaze & - & \xmark \\ 
    \bottomrule
  \end{tabular}
  \caption{\label{arcvideorefer} Exemplar \textit{Video Referring Expression and Comprehension} architectures.}
\end{table*}

\paragraph{Results.} As discussed earlier, several models have been created to approach the task of \textit{Video Referring Expression Comprehension}. In Table~\ref{resvideorefer} we present results obtained with a subset of models built using the ORGaze dataset presented earlier in Section~\ref{sssec:videoreferdata}.

\begin{table}[!ht]
\centering
\begin{tabular}{lccc}
\hline  %\toprule
\rowcolor{teal!35}
Methods &\multicolumn{1}{c}{Edgebox} & \multicolumn{1}{c}{FRCNN ($\uparrow$)} & \multicolumn{1}{c}{LOP ($\uparrow$)}\\
\hline \addlinespace[0.3em]     %\midrule
MNLM~\shortcite{kirosuni:2014}&-& 23.954 & 32.418 \\
VSEM~\shortcite{liuvid:2015}&-& 24.833 & 32.961 \\
MCB~\shortcite{fukui:2016}&-& 26.445& 33.366 \\
SimModel~\shortcite{plummer:2017}&4.5& 18.431& 35.556 \\
WithGaze~\shortcite{balajee:2018} &- & \textbf{47.256}& \textbf{47.012}  \\
\bottomrule
\end{tabular}
\caption{\label{resvideorefer} Comparison of Top-1 Accuracy (\%) of different methods on the ORGaze dataset.}
\end{table}

\subsubsection{Video Referring Expression Comprehension and Generation - Discussion}
\label{sssec:videoreferopenques}
The \textit{Video Referring Expression Comprehension} task is benchmarked using a single dataset. Evaluated using different task-specific metrics, the approach proposed by~\shortciteA{balajee:2018}, which uses the gaze information, produces the best results.

\section{Visual Question Answering, Reasoning, and Entailment}
\label{sec:vqaandrande}
In this section, we explore three different tasks, namely, \textit{Visual Question Answering}, \textit{Visual Reasoning}, and \textit{Visual Entailment}. The goal of each of these tasks are different. However, they share the common intention of answering questions when conditioned on a visual input. In the following sections, we elaborate on each of these three tasks separately.

\subsection{Visual Question Answering}
\label{ssec:vqa}
The goal of \textit{Visual Question Answering} (VQA) is to learn a model that comprehends visual content at both the global and local level for finding an association with pairs of questions and answers in the natural language form. The visual information for VQA includes both images and videos.

\subsubsection{Image Question Answering - Introduction}
\label{sssec:imqaintro}
\begin{figure}[!htb]
    \centering
        \includegraphics[width=\textwidth]{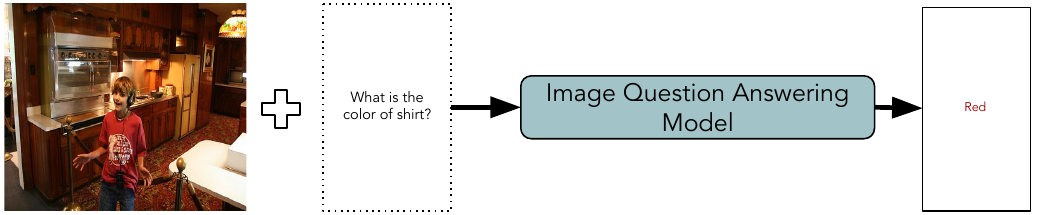}
    \caption{Given an \textit{image} and a \textit{question} about the image, an Image Question Answering model produces an answer to it.}\label{fig:vqa}
\end{figure}

The aim of \textit{Image Question Answering} (Image Q\&A) is to answer natural language questions about the contents of images. Earlier research efforts have focused on designing different algorithms and constructing datasets to address this challenge. The first approaches~\shortcite{malinowski:2014,malinowski:2015,geman:2015} considered Image Q\&A as a Visual Turing Test, where the expectation was to incorporate human-level abilities for semantically accessing the visual information to answer different questions. These were then improved as fill-in-the-blank tasks~\shortcite{yu:2015}, where the goal of the system was focused on multiple-choice question-answering for images. Also, it was expanded to address both multilingual~\shortcite{gao:2015} and automatic question generation, in which descriptions of sentences are converted into questions~\shortcite{ren:2015}. However, it lacked natural language questioning ability of humans. Hence, a broader task was proposed with an aim of addressing open-ended Image Q\&A~\shortcite{antol:2015,agrawal:2017}, where the challenge was to ask a free-form natural language question about an image and make the system to answer the question. Figure~\ref{fig:vqa} provides a schematic representation of the task where a free-form question about the contents of an image is asked to obtain an answer.

However, designing such a system can contain several other challenges, such as coming up with strong baselines~\shortcite{jabri:2016}. To address these, binary image Q\&A~\shortcite{zhang:2016} was explored by providing complementary images for abstract scenes. 
%The task is observed from the perspective of visual verification of concepts inquired in the questions.
These complementary images were used to provide visual verification of concepts contained in the questions.
Some of the questions were understood as a loose, global association between Q\&A sentences and images. Hence, more confined and dedicated tasks were created for relating local regions in the images~\shortcite{zhu:2016} by addressing object-level grounding. Some approaches~\shortcite{zhangcount:2018} concentrated only on counting objects in natural images. There are many methods that are proposed to address the challenging image Q\&A task. The details about different methods are already covered in earlier surveys~\shortcite{kafle:2017,wuvqa:2017}. Therefore, we briefly present new methods that were introduced after the publication of these surveys.

Recent works aim at interpretability or explainability by overcoming priors~\shortcite{agarwal:2018}, concentrating better on the image to extract relevant information~\shortcite{goyal:2019}, generating human-interpretable rules that provide better insights~\shortcite{manjunatha:2018}, and cycle-consistency~\shortcite{shah:2019}, while other works try to understand the text inside an image to answer and reason about it~\shortcite{singh:2019}. More recent works sought to incorporate outside knowledge~\shortcite{marino:2019} in the image Q\&A framework to support real-world knowledge-aware question answering~\shortcite{shahkvqa:2019}. 

There are different kinds of learning approaches used for image Q\&A, such as Multi-task learning and Federated learning. A multi-task learning approach~\shortcite{nguyenmulti:2019} is used to learn a vision-language representation that is shared by many tasks from their diverse datasets to address image Q\&A. In contrast, federated learning is used with the aimNet~\shortcite{liufederated:2020} and is validated on federated learning settings that include both horizontal and vertical federated learning. To focus on language priors, a modular language attention mechanism is used by~\shortciteA{jingovercoming:2020} to parse a question into three phrase representations, namely type representation, object representation, and concept representation. It has prevented language priors from dominating the answering process.

\subsubsection{Image Question Answering - Datasets}
\label{sssec:imqadata}
Several datasets were created in the past decade to address the challenge of image question answering. In the following, we cover the datasets that are extensively used for this Human-Computer Interaction (HCI) themed task.
\paragraph{VQA v1.0.}\label{para:vqa1-data}VQA v1.0\footnote{\url{https://visualqa.org} \label{fnote: vqa1.0-dataset-url}} \shortcite{antol:2015} contains open-ended questions about images. These questions target different areas of an image, including background details and the underlying contexts. The answers are also open-ended and contain either a few words or a closed set of answers that can be provided in a multiple-choice format.  Table~\ref{table:vqa1.0-dataset-real-split} and Table~\ref{table:vqa1.0-dataset-abstract-split} present the dataset splits of images with \textit{real} and \textit{abstract} scenes observed in the dataset respectively.

\begin{table*}[!ht]
\newcommand{\midruleDVSScripts}{\cmidrule(lr){1-4} \cmidrule(lr){5-6}}
\small
\centering
\begin{tabular}{l | c c c | c c}
        \hline  %\toprule
        \rowcolor{teal!35}
             Dataset    &Real   &Questions  &Answers        &\multicolumn{2}{c}{\textbf{Textual Annotations}} \\\cline{5-6}
        \rowcolor{teal!35}
             Split      &Scenes &per Image  &per Question   &Questions    &Answers \\
        \hline \addlinespace[0.3em] %\midruleDVSScripts
            Training    &82,783 &3          &10             &248,349        &2,483,490  \\
            Validation  &40,504 &3          &10             &121,512        &1,215,120    \\
            Test        &81,434 &3          &10             &244,302        &2,443,020    \\
\bottomrule
\end{tabular}
\caption{\label{table:vqa1.0-dataset-real-split} Splits of the VQA v1.0 dataset with \textit{real} scenes.}
\end{table*}

\begin{table*}[!ht]
\newcommand{\midruleDVSScripts}{\cmidrule(lr){1-4} \cmidrule(lr){5-6}}
\small
\centering
\begin{tabular}{l | c c c | c c}
        \hline  %\toprule
        \rowcolor{teal!35}
             Dataset        &Abstract   &Questions  &Answers        &\multicolumn{2}{c}{\textbf{Textual Annotations}} \\\cline{5-6}
        \rowcolor{teal!35}
             Split          &Scenes     &per Image  &per Question   &Questions    &Answers \\
        \hline \addlinespace[0.3em] %\midruleDVSScripts
            Training        &20,000     &3          &10             &60,000        &600,000  \\
            Validation      &10,000     &3          &10             &30,000        &300,000    \\
            Test            &20,000     &3          &10             &60,000        &600,000    \\
\bottomrule
\end{tabular}
\caption{\label{table:vqa1.0-dataset-abstract-split} Splits of the VQA v1.0 dataset with \textit{abstract} scenes.}
\end{table*}

\paragraph{VQA v2.0.}\label{para:vqa2-data} VQA v2.0 extends VQA v1.0 and has three parts: \textit{Balanced Real Images}, \textit{Balanced Binary Abstract Scenes}, and \textit{Abstract Scenes}. Table~\ref{table:vqa2.0-dataset-bal-real-split} and Table~\ref{table:vqa2.0-dataset-bb-abstract-split} presents the dataset splits of the images with balanced real and binary abstract scenes observed in the dataset respectively. However, abstract scenes in VQA v2.0 are same as that of VQA v1.0.

\begin{table*}[!ht]
\newcommand{\midruleDVSScripts}{\cmidrule(lr){1-3} \cmidrule(lr){4-6}}
\small
\centering
\begin{tabular}{l | c c | c  c c}
        \hline  %\toprule
        \rowcolor{teal!35}
             Dataset    &Real       &Answers        &\multicolumn{3}{c}{\textbf{Textual Annotations}} \\\cline{4-6}
        \rowcolor{teal!35}
             Split      &Images     &per Question   &Questions      &Answers        &Complementary Pairs\\
        \hline \addlinespace[0.3em] %\midruleDVSScripts
            Training    &82,783     &10             &443,757        &4,437,570      &200,394    \\
            Validation  &40,504     &10             &214,354        &2,143,540      &95,144    \\
            Test        &81,434     &10             &447,793        &4,477,930      & -      \\
\bottomrule
\end{tabular}
\caption{\label{table:vqa2.0-dataset-bal-real-split} Splits of the VQA v2.0 dataset with balanced real images.}
\end{table*}

The term \textit{complementary pairs} in Table~\ref{table:vqa2.0-dataset-bal-real-split} means that a given question is associated with a pair of similar images such that the answer is different depending on the image (i.e. two different answers)

\begin{table*}[!ht]
\newcommand{\midruleDVSScripts}{\cmidrule(lr){1-1} \cmidrule(lr){2-3} \cmidrule(lr){4-5}}
\small
\centering
\begin{tabular}{l | c c | c c}
        \hline  %\toprule
        \rowcolor{teal!35}
             Dataset        & Binary Abstract   &Answers        &\multicolumn{2}{c}{\textbf{Textual Annotations}} \\\cline{4-5}
        \rowcolor{teal!35}
             Split          &Scenes             &per Question   &Questions    &Answers \\
        \hline \addlinespace[0.3em] %\midruleDVSScripts
            Training        &20,629             &10             &22,055        &220,550  \\
            Validation      &10,696             &10             &11,328        &113,280    \\
            % Test            & -                 &-              & -            &-    \\
\bottomrule
\end{tabular}
\caption{\label{table:vqa2.0-dataset-bb-abstract-split} Splits of VQA v2.0 with balanced binary abstract scenes.}
\end{table*}

\paragraph{Outside Knowledge VQA (OK-VQA).}OK-VQA\footnote{\url{https://okvqa.allenai.org} \label{fnote: ok-vqa-dataset-url}} \shortcite{marino:2019} uses a subset of MSCOCO (see Section~\ref{para:mscoco-data}) and is constructed with additional annotations such as questions, answers, knowledge category, etc. Table~\ref{table:ok-vqa-dataset-stats} presents more details about the dataset, while the Table~\ref{table:ok-vqa-dataset-splits} shows the splits of it.

\begin{table}[!ht]
\small
    \centering
    \begin{tabular}{l c c c c c c c}
    \hline  %\toprule
    \rowcolor{teal!35}
        Total    &Total       &Answers per   & Unique     & Unique    & Unique          & Total       & Average\\
    \rowcolor{teal!35}
        Images   &Questions   &Question      & Questions  & Answers   & Ques. Words     & Categories  & Ans. Length\\
    \hline \addlinespace[0.3em]     %\midrule
        14,031   & 14,055      & 5            & 12,591     & 14,454    & 7,178          & 10 + 1      & 1.3  \\
    \bottomrule
    \end{tabular}
    \caption{\label{table:ok-vqa-dataset-stats} Statistics of the OK-VQA dataset.}
\end{table}

%\begin{wraptable}{R}{7cm}
\begin{table*}[!ht]
\small
    \centering
    \begin{tabular}{l | c c}
    \hline  %\toprule
    \rowcolor{teal!35}
        Split       & Percent (\%)  & Questions \\
    \hline \addlinespace[0.3em]     %\midrule
        Training    & 64          & 9,009  \\
        Test     & 36         & 5,046   \\
    \midrule
        Total       & 100         & 14,055    \\
    \bottomrule
    \end{tabular}
    \caption{\label{table:ok-vqa-dataset-splits} Splits of the OK-VQA dataset.}
\end{table*}
%\end{wraptable}

\paragraph{Knowledge-aware VQA (KVQA).}The KVQA\footnote{\url{http://malllabiisc.github.io/resources/kvqa} \label{fnote: kvqa-dataset-url}} \shortcite{shahkvqa:2019} dataset was designed to emphasize questions that require access to external knowledge. Table~\ref{table:kvqa-dataset-stats} presents more details about the dataset, while Table~\ref{table:ok-vqa-dataset-splits} shows the splits of it. In order to get a mean score, the KVQA dataset provides five such splits.

\begin{table*}[!ht]
\small
    \centering
    \begin{tabular}{l c c c c c c}
    \hline %\toprule
    \rowcolor{teal!35}
        Total    &Q\&A       & Unique           & Unique      & Avg.    & Avg.      & Avg. number of\\
    \rowcolor{teal!35}
        Images   &Pairs      & Named Entities   & Answers     & Ques. Len  & Ans. Len     & Questions per Image\\
    \hline \addlinespace[0.3em] %\midrule
        24,602   & 183,007       & 18,880       & 19,571      & 10.14      & 1.64         & 7.44  \\
    \bottomrule
    \end{tabular}
    \caption{\label{table:kvqa-dataset-stats} Statistics of the KVQA dataset.}
\end{table*}

\begin{table*}[!ht]
\small
    \centering
    \begin{tabular}{l | c c c}
    \hline  %\toprule
    \rowcolor{teal!35}
        Split       & Percent (\%)  & Images    & Q\&A pairs \\
    \hline \addlinespace[0.3em]     %\midrule
        Training    & 70         & 17k       & 130k  \\
        Validation  & 20          & 5k        & 34k   \\
        Test     & 10         & 2k        & 19k   \\
    \bottomrule
    \end{tabular}
    \caption{\label{table:kvqa-dataset-splits} Splits of the KVQA dataset.}
\end{table*}

\subsubsection{Image Question Answering - Evaluation Measures, Models, and Results}
\label{sssec:imqaall}

In this section we describe only the evaluation measures used for \textit{Image Question Answering} as \textbf{Models}, \textbf{Results}, and some \textbf{Discussion} are extensively presented in the recent surveys~\shortcite{wuvqa:2017}.

\paragraph{Evaluation Measures} Image Q\&A models are evaluated based on the Accuracy measure.

\subsubsection{Video Question Answering - Introduction}
\label{sssec:viqaintro}
The goal of \textit{Video Question Answering} (Video Q\&A) is to answer natural language questions about videos. Unlike Image Q\&A, Video Q\&A is less explored. Nevertheless, there are a few works which have explored this spatio-temporal domain. One of the early attempts in this domain was jointly parsing the videos with corresponding text to answer queries~\shortcite{tu:2014}. Further, an open-ended Movie Q\&A~\shortcite{tapaswi:2016} with multiple-choice question pairs was designed to solve challenging questions that require semantic reasoning over a long temporal domain. Additionally, to limit the involvement of crowdworkers, the task was modified using fill-in-the-blank questions~\shortcite{zhu:2017,mazaheri:2017} and were automatically generated from different manually created video description datasets (Section~\ref{ssec:vddatasets}). Other works~\shortcite{zeng:2017} modified this dataset to support answering free-form natural language questions. Beyond this, open-ended video question answering is also addressed with methods such as spatio-temporal attentional encoder-decoder learning framework~\shortcite{zhao:2017}. There has been interest shown in jointly addressing multiple tasks that handle video and language. High-level concept words~\shortcite{yu:2017} are detected in order to be integrated with any video and language models addressing fill-in-the blank and multiple-choice test. Spatio-temporal reasoning from videos to answer questions has also been addressed by designing a spatial and temporal attention mechanism~\shortcite{jang:2017}.

\begin{figure}[!htb]
    \centering
        \includegraphics[width=\textwidth]{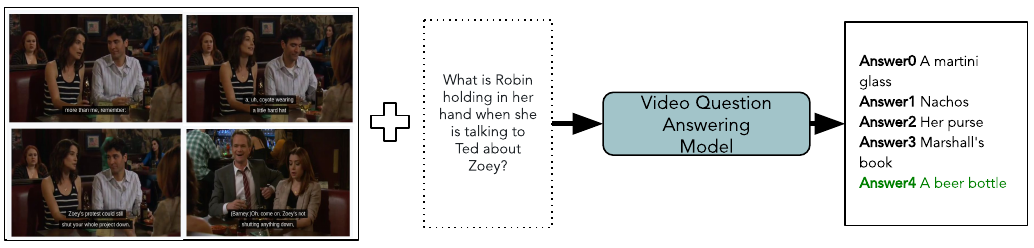}
    \caption{Given a \textit{video} (represented as sequence of frames from TV Q\&A dataset) and \textit{question}, a Video Question Answering model finds the right answer from Multiple Options.}\label{fig:videoqa}
\end{figure}

Recently, due to large interest in Video Q\&A, similar to Movie Q\&A, six popular TV shows were used to create a dataset, where questions are compositional~\shortcite{lei:2018}. The TV Q\&A dataset made the proposed multi-stream models to jointly localize relevant moments within a clip, comprehend subtitle-based dialogue, and then recognize relevant visual concepts. Furthermore, spatio-temporal grounding~\shortcite{lei:2019} is employed to link depicted objects to visual concepts in questions and answers. Figure~\ref{fig:videoqa} shows an example of this task, in which the model is given a video and a question and is asked to choose an answer from multiple choices.

\subsubsection{Video Question Answering - Datasets}
\label{sssec:viqadata}
Similar to image question answering, several datasets were created to address the challenge of video question answering. In the following, we cover those datasets that are popular and extensively used.

\paragraph{MovieQA.} The MovieQA\footnote{\url{http://movieqa.cs.toronto.edu/home} \label{fnote: movieqa-dataset-url}} \shortcite{tapaswi:2016} dataset is used to evaluate story comprehension of both video and text in an automatic manner. The dataset consists of almost 15,000 multiple choice questions and answers obtained from over 400 movies having high diversity. Table~\ref{table:movieqa-splits-statistics} reports the statistics and splits of the dataset.

\begin{table}[!ht]
\small
\centering
\begin{tabular}{l c c c c}
\hline  %\toprule
\rowcolor{teal!35}
                            & Training          & Validation        & Test           & Total               \\
\hline    %\midrule
%\rowcolor{olive!55}
\multicolumn{5}{c}{Movies with Plots and Subtitles} \\
\hline \addlinespace[0.3em]     %\midrule
             Movies         & 269               & 56                & 83                & 408                 \\
            QA pairs        & 9848              & 1958              & 3138              & 14944               \\
            Q words         & 9.3               & 9.3               & 9.5               & 9.3 $\pm$ 3.5       \\
            CA. words       & 5.7               & 5.4               & 5.4               & 5.6 $\pm$ 4.1       \\
 %           WA. words     & 5.2               & 5.0               & 5.1               & 5.1 $\pm$ 3.9       \\
\addlinespace[0.3em] \hline %\midrule
\multicolumn{5}{c}{Movies with Video Clips} \\
\hline \addlinespace[0.3em] %\midrule
            Movies                & 93      & 21                    & 26                & 140               \\
            QA pairs              & 4318    & 886                   & 1258              & 6462              \\
            Video clips           & 4385    & 1098                  & 1288              & 6771              \\
            Mean clip Length     & 201.0 s   & 198.5 s                 & 211.4s             & 202.7 $\pm$ 216.2 s \\
            Mean QA shots         & 45.6    & 49.0                  & 46.6              & 46.3  $\pm$ 57.1  \\
\bottomrule
\end{tabular}
\caption{\label{table:movieqa-splits-statistics} Statistics \& Splits of the MovieQA dataset. The column `Total' represents mean counts with standard deviations.}
\end{table}

\paragraph{TVQA.}The TVQA\footnote{\url{http://tvqa.cs.unc.edu} \label{fnote: tvqa-dataset-url}} \shortcite{lei:2018} dataset was created from videos of six different English TV shows, viz. \textit{Friends}, \textit{The Big Bang Theory}, \textit{How I Met Your Mother}, \textit{House M.D.}, \textit{Grey's Anatomy}, and \textit{Castle}. It consists of 460 hours of video and the questions are designed to be compositional, expecting the models to comprehend subtitles-based dialogue and to recognize relevant visual concepts. Table~\ref{table:tvqa-dataset-stats} presents the statistics of the dataset, while Table~\ref{table:tvqa-dataset-splits} shows the splits.

\begin{table}[!ht]
\small
    \centering
    \begin{tabular}{l c c c c c}
    \hline  %\toprule
    \rowcolor{teal!35}
         Video          & Video Clip    & Q\&A      & Total      & Questions per   & Answers per \\
    \rowcolor{teal!35}
         Clips          & Length      & Pairs     & Duration   & Video Clip      & Video Clip  \\
    \hline \addlinespace[0.3em]    %\midrule
        21,793          & 60 to 90 s    & 152,545    & 460 h  & 7               & 5  \\
    \bottomrule
    \end{tabular}
    \caption{\label{table:tvqa-dataset-stats} Statistics of the TVQA dataset.}
\end{table}

The testing data of TVQA is further split into two subsets named ``test-public'' containing 7,623 Q\&A pairs and ``test-reserved'' consisting of 7,630 Q\&A pairs. The \textit{test-public} set is available for the TVQA leaderboard\footnote{\url{http://tvqa.cs.unc.edu/leaderboard.html}} whereas \textit{test-reserved} is preserved for future use.

\begin{table}[!ht]
\small
    \centering
    \begin{tabular}{l | c c}
    \hline  %\toprule
    \rowcolor{teal!35}
        Split           & Percent (\%)   & Q\&A pairs \\
    \hline \addlinespace[0.3em]    %\midrule
        Training        & 80          & 122,039  \\
        Validation      & 10          & 15,253   \\
        Test            & 10          & 15,253   \\
    \bottomrule
    \end{tabular}
    \caption{\label{table:tvqa-dataset-splits} Splits of the TVQA dataset.}
\end{table}

The TVQA+\footnote{\url{http://tvqa.cs.unc.edu/download_tvqa_plus.html} \label{fnote: tvqa+-dataset-url}}~\shortcite{lei:2019} is an augmented subset of the original TVQA dataset where the augmentation comes in the form of bounding boxes linking depicted objects to visual concepts in both questions and answers. Table~\ref{table:tvqa+-dataset-statistics} presents the splits of TVQA+ dataset.

\begin{table*}[!ht]
\small
\centering
\begin{tabular*}{\textwidth}{l | c c c c c c c}
\hline 
\multirow{2}{*}{\cellcolor{teal!35}}    & \multirow{2}{*}{\cellcolor{teal!35}}  & \multirow{2}{*}{\cellcolor{teal!35}} & \cellcolor{teal!35}Avg. Span     &\cellcolor{teal!35} Avg. Video    &\cellcolor{teal!35} Annotated  &\cellcolor{teal!35} Bound.  & \multirow{2}{*}{\cellcolor{teal!35}}  \\
\multirow{-2}{*}{\cellcolor{teal!35} Split} & \multirow{-2}{*}{\cellcolor{teal!35} Q\&As}   & \multirow{-2}{*}{\cellcolor{teal!35} Clips}                       &\cellcolor{teal!35} Length (s)  &\cellcolor{teal!35} Length (s)  &\cellcolor{teal!35} Images     &\cellcolor{teal!35} Boxes     & \multirow{-2}{*}{\cellcolor{teal!35} Categories}  \\
\hline
Training    & 23,545        & 3,364     & 7.20      & 61.49         & 118,930       & 249,236       & 2,281 \\
Validation  & 3,017         & 431       & 7.26      & 61.48         & 15,350        & 32,682        & 769 \\
Test        & 2,821         & 403       & 7.18      & 61.48         & 14,188        & 28,908        & 680 \\ \hline
Total       & 29,383        & 4,198     & 7.20      & 61.49         & 148,468       & 310,826       & 2,527  \\\hline
\end{tabular*}
\caption{\label{table:tvqa+-dataset-statistics} Splits of the TVQA+ dataset.}
\end{table*}

\subsubsection{Video Question Answering - Evaluation Measures, Models and Results}
\label{sssec:viqaall}

In this section, we present the evaluation measures, models, and results achieved with various architectures of Video Q\&A.

\paragraph{Evaluation Measures.} Video Q\&A models are evaluated based on Accuracy. In addition, other measures such as Temporal mean Intersection-over-Union (Temp. mIoU)~\shortcite{hendricks:2017}, Answer-Span joint Accuracy (ASA), that jointly evaluates both answer prediction and span prediction, and object grounding performance calculated with mean Average Precision (Grd. mAP)~\shortcite{lei:2019} are used.

\paragraph{Models.}  The models which are created to address the task of \textit{Video Question Answering} aim to provide an overall understanding of the visual and the aligned textual content such as subtitles. In Table~\ref{arcvideoqa}, we present some exemplar architectures (refer to \textit{Combined} column) created to address the task by integrating both video and language. We also include a column that showcases the optimization techniques used to train those models.

\begin{table*}[!ht]
\small
  \centering
  \begin{tabular}{lcccccc}
    \hline  %\toprule
    \rowcolor{teal!35}
    {Approach} & Video & Frame & Language & Combined & Optimizer & RL \\ 
    \hline \addlinespace[0.3em]     %\midrule
    \shortcite{jang:2017}  & C3D & ResNet-152 & LSTM & ST-VQA & ADAM & \xmark \\ 
    \shortcite{lei:2018}  & - & R-CNN+ResNet-101  & BiLSTM  & Two-stream  & - & \xmark  \\ 
    \shortcite{lei:2019} &  - & R-CNN+ResNet-101 & BERT & STAGE & ADAM & \xmark \\
    \bottomrule
  \end{tabular}
  \caption{\label{arcvideoqa} Exemplar \textit{Video Question Answering} architectures.}
\end{table*}

\paragraph{Results.} Several models have been created to approach the task of \textit{Video Question Answering}. At the same time, many datasets have been created to provide diversity in the content so that they boost the generalization ability of the models. In this section, we cover the results achieved by the models on some representative datasets. Table~\ref{resvideoqatvqa} and Table~\ref{resvideoqatvqaplus} presents results obtained with a subset of models built using the TVQA and TVQA+ datasets presented in Section~\ref{sssec:viqadata}. Results for TVQA\footnote{\url{http://tvqa.cs.unc.edu/leaderboard.html}} and TVQA+\footnote{\url{https://competitions.codalab.org/competitions/22705#results}}   can also be found on the respective leaderboards.

\begin{table*}[!ht]
\small
  \centering
  \begin{tabular}{lc}
    \hline  %\toprule
    \rowcolor{teal!35}
    {Model} & Accuracy ($\uparrow$) \\ 
    \hline \addlinespace[0.3em]     %\midrule
    Random & 20.00 \\
    Retrieval-SkipThought & 24.77 \\
    Longest Answer & 30.22 \\
    NNS-SkipThought (Subtitle) & 38.29 \\
    NNS-TFIDF (Subtitle) & 50.79 \\
    Two-stream (Subtitle+Videos)~\shortcite{lei:2018} & 66.36 \\
    Three-stream (Subtitle+Videos+Questions)~\shortcite{lei:2018} & \textbf{68.48} \\
    \bottomrule
  \end{tabular}
  \caption{\label{resvideoqatvqa} Accuracy attained on TVQA test (public) set. All models use timestamp annotation without which the scores achieved by them are lower.}
\end{table*}

\begin{table*}[!ht]
\small
  \centering
  \begin{tabular}{lcccc}
    \hline  %\toprule
    \rowcolor{teal!35}
    {Model} & Accuracy & Grd. mAP ($\uparrow$) & Temp. mIOU ($\uparrow$) & ASA ($\uparrow$) \\ 
    \hline \addlinespace[0.3em]     %\midrule
    ST-VQA~\shortcite{jang:2017}  & 48.28 & - & - & - \\ 
    Two-stream~\shortcite{lei:2018}  & 68.13 & - & - & - \\
    STAGE-LXMERT~\shortcite{lei:2019} & 71.46 & 21.01 & 26.31 & 18.04 \\
    STAGE~\shortcite{lei:2019} & \textbf{74.83} & \textbf{27.34} & \textbf{32.49} & \textbf{22.23} \\ 
    \hline
    Human~\shortcite{lei:2019} & 90.46 & - & - & - \\
    \bottomrule
  \end{tabular}
  \caption{\label{resvideoqatvqaplus} Results obtained on TVQA+ test set. }
\end{table*}

\subsubsection{Video Question Answering - Discussion}
\label{sssec:viqaopenques}
It has been observed from STAGE~\shortcite{lei:2019} that aligned fusion is essential for improving Video Q\&A performance. STAGE uses all of the existing information such as Subtitles, Video, and Questions to build an efficient model. It has also proven to be effective if the models have access to the timestamp information as shown in Table~\ref{resvideoqatvqa}.

\subsection{Visual Reasoning}
\label{ssec:vr}
The goal in visual reasoning is to learn a model that comprehends the visual content by reasoning about it. Both images and videos are used as visual inputs for visual reasoning. In the following, we provide a detailed description of this complex and challenging task.

\subsubsection{Image Reasoning - Introduction}
\label{sssec:imreasonintro}
The goal of image reasoning is to answer sophisticated queries by reasoning about the visual world. Initial efforts~\shortcite{johnsonclevr:2017} aimed at designing diagnostic tests going beyond benchmarks such as VQA.  They reduced the biases by having detailed annotations describing the kind of reasoning each question requires. It has also been observed that VQA models struggle when comparing the attributes of objects, or when novel attribute combinations needs to be recognized (such as in compositional reasoning). A novel approach~\shortcite{johnson:2017} used a program generator to construct an explicit representation of the reasoning process, and an execution engine to execute the resulting program, producing an answer. Then, end-to-end module networks~\shortcite{hureason:2017} were proposed which learn to reason by directly predicting instance-specific network layouts without the aid of a parser as used in neural module networks.~\shortciteA{santoro:2017} went beyond and proposed Relation Networks (RNs) as a simple plug-and-play module to solve the problem of visual reasoning. RNs are further used to learn relation-aware visual features for content based image retrieval~\shortcite{messina:2018} and also Multi-Relational Networks~\shortcite{chang:2018}. Furthermore, global context reasoning~\shortcite{cao:2018} is explored for better aligning image and language domains in diverse and unrestricted cases.

A recent approach~\shortcite{perez:2018} introduced a general-purpose conditioning method called Feature-wise Linear Modulation (FiLM) layers which influence neural network computation via a simple, feature-wise affine transformation based on conditioning information. FiLM was modified by~\shortciteA{strubv:2018} to generate parameters of FiLM layers going up the hierarchy of a convolutional network in a multi-hop fashion rather than all at once. Cascaded Mutual Modulation (CMM)~\shortcite{yaocasc:2018} is an end-to-end visual reasoning model that also uses the FiLM technique to enable the textual/visual pipeline to mutually control each other. Another approach modified neural modular networks~\shortcite{huexp:2018} such that it performs compositional reasoning by automatically inducing a desired sub-task decomposition without relying on strong supervision.~\shortciteA{mascharka:2018} proposed a set of visual-reasoning primitives which, when composed, manifest as a model capable of performing complex reasoning tasks in an explicitly interpretable manner. Also, in the context of interpretable learning frameworks, Learning-By-Asking (LBA)~\shortcite{misralba:2018} attempted to closely mimic natural learning with the goal to make it more data efficient than the traditional VQA setting. Further, compositional attention networks~\shortcite{hudsonmac:2018} were designed as fully differentiable neural network architectures to facilitate explicit and expressive reasoning. The goal of this architecture is to provide a strong prior for iterative reasoning, allowing it to support structured learning, as well as to generalize from a modest amount of data.

Recently, neural-symbolic visual question answering~\shortcite{yi:2018} attempted to combine deep representation learning with symbolic program execution. It first recovers structural scene representation from the image and a program trace from the question. This was extended with a Neuro-Symbolic Concept Learner (NS-CL)~\shortcite{mao:2019} that learns visual concepts, words, and semantic parsing of sentences without explicit supervision. It learns by simply looking at images and reading paired questions and answers. Further, a multimodal relational network (MuRel)~\shortcite{cadene:2019} was proposed to learn end-to-end reasoning over real images. Additionally,~\shortciteA{aditya:2019} used spatial knowledge to aid visual reasoning. Their framework combined knowledge distillation, relational reasoning, and probabilistic logical languages. Existing diagnostic tests have been further modified with referring expressions to handle bias~\shortcite{liuclevr:2019} and with structural, relational, and analogical reasoning in a hierarchical representation~\shortcite{zhangraven:2019}. Explainable and explicit neural modules~\shortcite{shi:2019} have also been explored with scene graphs. Objects as nodes and pairwise relationships as edges were used for explainable and explicit reasoning with structured knowledge.

\begin{figure}[!htb]
    \centering
        \includegraphics[width=\textwidth]{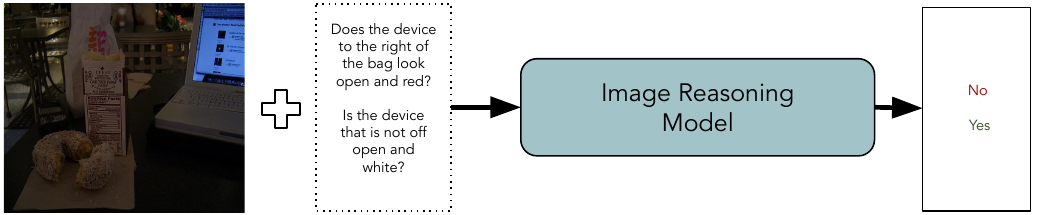}
    \caption{Given a \textit{real-world image} and a \textit{question}, an Image Reasoning Model reasons about the question to produce an answer.}\label{fig:compvqa}
\end{figure}

Further expanding the scope of inquiry on this subject, \shortciteA{andreasjcqa:2016,andreasj:2016} exploit the compositional linguistic structure of complex questions by forming neural module networks which query about the abstract shapes observed in an image. Improvement is further seen in how images are interpreted. For example, compositional question answering~\cite{hudson:2019} was addressed with scene graph structures on real-world images going beyond abstract shapes.  Figure~\ref{fig:compvqa} demonstrates the task of reasoning about real-world images.

\shortciteA{zellers:2019} introduced a reasoning task that requires commonsense knowledge, while the goal of NLVR~\shortcite{suhr-nlvr:2017} and NLVR2~\shortcite{suhr-nlvr2:2019} tasks is to determine whether a sentence is true about a visual input or not.

\subsubsection{Image Reasoning - Datasets}
\label{sssec:imagereasoningdata}
For image reasoning, both \textit{real} and \textit{synthetic} image datasets have been developed. In the following, we present the datasets belonging to both of these categories. 
\paragraph{Compositional Language and Elementary Visual Reasoning (CLEVR).}\label{para:clevrdata}CLEVR\footnote{\url{https://cs.stanford.edu/people/jcjohns/clevr} \label{fnote: clevr-dataset-url}} \shortcite{johnsonclevr:2017} is a diagnostic dataset created using a 3D computer graphics toolkit known as Blender\footnote{\url{https://www.blender.org}}. It consists of synthetic images of simple 3D objects that vary in their attributes, viz. size, color, shape, and material. Images contain three to ten different combinations of these objects and attributes and are arranged in different spatial positions. Such complex configurations require good visual reasoning capabilities from VQA models to produce correct answers. Table~\ref{table:clevr-dataset} presents the splits of dataset.

\begin{table}[!ht]
\small
    \centering
    \begin{tabular}{l | c c c c}
    \hline  %\toprule
    \rowcolor{teal!35}
        Split       & Images    & Questions     & Unique Questions  & Overlap with train\\
    \hline \addlinespace[0.3em]     %\midrule
        Training    & 70,000    & 699,989       & 608,607           & - \\
        Validation  & 15,000    & 149,991       & 140,448           & 17,338   \\
        Test        & 15,000    & 149,988        & 140,352           & 17,335    \\
    \midrule
        Total       & 100,000   & 999,968       & 853,554           & - \\
    \bottomrule
    \end{tabular}
    \caption{\label{table:clevr-dataset} Splits of the CLEVR dataset.}
\end{table}

\paragraph{Natural Language Visual Reasoning (NLVR).}
The Cornell Natural Language for Visual Reasoning dubbed as NLVR\footnote{\url{http://lil.nlp.cornell.edu/nlvr} \label{fnote:nlvr-dataset-url}}~\shortcite{suhr-nlvr:2017} is a multimodal dataset that comes with natural language sentences grounded in synthetic images. The images are rendered and encapsulate different objects such as triangles, circles, and squares. These objects come in various sizes and are placed at different positions within images. The descriptions of the images were manually written by crowdworkers. Table~\ref{table:nlvr-dataset} presents the official splits of the dataset for evaluation purposes.

\begin{table}[!ht]
\small
    \centering
    \begin{tabular}{l | c c}
    \hline  %\toprule
    \rowcolor{teal!35}
        Split       & Unique Sentences  & Examples\\
    \hline \addlinespace[0.3em]     %\midrule
        Training    & 3,163             & 74,460 \\
        Validation  & 267               & 5,940 \\
        Test-P      & 266               & 5,934 \\
        Test-U      & 266               & 5,910 \\
    \midrule
        Total       & 3,962             & 92,244 \\
    \bottomrule
    \end{tabular}
    \caption{\label{table:nlvr-dataset} Splits of the NLVR dataset. \textit{Test-P} and \textit{Test-U} means Test set (public) and Test set (unreleased) respectively.}
\end{table}

\paragraph{Natural Language Visual Reasoning \textit{for Real} (NLVR2).}
The limitations such as limited expressivity and semantic diversity that arose due to the synthetic nature of the NLVR dataset, has been addressed in the next incarnation of NLVR named as Natural Language for Visual Reasoning for Real, NLVR2\footref{fnote:nlvr-dataset-url}~\shortcite{suhr-nlvr2:2019}. Similar to NLVR, the images in NLVR2 also come as a pair along with a grounded natural language description.  Table~\ref{table:nlvr2-dataset} presents the official splits of the dataset.

\begin{table}[!ht]
\small
    \centering
    \begin{tabular}{l | c c}
    \hline  %\toprule
    \rowcolor{teal!35}
        Split       & Unique Sentences  & Examples\\
    \hline \addlinespace[0.3em]     %\midrule
        Training    & 23,671            & 86,373 \\
        Validation  & 2,018             & 6,982 \\
        Test-P      & 1,995             & 6,967 \\
        Test-U      & 1,996             & 6,970 \\
    \midrule
        Total       & 29,680            & 107,292 \\
    \bottomrule
    \end{tabular}
    \caption{\label{table:nlvr2-dataset} Splits of the NLVR2 dataset. \textit{Test-P} denotes Test set Public, whereas \textit{Test-U} means Test set Unreleased.}
\end{table}

\paragraph{CLEVR-CoGenT.}A modified version of CLEVR is the Compositional Generalization Test (CLEVR-CoGenT)\footref{fnote: clevr-dataset-url}~\shortcite{johnsonclevr:2017} dataset. It is used to test models' ability to find novel combinations of attributes at test-time. There are two types of \textit{conditions} in this dataset, viz. Condition A and Condition B, where based on the condition, the color of the geometrical shape can vary as show in Table \ref{table:clevr-condition-types}. Based on these \textit{conditions}, the CLEVR-CoGenT dataset is divided for evaluation purposes as shown in Table \ref{table:clevr-cogent-dataset-split}.
\begin{table}[!ht]
\small
    \centering
    \begin{tabular}{l c c}
    \hline  %\toprule
    \rowcolor{teal!35}
        Geometrical Shape           &Condition     &Colors of Geometrical Shape \\
    \hline \addlinespace[0.3em]    %\midrule
        \multirow{2}{*}{Cubes}      &A              &gray, blue, brown, yellow \\
                                    &B              &red, green, purple, cyan \\
        \hline
        \multirow{2}{*}{Cylinders}  &A              &red, green, purple, cyan \\
                                    &B              &gray, blue, brown, yellow \\
        \hline
        \multirow{2}{*}{Spheres}    &A              &any color \\
                                    &B              &any color \\
    \bottomrule
    \end{tabular}
    \caption{ \label{table:clevr-condition-types} \textit{Conditions} in the CLEVR-CoGenT dataset.}
   
\end{table}
\begin{table}[!ht]
\small
    \centering
    \begin{tabular}{l | c c c c}
    \hline  %\toprule
    \rowcolor{teal!35}
        Split                           &Condition  &Images      &Questions   \\
    \hline \addlinespace[0.3em]    %\midrule
        Training                        & A         &70,000      &699,960     \\
    \midrule
        \multirow{2}{*}{Validation}     & A         &15,000     &150,000    \\
                                        & B         &15,000     &149,991    \\
        \midrule
        \multirow{2}{*}{Test}           & B         &15,000     &149,980    \\
                                        & B         &15,000     &149,992    \\
    \bottomrule
    \end{tabular}
    \caption{\label{table:clevr-cogent-dataset-split} Splits of the CLEVR-CoGenT dataset.}
\end{table}

\paragraph{GQA.}\label{para:gqadata}The GQA\footnote{\url{https://cs.stanford.edu/people/dorarad/gqa} \label{fnote: gqa-dataset-url}}~\shortcite{hudson:2019} dataset was created to address the shortcomings in earlier VQA datasets. GQA consists of compositional questions over real-world images. Each image is associated with a scene graph of the image's objects, attributes, and relations. Also, each question is associated with a structured representation of its semantics. Table~\ref{table:gqa-dataset} presents the statistics and splits of the dataset.

\begin{table}[!ht]
\small
    \centering
    \begin{tabular}{l c c | c c c | c}
    \hline  %\toprule
    \rowcolor{teal!35}
        Images        & Questions       & Vocabulary Size  & Training   & Validation    & Testing   & Challenge\\
    \hline \addlinespace[0.3em]    %\midrule
        113,018       & 22,669,678      & 3,097            & 70\%       & 10\%          & 10\%      & 10\%   \\
    \bottomrule
    \end{tabular}
    \caption{\label{table:gqa-dataset} Statistics \& splits of the GQA dataset.}
\end{table}

\paragraph{Relational and Analogical Visual rEasoNing (RAVEN).}\label{para:ravendata}The RAVEN\footnote{\url{http://wellyzhang.github.io/project/raven.html} \label{fnote: raven-dataset-url}}~\shortcite{zhangraven:2019} dataset was designed to perform relational and analogical visual reasoning. It is built by keeping in mind Raven's Progressive Matrices (RPM)~\shortcite{burke:1958}. Furthermore, it associates vision with structural, relational, and analogical reasoning in a hierarchical representation. The dataset is split into training, validation, and testing in the ratio 6:2:2 respectively. Table~\ref{table:raven-dataset} presents the statistics of the dataset.

\begin{table}[!ht]
\small
    \centering
    \begin{tabular}{l c c c c c}
    \hline  %\toprule
    \rowcolor{teal!35}
                                & RPM         & Tree-structure    & Structural      & Rule            & Avg. rules \\
    \rowcolor{teal!35}
    \multirow{-2}{*}{Images}    & Problems    & per problem       & Labels          & Annotations     & per problem  \\
    \hline \addlinespace[0.3em]    %\midrule
        1,120,000               & 70,000      & 16                & 1,120,000       & 440, 000        & 6.29     \\
    \bottomrule
    \end{tabular}
    \caption{\label{table:raven-dataset} Statistics of the RAVEN dataset.}
\end{table}

\paragraph{Visual Commonsense Reasoning (VCR).}\label{para:vcrdata}VCR\footnote{\url{https://visualcommonsense.com} \label{fnote: vcr-dataset-url}}~\shortcite{zellers:2019} is a large-scale dataset for achieving cognition-level visual understanding. It contains about 110k images, 290k multiple choice questions and correspondingly 290k correct answers and rationales. This dataset is very diverse and, consequently, it is challenging. Table~\ref{table:vcr-dataset} presents the official splits and some high-level statistics of the dataset.
\begin{table}[!ht]
    \small\centering
    \begin{tabular}{l | c c c}
        \hline  %\toprule
        \rowcolor{teal!35}
        Dataset Characteristic              &Train      &Validation     &Test \\
        \hline \addlinespace[0.3em]    %\midrule
        Number of questions                 & 212,923   & 26,534        & 25,263 \\
        Number of answers per question      & 4         & 4             & 4 \\
        Number of rationales per question   & 4         & 4             & 4 \\ \midrule
        Number of images                    & 80,418    & 9,929         & 9,557 \\
        Number of movies covered            & 1,945     & 244           & 189 \\ \midrule
        Average question length             & 6.61      & 6.63          & 6.58 \\
        Average answer length               & 7.54      & 7.65          & 7.55 \\ 
        Average rationale length            & 16.16     & 16.19         & 16.07 \\ 
        Average num. of objects mentioned   & 1.84      & 1.85          & 1.82 \\ 
        \bottomrule
    \end{tabular}
    \caption{\label{table:vcr-dataset} High-level statistics of the VCR dataset. One fold in the dataset was held-out for blind evaluation at a later date. Hence, the statistics of that fold are not shown here. }
\end{table}

\paragraph{Visual {COM}monsense r{E}asoning in {T}ime (Visual {COMET}).}\label{para: visualcomet}Visual COMET\footnote{\url{https://visualcomet.xyz}\label{fnote: visualcomet-dataset-url}}~\shortcite{park:2020} is a large-scale dataset of Visual Commonsense Graphs for reasoning about the dynamic context of static images in order to achieve cognitive visual scene understanding. VisualCOMET contains images with person grounding (i.e., multimodal co-reference chains) and the images are connected with inference sentences. Table~\ref{table:visual-comet-dataset} presents the official splits and more statistics about the dataset.

\begin{table}[!ht]
    \centering
    \begin{tabular}{l|c c | c c c| c}
    \hline  %\toprule
    \rowcolor{teal!35}
                            & Images/   & Events at      &\multicolumn{3}{c|}{Inferences on}                 & Total \\\cline{4-6}
    \rowcolor{teal!35}
    \multirow{-2}{*}{Split} & Places    & Present    &Events Before   &Intents at Present   &Events After   & Inferences \\
    \hline \addlinespace[0.3em]    %\midrule
            Train           & 47,595    & 111,796    & 467,025        & 237,608             &469,430        & 1,174,063 \\
            Dev             & 5,973     & 13,768     & 58,773         & 28,904              &58,665         & 146,332  \\
            Test            & 5,968     & 13,813     & 58,413         & 28,568              &58,323         & 145,309  \\
    \hline
            Total           & 59,356    & 139,377    & 584,211        & 295,080             & 586,418       & 1,465,704 \\
    \bottomrule
    \end{tabular}
    \caption{Statistics and splits of the Visual Commonsense Graph dataset.}
    \label{table:visual-comet-dataset}
\end{table}

\subsubsection{Image Reasoning - Evaluation Measures, Models, and Results}
\label{sssec:imagereasall}

In this section, we review the measures used to evaluate different models of \textit{Image Reasoning} and the results obtained by them.

\paragraph{Evaluation Measures.} The standard evaluation measures such as Accuracy are used for benchmarking purposes. However, there are evaluation measures that are explicitly used for \textit{Image Reasoning} (e.g., CLEVR), viz. \textbf{Querying Attribute (QA)} that uses questions to ask about an attribute of a particular object, \textbf{Compare Attribute (CA)} which uses comparison questions for asking whether two objects have the same value for some attribute, \textbf{Compare Numbers (CN)} which uses comparison questions to ask which of two object sets is larger, \textbf{Count} which asks counting questions to find the number of objects fulfilling some conditions, and \textbf{Exist} which asks existence questions to check whether a certain type of object is present or not.

\paragraph{Models.} The models that are designed to approach the task of \textit{Image Reasoning} are built such that they provide an effective way of reasoning about vision with language as additional input. In Table~\ref{arcimagereason}, we present some exemplar architectures (refer to \textit{Combined} column) created to address the task by integrating both image and language. We also include a column that showcases the optimization techniques used to train the \textit{Image Reasoning} models.

\begin{table*}[!ht]
\small
  \centering
  \begin{tabular}{lccccc}
    \hline  %\toprule
    \rowcolor{teal!35}
    Approach & Image & Language & Combined & Optimizer & RL\\
    \hline \addlinespace[0.3em]     %\midrule
    ~\shortcite{johnsonclevr:2017} & ResNet-101 & LSTM & SA+MLP & ADAM & \xmark\\
    ~\shortcite{hureason:2017} & VGG & LSTM & N2NMN & ADAM & \cmark\\
    ~\shortcite{johnson:2017} & ResNet-101 & LSTM & PGEE & ADAM & \cmark\\
    ~\shortcite{santoro:2017} & Custom & LSTM & RN & ADAM & \xmark \\
    ~\shortcite{cao:2018} & ResNet-101 & BiLSTM & ACMN & ADAM & \xmark \\
    ~\shortcite{perez:2018} & ResNet-101 & GRU & FiLM & ADAM & \xmark \\
    ~\shortcite{hudsonmac:2018} & ResNet-101 & BiLSTM & MAC & ADAM & \xmark\\
    ~\shortcite{mascharka:2018}& ResNet-101 & - & TbD & ADAM & \xmark \\
    ~\shortcite{haurilets:2019}& ResNet-152 & LSTM & FinalDestGraph & ADAM & \xmark \\
    ~\shortcite{hu:2019} & ResNet-101 & LSTM & LCGN & ADAM & \xmark \\
    ~\shortcite{mao:2019} & ResNet-34 & BiGRU & NS-CL & - & \cmark\\
    \bottomrule
  \end{tabular}
  \caption{\label{arcimagereason} Exemplar \textit{Image Reasoning} architectures. ``Custom'' - Own CNN architecture.}
\end{table*}

\paragraph{Results.} The models designed on different \textit{Image Reasoning} datasets aim to achieve generalization. In this section, we cover the results achieved by the models from some representative datasets. Table~\ref{resimagereasonclevr}, Table~\ref{resimagereasongqa}, Table~\ref{resimagereasongvcr}, and Table~\ref{resimagereasonraven} presents results obtained with a subset of models built using the datasets such as CLEVR, GQA, VCR, and RAVEN that were presented in Section~\ref{sssec:imagereasoningdata}. Results for the NLVR and NLVR2 tasks can be found on the respective leaderboards\footnote{\url{http://lil.nlp.cornell.edu/nlvr}}.  

\begin{table}[!ht]
\small
  \centering
  \begin{tabular}{lccccccc}
    \hline      %\toprule
    \rowcolor{teal!35}
    {Model} & Count & Exist & CN & QA & CA & Overall \\
    \hline \addlinespace[0.3em]     %\midrule
    CNN+LSTM+SA+MLP~\shortcite{johnsonclevr:2017} & 59.7 & 77.9 & 75.1 & 80.9 & 70.8 & 73.2 \\
    N2NMN+700KProgLabel~\shortcite{hureason:2017} & 68.5 & 85.7 & 84.9 & 90.0 & 88.7 & 83.7 \\
    PGEE+700KProgLabel~\shortcite{johnson:2017} & 92.7 & 97.1 & 98.7 & 98.1 & 98.9 & 96.9 \\
    CNN+LSTM+RN~\shortcite{santoro:2017} & 90.1 & 97.8 & 93.6 & 97.9 & 97.1 & 95.5 \\
    ACMN~\shortcite{cao:2018} & 94.2 & 81.3 & 81.6 & 90.5 & 97.1 & 89.3 \\
    CNN+GRU+FiLM~\shortcite{perez:2018} & 94.3 & 99.1 & 96.8 & 99.1 & 99.1 & 97.7 \\
    MAC~\shortcite{hudsonmac:2018} & 97.2 & \textbf{99.5} & 99.4 & 99.3 & 99.5 & 98.9 \\
    TbD+700KProgLabel~\shortcite{mascharka:2018} & 97.6 & 99.2 & 99.4 & \textbf{99.5} & 99.6 & \textbf{99.1} \\
    FinalDestGraph~\shortcite{haurilets:2019} & 91.3 & 98.6 & \textbf{99.6} & \textbf{99.5} & \textbf{99.8} & 97.5 \\
    LCGN+single-hop~\shortcite{hu:2019} & - & - & - & - & -& 97.9 \\
    NS-CL~\shortcite{mao:2019} & \textbf{98.2} & 98.8 & 99.0 & 99.3 & 99.1 & 98.9 \\
    \bottomrule
  \end{tabular}
  \caption{\label{resimagereasonclevr} Comparison of different models on the CLEVR dataset.}
\end{table}

\begin{table}[!ht]
\small
  \centering
  \begin{tabular}{lccccccc}
    \hline  %\toprule
    \rowcolor{teal!35}
    {Model} & val & test-dev & test \\
    \hline \addlinespace[0.3em]     %\midrule
    CNN+LSTM~\shortcite{hudson:2019} & 49.2 & - & 46.6 \\
    Bottom-up~\shortcite{anderson:2018} & 52.2 & - & 49.7 \\
    MAC~\shortcite{hudsonmac:2018} & 57.5 & - & 54.1 \\
    LCGN+single-hop~\shortcite{hu:2019} & \textbf{63.8} & \textbf{55.6} & \textbf{56.0} \\
    \bottomrule
  \end{tabular}
  \caption{\label{resimagereasongqa} Comparison of accuracy (\%) scores of different methods on the validation (val), test-dev, and test splits of the GQA dataset.}
\end{table}

\begin{table}[!ht]
\small
\centering

\begin{tabular}{lcccccc }
 \hline
  \rowcolor{teal!35}
\multirow{2}{*}{} 
& \multicolumn{2}{c}{(Q$\rightarrow$ A)} 
& \multicolumn{2}{c}{(QA$\rightarrow$ R)} 
& \multicolumn{2}{c}{(Q$\rightarrow$ AR)} \\
\rowcolor{teal!35}
\multirow{-2}{*}{Model}  & val & test & val & test & val & test \\
\hline
R2C~\shortcite{zellers:2019} & 63.8 & 65.1 & 67.2 & 67.3 & 43.1 & 44.0 \\
ViLBERT~\shortcite{luvilbert:2019} & 72.4 &73.3 &74.5 &74.6 &54.0 &54.8 \\
B2T2~\shortcite{albertifusion:2019}&71.9 &72.6 &76.0 &75.7 &54.9 &55.0 \\
VL-BERT~\shortcite{suvlbert:2019}&73.7 &74.0 &74.5 &74.8 &55.0 &55.5 \\
Unicoder-VL~\shortcite{liunicoder:2019}  &72.6 &73.4 &74.5 &74.4 &54.5 &54.9    \\
\hline
\end{tabular}
 \caption{\label{resimagereasongvcr} Comparison of accuracy (\%) scores of different models on the validation (val) and test splits of the VCR dataset.}
\end{table}

\begin{table}[!ht]
\small
  \centering
  \begin{tabular}{lccccccc}
    \hline  %\toprule
    \rowcolor{teal!35}
                            & & 2x2  & 3x3     &     &   &   &  \\
    \rowcolor{teal!35}
    \multirow{-2}{*}{Model} &\multirow{-2}{*}{Acc} & Grid & Grid &\multirow{-2}{*}{L-R}  &\multirow{-2}{*}{U-D}  &\multirow{-2}{*}{O-IC}  &\multirow{-2}{*}{O-IG}  \\
    \hline \addlinespace[0.3em]     %\midrule
    WReNDRT~\shortcite{santoro:2018} & 15.02  & 23.26 & 29.51 & 6.99 & 8.43 & 8.93 & 12.35 \\
    ResNetDRT~\shortcite{zhangraven:2019} & \textbf{59.56} &  \textbf{46.53} & \textbf{50.40} & \textbf{65.82} & \textbf{67.11} & \textbf{69.09} & \textbf{60.11} \\
    \midrule
    Human~\shortcite{zhangraven:2019} & 84.41 & 81.82 & 79.55 & 86.36 & 81.81 & 86.36 & 81.81 \\
    PerfectSolver & 100 & 100 & 100 & 100 & 100 & 100 & 100 \\
    \bottomrule
  \end{tabular}
  \caption{\label{resimagereasonraven} Comparison of accuracy (\%) scores of different models on the RAVEN dataset.}
\end{table}

\subsubsection{Image Reasoning - Discussion}
\label{sssec:imagereasopenques}
The task of \textit{Image Reasoning} has been studied using different types of datasets. Initially, a synthetic dataset, viz. CLEVR, was used. Later, real-world datasets like GQA were created for developing more complex vision and language integration models. Table~\ref{resimagereasonclevr} shows the results for the CLEVR dataset. Recently introduced Neuro-Symbolic Concept Learner (NS-CL)~\shortcite{mao:2019} reaches state-of-the-art results without explicit supervision on visual concepts, words, and semantic parsing of sentences. However, for the real-world image datasets like GQA, the approach by~\shortciteA{hu:2019} that creates Language-Conditioned Graph Networks (LGCN) providing different hops to effectively support relational reasoning achieve best results. Most of the works that outperform on the VCR task are pretrained and fine-tuned as shown in Table~\ref{resimagereasongvcr}.

The RAVEN dataset differs from both CLEVR and GQA as it depends only on the image input. We can observe from Table~\ref{resimagereasonraven} that a perfect solver achieves 100\% accuracy, while the approach introduced by~\shortciteA{zhangraven:2019} achieves reasonable system performance.

\subsubsection{Video Reasoning - Introduction}
\label{sssec:videoreasontaskintro}
\begin{figure}[!htb]
    \centering
        \includegraphics[width=\textwidth]{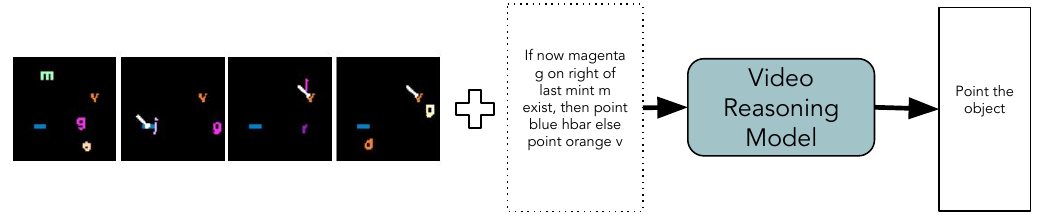}
    \caption{Given a \textit{video} (represented as a sequence of synthetic 2D images~\shortcite{yangcog:2018}) and a \textit{question}, a Video Reasoning Model reasons about the video to perform the task presented to it in the question.}\label{fig:videoreason}
\end{figure}

When compared to image reasoning, the video reasoning task is in its nascent stages and hence there is no clearly defined goal. However, for video reasoning, there exists a task of configurable visual question and answer (COG) designed by~\shortciteA{yangcog:2018}. The goal of COG is to address problems related to visual and logical reasoning and memory. To be more concrete, the task is aimed at deducing the correct answer by pointing to the right object while taking into account the changes of the scene i.e., from both spatial and temporal perspective. Figure~\ref{fig:videoreason} demonstrates the task of temporal reasoning about synthetic 2D scenes resembling video input.

Further, ~\shortciteA{haurilets:2019} addressed both image and video reasoning by introducing the concept of a question-based visual guide to constrain the potential solution space by learning an optimal traversal scheme. In their approach, the final destination nodes alone are used to produce the answers.

\subsubsection{Video Reasoning - Datasets}
\label{sssec:videoreasondata}
There are not many datasets for video reasoning. One of the few examples is listed below.

\paragraph{Configurable Visual Question and Answer (COG).}\label{para:videoreasoningdata}COG\footnote{\url{https://github.com/google/cog#datasets} \label{fnote: cog-dataset-url}}~\shortcite{yangcog:2018} was created to parallel experiments in humans and animals. Table~\ref{table:cog-dataset} presents splits of the dataset.

\begin{table}[!ht]
\small
    \centering
    \begin{tabular}{l | c c c}
    \hline  %\toprule
    \rowcolor{teal!35}
                            & Total             & Examples per \\
    \rowcolor{teal!35}
    \multirow{-2}{*}{Split} & Examples          & Task Family  \\
    \hline \addlinespace[0.3em]    %\midrule
        Training            & 10,000,320         &  227,280 \\
        Validation          & 500,016           &  11,364 \\
        Test                & 500,016           & 11,364 \\
    \bottomrule
    \end{tabular}
    \caption{\label{table:cog-dataset} Splits of the COG dataset.}
\end{table}

\subsubsection{Video Reasoning - Evaluation Measures, Models, and Results}
\label{sssec:videoreasall}

In this section, we review the measures used to evaluate different models of \textit{Video Reasoning} and the results obtained by them.

\paragraph{Evaluation Measures.} For \textit{Video Reasoning} (e.g., COG) the evaluation measures used are based on changes of the scene in three different query types.
\begin{itemize}
    \item \textbf{Pointing (Point)} queries ask the system to point to a certain object.
    \item \textbf{Yes/No} seeks a binary decision, while \textbf{Conditional (Condit)} is composed of questions based on objects that needs to fulfill certain conditions.
    \item \textbf{Attribute-related (Atts)} which is composed of questions about certain attributes.
\end{itemize}

\paragraph{Models.} Many models have been created to approach the task of \textit{Video Reasoning}. In Table~\ref{arcvideoreason}, we present some exemplar architectures (refer to \textit{Combined} column) created to address the task by integrating video and language.

\begin{table*}[!ht]
\small
  \centering
  \begin{tabular}{lccccc}
    \hline  %\toprule
    \rowcolor{teal!35}
    {Approach} & Video & Frame & Language & Combined & RL \\ 
    \hline \addlinespace[0.3em]     %\midrule
    ~\shortcite{yangcog:2018} & - & Custom & LSTM & WorkMemory & \xmark \\
    ~\shortcite{haurilets:2019} & -& ResNet-152 & LSTM & FinalDestGraph & \xmark \\
    \bottomrule
  \end{tabular}
  \caption{\label{arcvideoreason} Exemplar \textit{Video Reasoning} architectures.}
\end{table*}

\paragraph{Results.} As discussed earlier, several models have been created to approach the task of \textit{Video Reasoning}. In Table~\ref{resvideoreason}, we present the results obtained with a subset of models built using the COG dataset presented in Section~\ref{sssec:videoreasondata}.

\begin{table}[!ht]
\small
  \centering
  \begin{tabular}{lccccc}
    \hline  %\toprule
    \rowcolor{teal!35}
    {Model} & Atts & Condit & Point & Yes/No & All \\
    \hline \addlinespace[0.3em]     %\midrule
    WorkMemory~\shortcite{yangcog:2018} & - & - & - & - & 93.7 \\
    QuestionNodes~\shortcite{haurilets:2019} & 73.7 & 63.5 & 92.5 & 57.9 & 63.3 \\
    FinalDestGraph~\shortcite{haurilets:2019} & \textbf{99.2} & \textbf{98.4} & \textbf{100.0} & \textbf{95.0} & \textbf{97.2} \\
    \bottomrule
  \end{tabular}
  \caption{\label{resvideoreason} Comparison of measures using different methods on the COG dataset.}
\end{table}

\subsubsection{Video Reasoning - Discussion}
\label{sssec:videoreasonopenques}
The results presented in Table~\ref{resvideoreason} show that the recently proposed approach by~\shortciteA{haurilets:2019} achieves the best result on different task-specific measures. This approach proposes a question-based visual guide, which constrains the potential solution space by learning an optimal traversal scheme of a graph.

\subsection{Visual Entailment}
\label{ssec:venttask}
Goal of the \textit{Visual Entailment} task is to learn a model that predicts whether the visual content entails the augmented text along with hypothesis. Both images and videos are used as visual inputs. In the following, we describe the task, datasets used, and the approaches that have been proposed to tackle the problem.

\subsubsection{Image Entailment - Introduction}
\label{sssec:imageentailintro}
\begin{figure}[!htb]
    \centering
        \includegraphics[width=\textwidth]{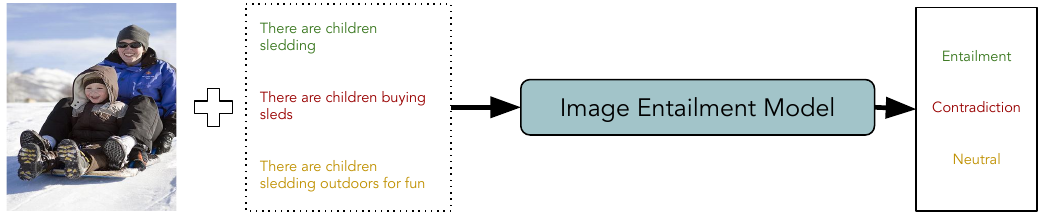}
    \caption{Given an image as a \textit{premise} and a natural language text as a \textit{hypothesis}, an Image Entailment Model predicts whether the hypothesis is an entailment, contradiction, or neutral by understanding the evidence(s) present in the image.}\label{fig:visualentail}
\end{figure}

To address the perceived drawbacks of VQA and visual reasoning, i.e. that they deal with similar objects and sentence structures,~\shortciteA{vu:2018} initially proposed a visually-grounded version of the Textual Entailment task where an image is augmented with textual premise and hypothesis. However, this task was refined by ~\shortciteA{xie:2019} to predict whether the image semantically entails the text, given image-sentence pairs, where the premise is defined by an image instead of a natural language sentence. Figure~\ref{fig:visualentail} illustrates the task, where the image as a \textit{premise} and a piece of text as \textit{hypothesis} are used by the Image Entailment model to predict whether the hypothesis is an \textit{entailment}, \textit{contradiction}, or \textit{neutral}.

\subsubsection{Image Entailment - Datasets}
\label{sssec:imentaildata}
The image entailment task is achieved using two different datasets. One dataset extends Natural Language Inference with Visually-grounded Natural Language Inference (V-SNLI)~\shortcite{vu:2018} while the other extends the Flickr30K dataset (see Section~\ref{para:flickr30k-data}) into a visual entailment dataset (SNLI-VE)\footnote{\url{https://github.com/necla-ml/SNLI-VE} \label{fnote: snli-ve-dataset-github-url}}~\shortcite{xie:2019}. Table~\ref{table:v-snli-dataset} and Table~\ref{table:snli-ve-dataset} presents the statistics and splits of these two datasets respectively.

\begin{table}[!ht]
\small
    \centering
    \begin{tabular}{l | c c c}
    \hline  %\toprule
    \rowcolor{teal!35}
         Split                          & Entailment    & Neutral       & Contradiction \\
    \hline \addlinespace[0.3em]    %\midrule
        Training                        & 182,167       & 181,515       & 181,938 \\
        Validation                      & 3,329         & 3,235         & 3,278 \\
        Test                            & 3,368         & 3,219         & 3,237 \\
        V-SNLI$_{\text{hard}}$ Test     & 1,058         & 1,068         & 1,135 \\
    \bottomrule
    \end{tabular}
    \caption{\label{table:v-snli-dataset} Splits of the V-SNLI dataset.}  %\colorbox{orange!45}{V-SNLI}
\end{table}

\begin{table}[!ht]
\small
    \centering
    \begin{tabular}{l | c c c c c }
    \hline  %\toprule
    \rowcolor{teal!35}
         Split          & Images    & Entailment    & Neutral   & Contradiction     & Vocab \\
    \hline \addlinespace[0.3em]    %\midrule
        Training        & 29,783    & 176,932       & 176,045   & 176,550           & 29,550 \\
        Validation      &1000       & 5,959         & 5,960     & 5,939             & 6,576 \\
        Test            & 1000      & 5,973         & 5,964     & 5,964             & 6,592 \\
    \bottomrule
    \end{tabular}
    \caption{\label{table:snli-ve-dataset} Splits of the SNLI-VE dataset.}
\end{table}

\subsubsection{Image Entailment - Evaluation Measures, Models, and Results}
\label{sssec:imageentailall}

In this section, we review the measures used to evaluate different models of \textit{Image Entailment} and the results obtained by them.

\paragraph{Evaluation Measures.} \textit{Image Entailment} task is evaluated using the Accuracy measure.

\paragraph{Models.} Two different models are created to approach the task of \textit{Image Entailment}. In Table~\ref{arcimageentail}, we present some exemplar architectures (refer to \textit{Combined} column) created to address the task. We also include a column that showcases the optimization techniques used to train those models.

\begin{table*}[!ht]
\small
  \centering
  \begin{tabular}{lccccc}
    \hline      %\toprule
    \rowcolor{teal!35}
    Approach & Image & Language & Combined & Optimizer & RL\\
    \hline \addlinespace[0.3em]     %\midrule
    ~\shortcite{vu:2018} & VGG & BiLSTM & V-BiMPM & ADAM & \xmark \\
    ~\shortcite{xie:2019} & ResNet-101 & GRU & EVE-Image & ADAM & \xmark\\
    \bottomrule
  \end{tabular}
  \caption{\label{arcimageentail} Exemplar \textit{Image Entailment} architectures.}
\end{table*}

\paragraph{Results.} The \textit{Image Entailment} models leverage both image and textual input representations to build an entailment pipeline. In Table~\ref{ressnlive}, Table~\ref{resvsnli}, and Table~\ref{resvsnlihard} we present results obtained with a subset of models that were built using the datasets presented in Section~\ref{sssec:imentaildata}.

\begin{table*}[!ht]
\small
  \centering
  \begin{tabular}{lcccc}
    \hline      %\toprule
    \rowcolor{teal!35}
    {Model} & Contradiction & Neutral & Entailment & Overall \\
    \hline \addlinespace[0.3em]     %\midrule
    Relation Network~\shortcite{santoro:2017} &  67.29 & 68.86 & 66.50 & 67.55\\
    Bottom-up~\shortcite{anderson:2017} & 70.52 & \textbf{70.96} & 65.23 & 68.90 \\
    Top-Down~\shortcite{anderson:2017} & 69.72 & 69.33 & 71.86 & 70.3 \\
    Hypothesis Only~\shortcite{gururangan:2018} &  67.60 & 67.71 & 64.83 & 66.71 \\
    %Image Captioning~\shortcite{choi:2018} &  66.25 & 70.69& 66.08 & 67.67 \\
    EVE-ROI~\shortcite{xie:2019} & 67.69 & 69.45 & \textbf{74.25} & 70.47 \\
    EVE-Image~\shortcite{xie:2019} & \textbf{71.56} & 70.52 & 71.39 & \textbf{71.16}\\
    \bottomrule
  \end{tabular}
  \caption{\label{ressnlive} Comparison of accuracies (\%) of different models on the SNLI-VE dataset.}
\end{table*}

\begin{table*}[!ht]
\small
  \centering
  \begin{tabular}{lcccc}
    \hline  %\toprule
    \rowcolor{teal!35}
    {Model} & Contradiction & Neutral & Entailment & Overall \\
    \hline \addlinespace[0.3em]     %\midrule
    Hypothesis Only~\shortcite{bowman:2015} &  66.29 & 66.36 & 72.65 & 68.49 \\
    LSTM (blind)~\shortcite{bowman:2015} &  79.7 & 76.79 & 87.71 & 81.49 \\
    V-LSTM~\shortcite{anderson:2017} &  71.39 & 68.06 & 87.14 & 75.70 \\
    BiMPM~\shortcite{wangbi:2017} & 86.25 & 82.79 & 90.03 & 86.41\\
    V-BiMPM~\shortcite{vu:2018} & \textbf{87.53} & \textbf{82.91} & \textbf{90.38} & \textbf{86.99} \\
    \bottomrule
  \end{tabular}
  \caption{\label{resvsnli} Comparison of accuracies (\%) of different models on the V-SNLI dataset.}
\end{table*}

\begin{table*}[!ht]
\small
  \centering
  \begin{tabular}{lcccc}
    \hline  %\toprule
    \rowcolor{teal!35}
    {Model} & Contradiction & Neutral & Entailment & Overall \\
    \hline \addlinespace[0.3em]     %\midrule
    Hypothesis Only~\shortcite{bowman:2015} &  25.29 & 20.22 & 31.28 & 25.57 \\
    LSTM (blind)~\shortcite{bowman:2015} &  60.79 & 50.19 & 72.12 & 60.99 \\
    V-LSTM~\shortcite{anderson:2017} &  46.34 & 32.02 & 69.09 & 49.03 \\
    BiMPM~\shortcite{wangbi:2017} & \textbf{77.62} & 59.36 & 80.43 & 72.55\\
    V-BiMPM~\shortcite{vu:2018} & 76.12 & \textbf{63.67} & \textbf{81.38} & \textbf{73.75} \\
    \bottomrule
  \end{tabular}
  \caption{\label{resvsnlihard} Comparison of accuracy (\%) scores of various models on V-SNLI$_{\text{hard}}$.}
\end{table*}

\subsubsection{Image Entailment - Discussion}
\label{sssec:imentailopenques}
The task of \textit{Image Entailment} was evaluated using two different datasets. Table~\ref{resvsnli} and Table~\ref{resvsnlihard} shows results obtained from V-SNLI in different settings. The approach proposed by~\shortciteA{vu:2018} that creates a visually grounded Bilateral Multi-Perspective Matching (BiMPM) model achieves the best result for the entailment task.

Similarly, evaluations conducted with SNLI-VE dataset (cf. Table~\ref{ressnlive}) show that the Explainable Visual Entailment (EVE) approach proposed by~\shortciteA{xie:2019} achieves the best overall result.
\subsubsection{Video Entailment - Introduction}
\label{sssec:videoentailintro}
\begin{figure}[!htb]
    \centering
        \includegraphics[width=\textwidth]{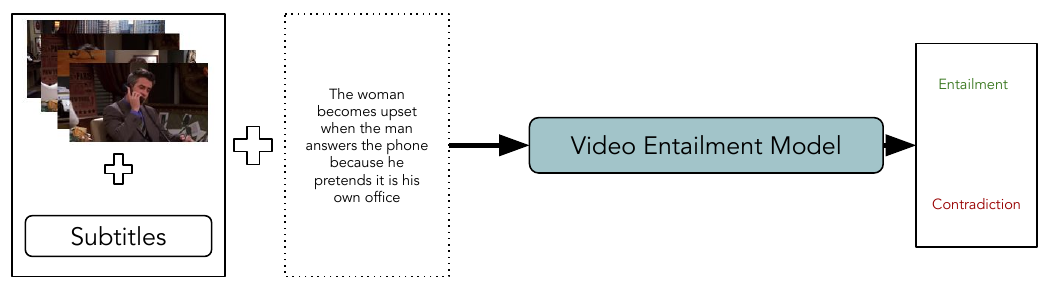}
    \caption{Given a video along with aligned subtitles as \textit{premise} and a paired natural language text as \textit{hypothesis}, the goal of a Video Entailment Model is to predict whether the hypothesis is an entailment or contradiction, by understanding the evidence(s) observed in the video. Example modified from~\shortciteA{liu-violin:2020}.} \label{fig:videoentail}
\end{figure}

Video entailment~\shortcite{liu-violin:2020} aims to infer whether the natural language hypothesis is entailed or contradicted when given a video clip aligned with the subtitles information. The video contains diverse temporal dynamics, event shifts, and social interactions. Figure~\ref{fig:videoentail} illustrates the task: given a video clip with aligned subtitles as premise and a natural language hypothesis based on the video content, a video entailment model needs to infer whether the hypothesis is entailed or contradicted by the given video clip.

\subsubsection{Video Entailment - Datasets}
\label{sssec:videoentaildata}
The Video Entailment task is proposed by~\shortciteA{liu-violin:2020}, with the introduction of a large-scale dataset called as VIdeO-and-Language INference (VIOLIN)\footnote{\url{https://github.com/jimmy646/violin}\label{fnote: violin-dataset-github-url}}. Detailed statistics of the dataset is presented in Table~\ref{table:violin-dataset-stats}.

\begin{table}[!ht]
\small
\begin{center}
\begin{tabular}{l c c c c c c }
\hline  %\toprule
\rowcolor{teal!35}
Video Source            &Num. of        & Num. of       &Avg. Clip      & Avg. Pos.     & Avg. Neg.     & Avg. Sub- \\
\rowcolor{teal!35}
(TV Show/Movie Clips)  & Episodes       & Clips         & Len           & Stmnt Len     & Stmnt Len      & Title Len \\
\hline \addlinespace[0.3em]    %\midrule
Friends                 & 234           & 2,676         & 32.89s        & 17.94         & 17.85         & 72.80\\
Desperate Housewives    & 180           & 3,466         & 32.56s        & 17.79         & 17.81         & 69.19\\
How I Met Your Mother   & 207           & 1,944         & 31.64s        & 18.08         & 18.06         & 76.78\\
Modern Family           & 210           & 1,917         & 32.04s        & 18.52         & 18.20         & 98.50\\
MovieClips              & 5,885         & 5,885         & 40.00s        & 17.79         & 17.81         & 69.20\\
\hline
All                     & 6,716         & 15,887        & 35.20s        & 18.10         & 18.04         & 76.40\\
\bottomrule
\end{tabular}
\end{center}
\vspace{-2mm}
\caption{Statistics of different video sources in the VIOLIN dataset.}
\label{table:violin-dataset-stats}
\vspace{-2mm}
\end{table}

For training and model evaluation purposes, the VIOLIN dataset is split into training, validation, and test splits in the ratio of 8:1:1. The exact number of triplet instances in each of the splits is shown in Table~\ref{table:violin-dataset-split}.

\begin{table}[!ht]
\small
    \centering
    \begin{tabular}{l | c c c c}
    \hline  %\toprule
    \rowcolor{teal!35}
                                & Number of     &Number of          &Number of\\
    \rowcolor{teal!35}
    \multirow{-2}{*}{Split}     & Videos (V)    &Hypotheses (H)     &Triplets (V, S, H)\\
    \hline \addlinespace[0.3em]    %\midrule
        Training                & 12,687        & 76,122            & 76,122 \\
        Validation              & 1,600         & 9,600             & 9,600 \\
        Testing                 & 1,600         & 9,600             & 9,600 \\
    \hline \addlinespace[0.3em]    %\midrule
        Total                   &15,887         & 95,322            & 95,322 \\
    \bottomrule
    \end{tabular}
    \caption{\label{table:violin-dataset-split} Splits of the VIOLIN dataset.} (V: Video, S: Subtitle, H: Hypothesis)
\end{table}

\subsubsection{Video Entailment - Evaluation Measures, Models, and Results}
\label{sssec:videoentailall}

In this section, we present the evaluation measures, models, and results achieved with various architectures introduced for solving the \textit{Video Entailment} task.

\paragraph{Evaluation Measures.} The \textit{Video Entailment} models are evaluated using Accuracy.

\paragraph{Models.} Very few models have been created to approach the task of \textit{Video Entailment}. The variation of the \textit{Video Entailment} models include the usage of different type of textual content such as subtitles, statements, etc. In Table~\ref{arcvideoentail}, we present some exemplar architectures (refer to \textit{Combined} column) created to address the task by integrating both video and language inputs. We also include a column that showcases the optimization techniques used to train those models.

\begin{table*}[!ht]
\small
  \centering
  \begin{tabular}{lcccccc}
    \hline  %\toprule
    \rowcolor{teal!35}
    {Approach} & Video & Frame & Language & Combined & Optimizer & RL \\ 
    \hline \addlinespace[0.3em]     %\midrule
    \shortcite{liu-violin:2020} & - & Detection Feat & BERT & SSV & ADAM & \xmark \\
    \bottomrule
  \end{tabular}
  \caption{\label{arcvideoentail} Exemplar \textit{Video Entailment} architectures. SSV - Statement+Subtitles+Visual.}
\end{table*}

\paragraph{Results.} Few models which have been designed to approach the task of \textit{Video Entailment} use different types of textual content aligned with video. In Table~\ref{resviolin} we present results obtained with subset of models built using the VIOLIN dataset presented in Section~\ref{sssec:videoentaildata}. For building textual or visual representations, models such as \textbf{SSV} has used pretrained vision and language integration models such as LXMERT~\shortcite{tanlxmert:2019}.

\begin{table*}[!ht]
\small
  \centering
  \begin{tabular}{lccc}
    \hline  %\toprule
    \rowcolor{teal!35}
    {Model} & Visual & Text & Accuracy \\
    \hline \addlinespace[0.3em]     %\midrule
    Statement~\shortcite{liu-violin:2020} & - & BERT & 54.20 \\
    Statement+Visual~\shortcite{liu-violin:2020} & Detection Feat & BERT & 59.45 \\
    Statement+Subtitles~\shortcite{liu-violin:2020} & - & BERT & 66.05 \\
    SSV~\shortcite{tanlxmert:2019} & LXMERT & LXMERT & 66.25 \\
    SSV~\shortcite{liu-violin:2020} & Detection Feat & BERT & \textbf{67.84} \\
    \bottomrule
  \end{tabular}
  \caption{\label{resviolin} Comparison of accuracies (\%) of different methods on the VIOLIN dataset.}
\end{table*}

\subsubsection{Video Entailment - Discussion}
\label{sssec:videoentailopenques}
The task of \textit{Video Entailment} was evaluated using the VIOLIN dataset and the recently proposed method by~\shortciteA{liu-violin:2020} has shown that using multi-source information arising from different types of data such as Statements, Subtitles, and Visual features are useful for building a robust model. In addition, textual features generated using contextualized word embedding models are effective as well.

\section{Visual Dialog}
\label{sec:visualdialog}
In this section, we explore the task of \textit{Visual Dialog}. The objective of visual dialog is different from the previously discussed tasks and involves a complex interaction between a human and an artificial agent.
\subsection{Image Dialog}
\label{ssec:imagedialog}

In the following, we describe the setting of \textit{Visual Dialog} where an image is used as the visual input.
\subsubsection{Image Dialog - Introduction}
\label{sssec:imagedialogintro}
\begin{figure}[!htb]
    \centering
        \includegraphics[width=0.85\textwidth]{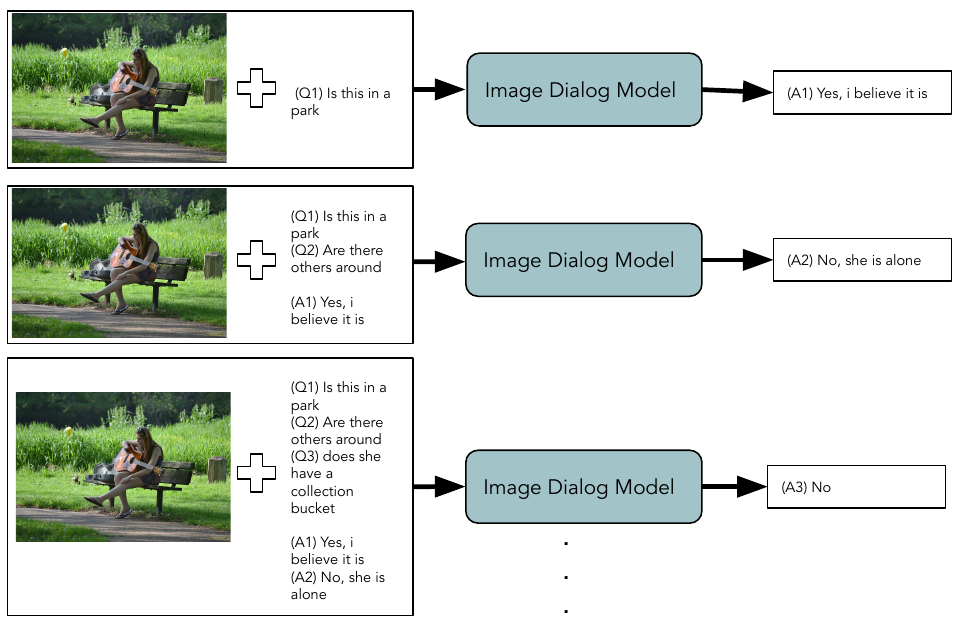}
    \caption{Given an \textit{image}, \textit{question} and the \textit{dialog history}, an Image Dialog Model generates an answer based on these multimodal information.}\label{fig:visualdialog}
\end{figure}

The goal of the \textit{image dialog} task is to create AI agents that can hold dialog with humans in a natural language of choice about a visual content~\shortcite{das:2017}, represented by an image. To be more specific, given an image, a history of dialogs, and a question about the image, the goal of an AI agent is to ground the question in the image, infer the context from the history, and then answer the question accurately. However, this problem can also be construed as a task where the goal of the AI system is to locate an unknown object in the image by asking a sequence of questions~\shortcite{vries:2017} or to hold natural-sounding conversations about a shared image~\shortcite{nasrin:2017}. In Figure~\ref{fig:visualdialog}, we provide a visual depiction to illustrate the said task.

Further, a standard agent can be extended to have a question and answer bot cooperating with each other for guessing images~\shortcite{dasco:2017}. To counter generic responses in dialog generation, knowledge transfer from dialog generation was explored with a discriminative dialog module trained to rank a list of candidate human responses~\shortcite{lu:2017}. However, other approaches constrained themselves to specific domains and proposed end-to-end optimization schemes~\shortcite{strub:2017}. \shortciteA{seo:2017} introduced attentive memory that exploits visual attention in the past to resolve the current reference. Recently, reinforcement learning and Generative Adversarial Networks (GANs) were also used to generate more human-like responses to questions in the image-based dialog~\shortcite{wu:2018}. Dialog can also be seen from the perspective of a system which asks questions, and demonstrates how a visual dialog can be generated from discriminative question generation and answering~\shortcite{jain:2018}. Furthermore, co-reference resolution was also investigated ~\shortcite{kottur:2018} to bridge the gap between nouns and pronouns with the usage of modules that form explicit and grounded co-reference resolution at word-level.

Recently, a novel attention mechanism called recursive visual attention~\shortcite{niu:2019} was proposed to resolve visual co-reference for visual dialog by browsing the dialog history. Another approach~\shortcite{zhengdialog:2019} formalized the task as inference in a graphical model with partially observed nodes and unknown graph structures, i.e., relations in dialog. Further, ~\shortciteA{guo:2019} extended one-stage solution to a two-stage solution by building an image-question-answer synergistic network to value the role of the answer for precise visual dialog. Other novel approaches~\shortcite{shekharbeyond:2019} were also designed where a visually-grounded encoder was employed to synergize between guessing and asking questions. Further, a cooperative learning regime was followed to improve the accuracy.

\subsubsection{Image Dialog - Datasets}
\label{sssec:imdialogdata}
For addressing the task of image dialog several datasets have been created. In the following, we elaborate each of them separately.

\paragraph{VisDial.}\label{para:visdialdata}For Image Dialog, there exists two versions of this dataset, VisDial v0.9 and VisDial 1.0\footnote{\url{https://visualdialog.org/data} \label{fnote: visual-dialog-dataset-url}}~\shortcite{das:2017}. VisDial was created using the MSCOCO dataset. For VisDial v0.9, splits are divided only into the training and validation set. Table~\ref{table:visual-dialog0.9-dataset} and Table~\ref{table:visual-dialog1.0-dataset} present details about the splits of VisDial v0.9 and VisDial v1.0 respectively.

\begin{table}[!ht]
\small
    \centering
    \begin{tabular}{l | c c c c}
    \hline  %\toprule
    \rowcolor{teal!35}
        Split       & Images    & Questions     & Answers   & Dialog Turns \\
    \hline \addlinespace[0.3em]    %\midrule
        Training    & 82,783    & 827,830       & 827,830   & 10  \\
        Validation  & 40,504    & 405,040       & 405,040   & 10 \\
        Test & -    & -       & -   & - \\
    \bottomrule
    \end{tabular}
    \caption{\label{table:visual-dialog0.9-dataset} Splits of the VisDial v0.9 dataset.}
\end{table}

\begin{table}[!ht]
\small
    \centering
    \begin{tabular}{l | c c c c}
    \hline  %\toprule
    \rowcolor{teal!35}
        Split       & Images    & Questions     & Answers       & Dialog Turns \\
    \hline \addlinespace[0.3em]    %\midrule
        Training    & 123,287   & 1,232,870     & 1,232,870     & 10  \\
        Validation  & 2,064     & 20,640        & 20,640        & 10 \\
        Test        & 8,000     & 80,000        & 80,000        & 1 \\
    \bottomrule
    \end{tabular}
    \caption{\label{table:visual-dialog1.0-dataset} Splits of the VisDial v1.0 dataset.}
\end{table}

\paragraph{CLEVR-Dialog.}\label{para:clevrdialogdata}The CLEVR-Dialog\footnote{\url{https://github.com/satwikkottur/clevr-dialog} \label{fnote: clevr-dialog-dataset-url}}~\shortcite{kottur:2019} dataset was developed for studying multi-round reasoning in visual dialog. The dialog grammar is grounded in the scene graphs of the CLEVR dataset (Section~\ref{para:clevrdata}), originally developed for reasoning about images. Table~\ref{table:clevr-dialog-dataset} provides statistics of the dataset, while Table~\ref{table:clevr-dialog-dataset-split} shows the dataset splits. 

\begin{table}[!ht]
\small
    \centering
    \begin{tabular}{c c c c c c c c c}
    \hline  %\toprule
    \rowcolor{teal!35}
        CLEVR  & Total      & Total         & Unique        & Unique    & Vocabulary   & Dialog     & Mean Ques.\\
    \rowcolor{teal!35}
        Images & Dialogs    & Questions     & Questions     & Answers   & Size         & Turns      & Length    \\
    \hline \addlinespace[0.3em]    %\midrule
        85k     & 425k     & 4.25M       & 73k               & 29       & 125           & 10        & 10.6\\
    \bottomrule
    \end{tabular}
    \caption{\label{table:clevr-dialog-dataset} Statistics of the CLEVR-Dialog dataset.}
\end{table}

\begin{table}[!ht]
\small
    \centering
    \begin{tabular}{l | c c c c}
    \hline  %\toprule
    \rowcolor{teal!35}
        Split       & Images    & Q\&A Pairs    & Instances     & Dialog Rounds\\
    \hline \addlinespace[0.3em]     %\midrule
        Training    & 70,000    & 3.5M          & 5             & 10    \\
        Validation  & 15,000    & 0.75M         & 5             & 10    \\
        Test & -    &-         & -            & -    \\
    \bottomrule
    \end{tabular}
    \caption{\label{table:clevr-dialog-dataset-split} Splits of the CLEVR-Dialog dataset.}
\end{table}

\subsubsection{Image Dialog - Evaluation Measures, Models and Results}
\label{sssec:imagedialogall}

In this section, we review the measures used to evaluate different models of \textit{Image Dialog} and the results achieved by these models.

\paragraph{Evaluation Measures.} The \textit{Image Dialog} models are evaluated using the \textit{Retrieval} metrics that have been discussed in Section~\ref{sssec:idgall}.

\paragraph{Models.} The models created to approach the \textit{Image Dialog} task continuously process a stream of images and textual dialog information. In Table~\ref{arcimagedialog}, we present some exemplar architectures (refer to \textit{Combined} column) designed to integrate image and textual dialog to address the task.

\begin{table*}[!ht]
\small
  \centering
  \begin{tabular}{lcccccc}
    \hline      %\toprule
    \rowcolor{teal!35}
    Approach & Image & Language & Combined & RL\\
    \hline \addlinespace[0.3em]     %\midrule
    ~\shortcite{das:2017} & VGG & LSTM & MemoryNetwork & \xmark\\
    ~\shortcite{lu:2017} & VGG & LSTM & HCIAE-NP-ATT &  \xmark \\
    ~\shortcite{seo:2017} & VGG & LSTM & AMEM & \xmark\\
    ~\shortcite{jain:2018} & VGG &  LSTM & SF &  \xmark \\
    ~\shortcite{kottur:2018} & ResNet-152 & LSTM & CorefNMN & \xmark \\
    ~\shortcite{wu:2018} & VGG & LSTM & CoAtt-GAN & \cmark \\
    ~\shortcite{niu:2019} & ResNet-152 & LSTM & RvA & \xmark\\
     ~\shortcite{zhengdialog:2019} & VGG & LSTM & GNN & \xmark\\
    ~\shortcite{guo:2019} & ResNet-101 & LSTM & Synergistic & \xmark \\
    \bottomrule
  \end{tabular}
  \caption{\label{arcimagedialog} Exemplar \textit{Image Dialog} Architectures (Discriminative and Generative).}
\end{table*}

\paragraph{Results.} Models that are created to solve the task of \textit{Image Dialog} effectively comprehends the complexity of the task. Several approaches are used to build the models with different versions of the same dataset. However, few approaches share some commonalities such as usage of Memory Networks~\shortcite{sukhbaatarend:2015}. Table~\ref{resvisv1dialogdisc} and  Table~\ref{resvisv1dialoggen} presents the results obtained with a subset of both  \textbf{discriminative} and \textbf{generative} models built using the ``VisDial0.9'' dataset. While Table~\ref{resvisv2dialogdisc} presents the results obtained only with a subset of \textbf{generative} models built using the ``VisDial1.0'' dataset presented earlier in Section~\ref{sssec:imdialogdata}.

\begin{table*}[!ht]
\small
  \centering
  \begin{tabular}{lcccccc}
    \hline  %\toprule
    \rowcolor{teal!35}
    {Model} & MRR & R@1 & R@5 & R@10 & Mean\\
    \hline \addlinespace[0.3em] %\midrule
    LF~\shortcite{das:2017} & 0.5807 & 43.82 & 74.68 & 84.07 & 5.78 \\
    HRE~\shortcite{das:2017} & 0.5846 & 44.67 & 74.50 & 84.22 & 5.72 \\
    HREA~\shortcite{das:2017} & 0.5868 & 44.82 &  74.81 &  84.36 & 5.66 \\
    MN~\shortcite{das:2017} & 0.5965 & 45.55 & 76.22 & 85.37 & 5.46 \\
    \midrule
    HCIAE-NP-ATT~\shortcite{lu:2017} & 0.6222 & 48.48 & 78.75 & 87.59 & 4.81 \\
    AMEM~\shortcite{seo:2017} & 0.6227 & 48.53 & 78.66 &  87.43 & 4.86 \\
    CoAtt~\shortcite{wu:2018} & 0.6398 & 50.29 & 80.71 & 88.81 & 4.47 \\
    SF~\shortcite{jain:2018} & 0.6242 &  48.55 & 78.96 & 87.75 &  4.70 \\
    SCA~\shortcite{wu:2018} & 0.6398 &  50.29 & 80.71 &  88.81 & 4.47 \\
    CorefNMN~\shortcite{kottur:2018} & 0.641 &  50.92 & 80.18 & 88.81 & 4.45 \\
    \midrule
    GNN~\shortcite{zhengdialog:2019} & 0.6285 & 48.95 & 79.65 & 88.36 & 4.57 \\
    RvA~\shortcite{niu:2019} & \textbf{0.6634} & \textbf{52.71} & \textbf{82.97} & \textbf{90.73} & \textbf{3.93} \\
    \bottomrule
  \end{tabular}
  \caption{\label{resvisv1dialogdisc} Results of different \textbf{discriminative models} on the validation split of the VisDial v0.9 dataset.}
\end{table*}

\begin{table*}[!ht]
\small
  \centering
  \begin{tabular}{lcccccc}
    \hline      %\toprule
    \rowcolor{teal!35}
    {Model} & MRR & R@1 & R@5 & R@10 & Mean\\
    \hline \addlinespace[0.3em]     %\midrule
    LF~\shortcite{das:2017} & 0.5199 & 41.83 & 61.78 & 67.59 & 17.07 \\
    HRE~\shortcite{das:2017} & 0.5237 & 42.29 & 62.18 & 67.92 & 17.07 \\
    HREA~\shortcite{das:2017} & 0.5242 & 42.28 &  62.33 &  68.17 & 16.79 \\
    MN~\shortcite{das:2017} & 0.5259 & 42.29 & 62.85 & 68.88 & 17.06 \\
    \midrule
    HCIAE-NP-ATT~\shortcite{lu:2017} & 0.5386 & 44.06 & 63.55 & 69.24 & 16.01 \\
    CorefNMN~\shortcite{kottur:2018} & 0.535 &  43.66 & 63.54 & 69.93 & 15.69 \\
    CoAtt~\shortcite{wu:2018} & 0.5411 & 44.32 & 63.82 & 69.75 & 16.47 \\
    CoAtt-RL~\shortcite{wu:2018} & \textbf{0.5578} & \textbf{46.10} & \textbf{65.69} & 71.74 & 14.43 \\
    \midrule
    RvA~\shortcite{niu:2019} & 0.5543 & 45.37 & 65.27 & \textbf{72.97} & \textbf{10.71} \\
    \bottomrule
  \end{tabular}
  \caption{\label{resvisv1dialoggen} Results of different \textbf{generative models} on the validation split of the VisDial v0.9 dataset.}
\end{table*}

\subsubsection{Image Dialog - Discussion}
\label{sssec:imagedialogopenques}
For the \textit{Image Dialog} task, two versions of the same dataset were used for evaluation. Similar approaches were used for the evaluation of both datasets with retrieval metrics. Nevertheless, the methods that achieve state-of-the-art performance on both datasets differ. Among the generative and discriminative methods on VisDial v0.9 dataset, the Recursive Visual Attention (RvA) approach proposed by~\shortciteA{niu:2019} achieves the best result. RvA refines the visual attention recursively by browsing through the dialog history until the agent has sufficient confidence in its visual co-reference resolution. This has also been shown to generate interpretable attention maps without additional annotations.

For the VisDial v1.0 dataset, the results presented in Table~\ref{resvisv2dialogdisc} show that Synergistic-ensemble by~\shortciteA{guo:2019} outperform RvA.

\begin{table*}[!ht]
\small
  \centering
  \begin{tabular}{lccccccc}
    \hline  %\toprule
    \rowcolor{teal!35}
    {Model} & MRR & R@1 & R@5 & R@10 & Mean & NDCG\\
    \hline \addlinespace[0.3em]     %\midrule
    LF~\shortcite{das:2017} & 0.5542 & 40.95 & 72.45 & 82.83 & 5.95 & 0.4531 \\
    LF-att~\shortcite{das:2017} & 0.5707 & 42.08 & 74.83 & 85.05 & 5.59 & 0.4976 \\
    HRE~\shortcite{das:2017} & 0.5416 & 39.93 & 70.45 & 81.50 & 6.41 & 0.4546 \\
    MN~\shortcite{das:2017} & 0.5549 & 40.98 & 72.30 & 83.30 & 5.92 & 0.4750  \\
    MN-att~\shortcite{das:2017} & 0.5690 & 42.43 & 74.00 & 84.35 & 5.59 & 0.4958  \\
    CorefNMN~\shortcite{kottur:2018} & 0.615 &  47.55 & 78.10 & 88.80 & 4.40 & 0.547 \\
    \midrule
    GNN~\shortcite{zhengdialog:2019} & 0.6137 & 47.33 & 77.98 & 87.83 & 4.57 & 0.5282 \\
    RvA~\shortcite{niu:2019} & 0.6303  & 49.03 & 80.40 & 89.83 & 4.18 & 0.5559 \\
    Synergistic-ensemble~\shortcite{guo:2019} & \textbf{0.6342} & \textbf{49.30} & \textbf{80.77} & \textbf{90.68} & \textbf{3.97} & \textbf{0.5788} \\
    \bottomrule
  \end{tabular}
  \caption{\label{resvisv2dialogdisc} Results of different \textbf{discriminative models} on the test-standard split of the VisDial v1.0 dataset.}
\end{table*}

\subsection{Video Dialog}
\label{ssec:videodialog}
In this part, we present details about the \textit{Visual Dialog} task in which a video is used as the visual input and a conversational chat with humans about the visual content is expected.
\subsubsection{Video Dialog - Introduction}
\label{sssec:videodialogintro}
\begin{figure}[!htb]
    \centering
        \includegraphics[width=0.85\textwidth]{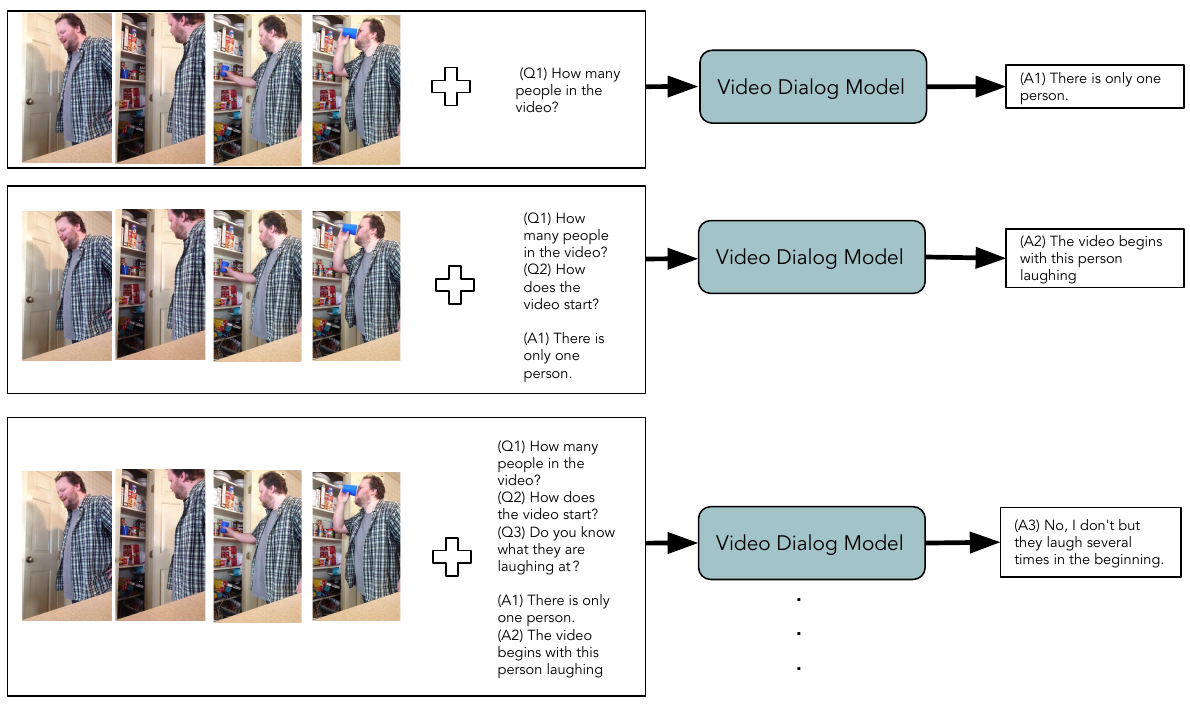}
    \caption{Given a \textit{video} (represented as a sequence of frames), a \textit{question}, and the \textit{dialog history}, a Video Dialog Model generates answers based on these information.}\label{fig:videodialog}
\end{figure}

The aim of video dialog is to leverage scene information containing both audio (which can be transcribed as subtitles) and visual frames to hold a dialog (i.e., an exchange) with humans in a natural language of choice about the multimedia content~\shortcite{alamriaudio:2019,alamri:2019}. A successful system is expected to ground concepts from the question in the video while leveraging contextual cues from the dialog history. Figure~\ref{fig:videodialog} illustrates the video dialog task.

Several approaches have been proposed to address the task, where initially multimodal attention-based video description features were used to improve dialog~\shortcite{hori:2019}. Further, a novel baseline~\shortcite{schwartz:2019} analyzed components such as data representation, extraction, attention, and answer generation in order to show that there can be relative improvements as compared to other approaches. 

\subsubsection{Video Dialog - Datasets}
\label{sssec:videodialogdata}
Audio Visual Scene-Aware Dialog (AVSD)\footnote{\url{https://video-dialog.com} \label{fnote: avsd-dataset-url}}~\shortcite{alamriaudio:2019} was created for the Scene-Aware Dialog Challenge, in which the agent grounds its responses on the dynamic scene, the audio, and the history (previous rounds) of the dialog. Table~\ref{table:avsd-dataset} presents some statistics and the splits of the AVSD dataset.
\begin{table}[!ht]
\small
    \centering
    \begin{tabular}{l | c c c}
    \hline  %\toprule
    \rowcolor{teal!35}
        Split       & Dialogs   & Turns         & Words \\
    \hline \addlinespace[0.3em]    %\midrule
        Training    & 7,985     & 123,480       & 1,163,969 \\
        Validation  & 1,863     & 14,680        & 138,314 \\
        Test        & 1,968     & 14,660        & 138,790 \\
    \bottomrule
    \end{tabular}
    \caption{\label{table:avsd-dataset} Splits of the AVSD dataset.}
\end{table}

\subsubsection{Video Dialog - Evaluation Measures, Models, and Results}
\label{sssec:videodialogall}

In this section, we review the evaluation measures used to benchmark different models of \textit{Video Dialog} and the results obtained by these models.

\paragraph{Evaluation Measures.} The \textit{Video Dialog} models are evaluated using the ``Retrieval metrics'' discussed in Section~\ref{sssec:idgall}.

\paragraph{Models.} Only a couple of models have been proposed so far to approach the task of \textit{Video Dialog}. These models aim to capture the temporal aspect of a video and incorporate it in the textual dialog.  In Table~\ref{arcvideodialog}, we present some exemplar architectures (refer to \textit{Combined} column) designed to address the task by integrating both video and language inputs. We also include a column that showcases the optimization techniques used to train those models. 

\begin{table*}[!ht]
\small
  \centering
  \begin{tabular}{lcccccc}
    \hline  %\toprule
    \rowcolor{teal!35}
     {Approach} & Video & Frame & Language & Combined & Optimizer & RL\\
    \hline \addlinespace[0.3em]     %\midrule
     ~\shortcite{hori:2019} & I3D & VGG & LSTM & MultimodalAtt & ADAM & \xmark \\
    ~\shortcite{schwartz:2019} & I3D & VGG & LSTM & i3d-rgb-spatial-10 & ADAM & \xmark \\
    \bottomrule
  \end{tabular}
  \caption{\label{arcvideodialog} Exemplar \textit{Video Dialog} architectures.}
\end{table*}

\paragraph{Results.}  As discussed earlier only few models have been created to address the task of \textit{Video Dialog}. In Table~\ref{resvideodialog} we present the results obtained with those models built using the ``AVSD'' dataset presented earlier in Section~\ref{sssec:videodialogdata}.

\begin{table*}[!ht]
\small
  \centering
  \begin{tabular}{lcccccc}
    \hline  %\toprule
    \rowcolor{teal!35}
    {Model} & B-1 & B-2 & B-3 & B-4 & METEOR & CIDEr \\
    \hline \addlinespace[0.3em]     %\midrule
    Att-base~\shortcite{hori:2019} & 0.273 & 0.173 & 0.117 & 0.084 & 0.117 & 0.766 \\
    Att-weightshare~\shortcite{schwartz:2019} & \textbf{0.293} & \textbf{0.191} & 0.133 & 0.097 & 0.127 & 0.923 \\
    i3d-rgb-spatial-10~\shortcite{schwartz:2019} & 0.290 & 0.190 & \textbf{0.133} & \textbf{0.097} & 0.127 & 0.928 \\
    Att-base-beam~\shortcite{schwartz:2019} & 0.285 & 0.187 & 0.131 & 0.096 & \textbf{0.128} & \textbf{0.941} \\
    \bottomrule
  \end{tabular}
  \caption{\label{resvideodialog} Results of different models on the ``AVSD'' dataset.}
\end{table*}

\subsubsection{Video Dialog - Discussion}
\label{sssec:videodialogopenques}
The \textit{Video Dialog} task is evaluated with the AVSD dataset. Different strategies have been explored to fuse the language and video features to create a strong baseline. In particular, the approach proposed by~\shortciteA{schwartz:2019}, which uses beam search and the attention mechanism (i.e., Att-base-beam) over different modalities, outperforms other baseline methods.

\section{Multimodal Machine Translation}
\label{sec:mmt}
In this section, we explore the task of \textit{Multimodal Machine Translation} (MMT). The goal of this task is to translate natural language sentences that describe visual content (e.g. image) in a source language into a target language by taking the visual content as an additional input to the source language sentences.
\subsection{Machine Translation with Image}
\label{ssec:mmtimagetask}
In the following, we elaborate on the \textit{Multimodal Machine Translation} task by considering image as the only visual input.

\subsubsection{Machine Translation with Image - Introduction}
\label{sssec:mmtimagetaskintro}

\begin{figure}[!htb]
    \centering
        \includegraphics[width=\textwidth]{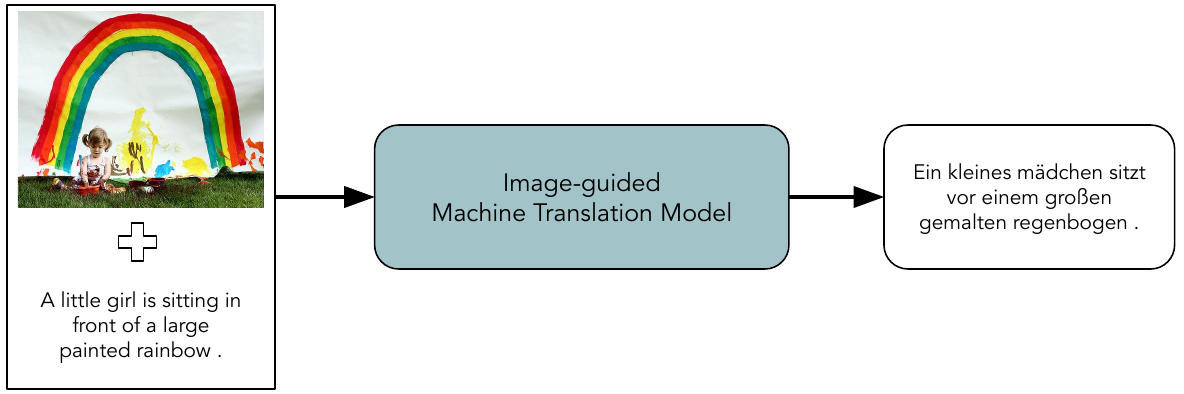}
    \caption{Given an \textit{image} and its \textit{description} in a source language (e.g. En), an Image-guided Machine Translation model produces a description in a target language (e.g. De).}\label{fig:mmtbase-img}
\end{figure}

The aim of MMT~\shortcite{specia:2016,hitschler:2016,elliott:2017,barrault:2018} is to translate sentences, that describe an image, in a source language into equivalent sentences in a target language. However, for any given image the description can be written in different source languages, resulting in multiple source language descriptions. This situation opens up the possibility to propose different variants of the MMT task. The first variant is a \textit{single source translation} task, in which the image description in a single source language is translated to the target language with additional cues from the corresponding image. Figure~\ref{fig:mmtbase-img} depicts this variant where an image is accompanied with its description in English and needs to be translated by the model into a description in German.

\begin{figure}[!htb]
    \centering
        \includegraphics[width=\textwidth]{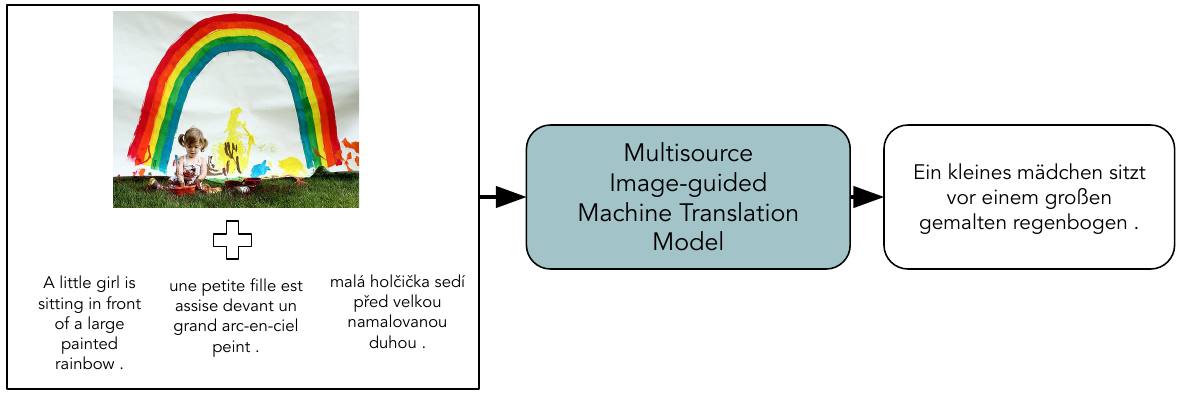}
    \caption{Given an \textit{image} and its \textit{description} in multiple source languages (e.g. en, fr, cs), a Multisource Image-guided Machine Translation model produces a description in a target language (e.g. de).}\label{fig:mmtmultisource}
\end{figure}

The second variant is a target language description generation task with additional source language cues, i.e., multiple source language descriptions of the same image termed as \textit{multisource MMT}. Figure~\ref{fig:mmtmultisource} illustrates this variant, where an image is accompanied with its descriptions in English (en), French (fr), and Czech (cs), that are all used to generate the German (de) translation.

Different approaches have been proposed to handle single source MMT by associating visual and textual features with multimodal attention~\shortcite{huangmt:2016}. Further, a novel approach where a doubly-attentive decoder incorporated visual features to bridge the gap between image description and translation was proposed~\shortcite{calixto:2017}. In a similar vein, global visual features were incorporated in an attention-based multimodal NMT~\shortcite{calixtoglob:2017}. This is achieved by attending to source-language words and parts of an image independently by means of two separate attention mechanisms.

MMT task can also be solved using two sub-tasks: learning to translate, and learning visually grounded representations~\shortcite{elliottimi:2017}, both combined in a multi-task learning framework. Further, an advanced multimodal compact bilinear pooling method~\shortcite{delbrouckemnlp:2017,delbrouck:2017} has also been used for MMT in which the outer product of two vectors combines the attention features of the two modalities. Another model~\shortcite{zhouva:2018} used a shared visual-language embedding and a translator for learning. This joint model leverages a visual attention grounding mechanism that links the visual semantics with the corresponding textual semantics. Due to the presence of large multimodal data on the web, noisy image captions have also been tried for MMT~\shortcite{schamoni:2018}. A latent variable model~\shortcite{calixto:2018} has also been attempted in which the latent variable can be seen as a multimodal stochastic embedding of an image and its description in a foreign language.

MMT models have also been used in an adversarial setting. \shortciteA{elliott:2018} found that even in the presence of visual features from unrelated images there is no significant performance degradation. Due to the recent success of unsupervised machine translation~\shortcite{lample:2018}, there is also a growing interest in extending it for unsupervised MMT~\shortcite{su:2018}. Other studies~\shortcite{caglayan:2019} have reduced criticism of MMT by showing that under the limited textual context, MMT models are capable of leveraging the visual input to generate better translations. Regarding multisource models, \shortciteA{libovicky:2017} explored MMT using neural multi-source sequence-to-sequence learning.

\subsubsection{Machine Translation with Image - Datasets}
\label{sssec:mmtimagedata}
The main dataset used with the models above (Section \ref{sssec:mmtimagetaskintro}) is the Multi30k-MMT\footnote{\url{https://www.statmt.org/wmt18/multimodal-task.html} \label{fnote: multi30k-mmt-dataset-url}} dataset~\shortcite{barrault:2018}, extended using the Flickr30k dataset. Along with English, it contains human translated German, French, and Czech language sentences. The splits of this dataset can be found in Table~\ref{table:multi30k-mmt-dataset}.

\begin{table}[!ht]
\small
    \centering
    \begin{tabular}{l | c c}
    \hline  %\toprule
    \rowcolor{teal!35}
        Split       & Images        & Captions \\
    \hline \addlinespace[0.3em] %\midrule
        Training    & 29,000        &  29,000 \\
        Validation  & 1,014         & 1,014 \\
        Test        & 1,000         & 1,000 \\
    \bottomrule
    \end{tabular}
    \caption{\label{table:multi30k-mmt-dataset} Splits of Multi30k-MMT for English, German, French, and Czech.}
\end{table}

\subsubsection{Machine Translation with Image - Evaluation Measures, Models, and Results}
\label{sssec:mmtimageall}

In this section, we review the evaluation measures used to benchmark different models of \textit{Machine Translation with Image} and the results obtained by these models.

\paragraph{Evaluation Measures.} To evaluate \textit{Machine Translation with Image} models, the ``Retrieval metrics'' presented in the Section~\ref{sssec:idgall} are used.

\paragraph{Models.} Several models have been created for the task of \textit{Machine Translation with Image}. The aim of these models is to tackle translation using either a single or multiple language textual sources along with an image. In Table~\ref{arcimagetranslation}, we present some exemplar architectures (refer to \textit{Combined} column) which integrate both image and language to address the task. We also include an ``Optimizer'' column that indicates the optimization techniques used to train those models.

\begin{table*}[!ht]
\small
  \centering
  \begin{tabular}{lccccc}
    \hline  %\toprule
    \rowcolor{teal!35}
    Approach & Image & Language & Combined & Optimizer & RL\\
    \hline \addlinespace[0.3em]     %\midrule
    ~\shortcite{calixto:2017} & ResNet-50 & BiGRU & DoubleAtt & Adadelta & \xmark \\                    
    ~\shortcite{calixtoglob:2017} & VGG & BiGRU & GVF & Adadelta & \xmark \\
    ~\shortcite{elliottimi:2017} & Inception-V3* & BiGRU & Imagination & ADAM & \xmark \\
    ~\shortcite{caglayanlium:2017} & ResNet-50 & BiGRU & Lium-cvc-ensemble & ADAM & \xmark \\
    ~\shortcite{calixto:2018} & ResNet-50 & BiGRU & VMMT$_{\text{F}}$ & ADAM & \xmark \\
    ~\shortcite{helcl:2018} & ResNet-50 & LSTM & CUNI-ensemble & ADAM & \xmark \\                 
    \bottomrule
  \end{tabular}
  \caption{\label{arcimagetranslation} Exemplar \textit{Machine Translation with Image} architectures. * - compares with ResNet-50 and VGG also.}
\end{table*}

\paragraph{Results.} In Table~\ref{mmtimagecomp2016} and Table~\ref{mmtimagecomp2018} we present the results obtained with a subset of models built using the Multi30k-MMT dataset presented earlier in Section~\ref{sssec:mmtimagedata}.

\begin{table}[!ht]
 \small
 \centering
\begin{tabular}{l|c|ccc}
\hline  %\toprule
\rowcolor{teal!35}
\multicolumn{5}{c}{Results of Different Methods} \\
\hline  %\midrule
\rowcolor{teal!35}
 Model         & Language              &  en $\rightarrow$ de     &  en $\rightarrow$ fr &  en $\rightarrow$ cs \\ 
\hline \addlinespace[0.3em]     %\midrule
                                                        &  BLEU & 36.5 & - & - \\
\multirow{-2}{*}{DoubleAtt~\shortcite{calixto:2017}}    &  METEOR & 55.0 & - & - \\ 
\midrule
                                                        &  BLEU & 37.3 & - & - \\
\multirow{-2}{*}{GVF~\shortcite{calixtoglob:2017}}      &  METEOR & 55.1 & - & - \\ 
\midrule
                                                            &  BLEU & 36.8 & - & - \\
\multirow{-2}{*}{Imagination~\shortcite{elliottimi:2017}}   &  METEOR & 55.8 & - & - \\ 
\midrule
                                                                    &  BLEU & 41.0 & 56.7 & - \\
\multirow{-2}{*}{Lium-cvc-ensemble~\shortcite{caglayanlium:2017}}   &  METEOR & 60.5 & 73.0 & - \\
\midrule
                                                                    &  BLEU & 37.6 & - & - \\
\multirow{-2}{*}{VMMT$_{\text{F}}$~\shortcite{calixto:2018}}        &  METEOR & 56.0 & - & - \\

\midrule
                                                        &  BLEU & 42.6 & 62.8 & 35.9 \\
\multirow{-2}{*}{CUNI-ensemble~\shortcite{helcl:2018}}  &  METEOR & 59.4 & 77.0 & 32.7 \\

\bottomrule
\end{tabular}
Translation\caption{\label{mmtimagecomp2016} Machine Translation with Image on the Multi30k test set [2016 (en $\rightarrow$ de), 2017 (en $\rightarrow$ fr), 2018 (en $\rightarrow$ cs)].}
\end{table}

\begin{table}[!ht]
 \small
 \centering
\begin{tabular}{l|c|ccc}
\hline  %\toprule
\rowcolor{teal!35}
\multicolumn{5}{c}{Results of Different Methods} \\
\hline  %\midrule
\rowcolor{teal!35}
 Model         & Language              &  en $\rightarrow$ de     &  en $\rightarrow$ fr &  en $\rightarrow$ cs \\
\hline \addlinespace[0.3em]     %\midrule
                                                        &  BLEU & 32.5 & 40.6 & 31.8 \\
\multirow{-2}{*}{CUNI-single~\shortcite{helcl:2018}}    &  METEOR & 52.3 & 61.0 & 30.6 \\
\midrule
                                                        &  BLEU & 38.5 & 44.1 & - \\
\multirow{-2}{*}{MeMAD~\shortcite{gronroos:2018}}       &  METEOR & 56.6 & 64.3 & - \\
\bottomrule
\end{tabular}
\caption{\label{mmtimagecomp2018} Machine Translation with Image on Multi30k test set [2018 (en $\rightarrow$ de, en $\rightarrow$ fr, en $\rightarrow$ cs)].}
\end{table}

\subsubsection{Machine Translation with Image - Discussion}
\label{sssec:mmtimageopenques}
This task is evaluated using only one dataset, e.g., Multi30k-MMT, containing descriptions in three source languages and one target language. Results presented in Table~\ref{mmtimagecomp2016} and Table~\ref{mmtimagecomp2018} refer to the shared task proposed in different years. We can observe that based on different years of test set release, varied sets of approaches outperform the baseline methods.  

\subsection{Machine Translation with Video}
\label{ssec:mmtvideotask}
In the following, we present more details about \textit{Multimodal Machine Translation} by using the video as the visual input.
\subsubsection{Machine Translation with Video - Introduction}
\label{sssec:mmtvideotaskintro}
The goal in video-guided machine translation~\shortcite{wangvatex:2019} is to translate a source language description into the target language equivalent using the video information as additional spatio-temporal context. 

\begin{figure}[!htb]
    \centering
        \includegraphics[width=\textwidth]{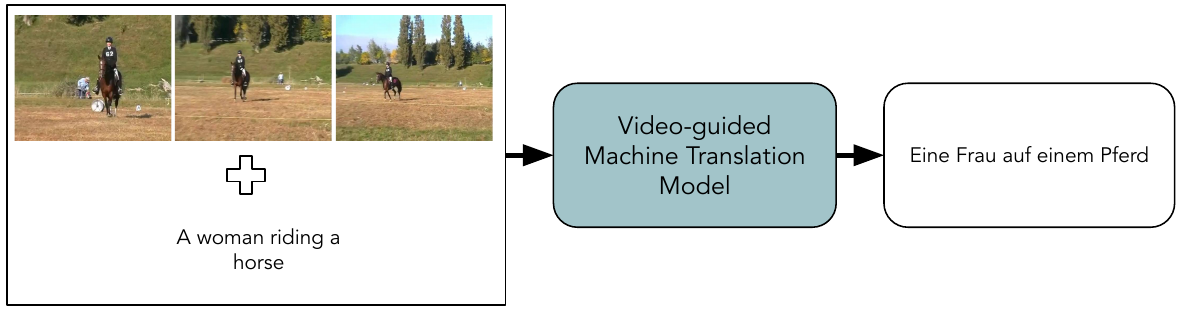}
    \caption{Schematic representation of Video-guided Machine Translation task.}\label{fig:mmtbase-vid}
\end{figure}

Figure~\ref{fig:mmtbase-vid} illustrates this task where the English language description accompanied by a video needs to be translated into the equivalent description in German.

\subsubsection{Machine Translation with Video - Datasets}
\label{sssec:mmtvideodata}
The VATEX\footnote{\url{http://vatex.org/main/index.html} \label{fnote: vatex-dataset-url}}~\shortcite{wangvatex:2019} dataset was created for English and Chinese languages to perform machine translation with video and also for the task of generating multilingual video descriptions. Table~\ref{table:vatex-dataset} presents more details about the dataset.

\begin{table}[!ht]
\small
    \centering
    \begin{tabular}{l | c c}
    \hline %\toprule
    \rowcolor{teal!35}
        Split           & Videos    & Action Label \\
    \hline \addlinespace[0.3em]    %\midrule
        Training        & 25,991    &  \cmark \\
        Validation      & 3,000     &  \cmark \\
        Public Test     & 6,000     &  - \\
        Secret Test     & 6,278     &   - \\
    \bottomrule
    \end{tabular}
    \caption{\label{table:vatex-dataset} Splits of the VATEX dataset. \textit{Secret Test} denotes human-annotated captions heldout for organizing challenges; Hence, this split is unavailable to the public.}
\end{table}

\subsubsection{Machine Translation with Video - Evaluation Measures, Models, and Results}
\label{sssec:mmtvideoall}
In this section, we review the measures used to evaluate different models of \textit{Machine Translation with Video} and the results obtained by them.

\paragraph{Evaluation Measures.} To evaluate the \textit{Machine Translation with Video} models, the Language metrics discussed in Section~\ref{sssec:idgall} are used.

\paragraph{Models.} Very few models have been created to investigate the task of \textit{Machine Translation with Video}. The temporal aspect of a video is crucial for providing effective translations. In contrast to \textit{Machine Translation with Image}, the task of \textit{Machine Translation with Video} only has models which are built using single textual source. In Table~\ref{arcvideotranslation}, we present some exemplar architectures (refer to \textit{Combined} column) which integrate both video and language inputs for addressing the task. We also include a column that showcases the optimization techniques used to train those models.

\begin{table*}[!ht]
\small
  \centering
  \begin{tabular}{lcccccc}
    \hline  %\toprule
    \rowcolor{teal!35}
    {Approach} & Video & Frame & Language & Combined & Optimizer & RL \\ 
    \hline \addlinespace[0.3em]     %\midrule
   ~\shortcite{wangvatex:2019} & I3D & - & LSTM &  NMT+LSTM VI & ADAM & \xmark \\ 
    \bottomrule
  \end{tabular}
  \caption{\label{arcvideotranslation} Exemplar \textit{Machine Translation with Video} architectures.}
\end{table*}

\paragraph{Results.} The models that have been created to address the task of \textit{Machine Translation with Video} is built using a single dataset, namely VATEX. In Table~\ref{resmmtvideo} we present results obtained with a subset of models built using the VATEX dataset presented earlier in Section~\ref{sssec:mmtvideodata}.

\begin{table}[!ht]
\small
  \centering
  \begin{tabular}{lccc}
    \hline  %\toprule
    \rowcolor{teal!35}
    {Model} & B-4 & METEOR  \\
    \hline \addlinespace[0.3em]     %\midrule
    NMT+LSTM VI~\shortcite{wangvatex:2019} [English $\rightarrow$ Chinese] & 30.20  & - \\
    NMT+LSTM VI~\shortcite{wangvatex:2019} [Chinese $\rightarrow$ English] & 27.18  & - \\
    \bottomrule
  \end{tabular}
  \caption{\label{resmmtvideo} Comparison of different methods on the VATEX dataset.}
\end{table}

\subsubsection{Machine Translation with Video - Discussion}
\label{sssec:mmtvideoopenques}
In Table~\ref{resmmtvideo}, we observe that only one method utilizing LSTM with video features from the pretrained I3D model (i.e., NMT+LSTM VI) is evaluated using the language metrics on the challenging VATEX dataset for both English and Chinese.

\section{Language-to-Vision Generation}
\label{sec:languagetovisiongen}
In this section, we explore the task of \textit{Language-to-Vision Generation}. The goal of this task is to generate visual content given their natural language descriptions. However, different variations of the task exist and will be discussed in the following.
\subsection{Language-to-Image Generation}
\label{ssec:ltoimagegen}
In the following, we describe the setting of \textit{Language-to-Image Generation} where an image is desired from a piece of natural language text (e.g., a sentence) describing the scene.
\subsubsection{Language-to-Image Generation - Introduction}
\label{sssec:ltoimagegenintro}
 A litany of different variations of the Language-to-Image Generation exists. For example, generation of an image can also be thought as a manipulation of an image. It allows for the generation of a new image using desired natural language description. We present some variations in the following.

\begin{figure}[!htb]
    \centering
        \includegraphics[width=\textwidth]{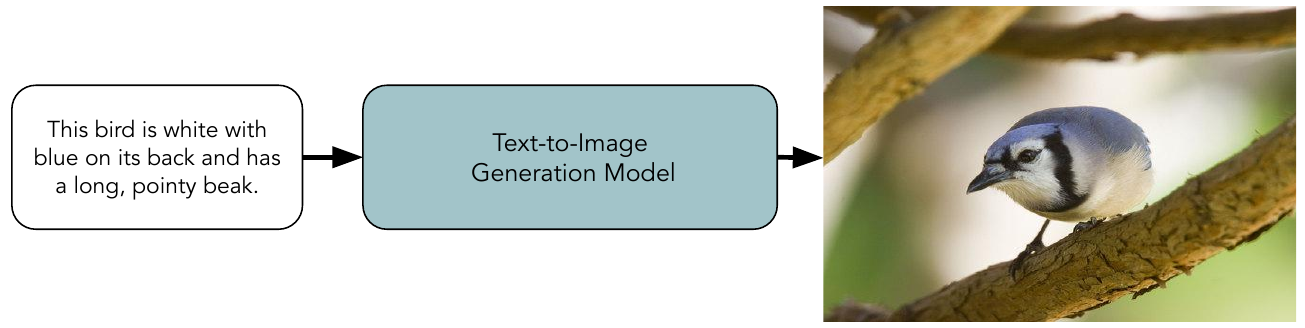}
    \caption{Given a \textit{natural language description}, a Language-to-Image Model generates an image by conditioning on the provided description.}\label{fig:texttoimage}
\end{figure}

\paragraph{Sentence-level Language-to-Image Generation.} The goal is to generate images conditioned on the natural language descriptions. It is considered as a fundamental problem in many applications. The success of Generative Adversarial Networks (GANs)~\shortcite{goodfellow:2014} has made possible the generation of interesting images of specific categories, such as room interiors, album covers, and faces~\shortcite{radford:2015}. This has led to an interest in bridging the gap between natural language text and image modeling. Figure~\ref{fig:texttoimage} shows the usage of natural language description for generating image with a Text-to-Image Generation Model. 

Initially, alignDRAW~\shortcite{mansimov:2015} was introduced to iteratively draw patches on a canvas, while attending to the relevant words in the description. Further, it was shown that visual concepts could be translated from characters to pixels~\shortcite{reed:2016} with a conditional GAN. This was further improved by taking instructions about what content should be drawn in which location in order to achieve high-quality image generation~\shortcite{reednips:2016}. Models which were developed to condition on classes for image generation~\shortcite{nguyen:2017} have also been used to generate images. However, the quality of images generated is much lower than when not conditioning on classes. Very close to this approach is Text-conditioned Auxiliary Classifier GAN (TAC-GAN)~\shortcite{dash:2017} which conditions images on both the sentence and class information, which has been shown to improve their structural coherence. To generate images with high resolution, several GANs were stacked together yielding stackGAN~\shortcite{zhang:2017,zhang:2018} that used a global sentence representation. This helped generate images of different sizes. To overcome the bottleneck of global-level sentence representation, attention-based GAN like AttGAN~\shortcite{xu:2018} was used to capture the fine-grained details at different sub-regions of the image. It pays attention to the relevant words in the natural language description. 

In other research efforts, a hierarchical approach~\shortcite{hong:2018} was taken by inferring the semantic layout of the image. Instead of learning a direct description to an image mapping, the generation process is decomposed into multiple steps. First a semantic layout from the text is constructed by the layout generator. Then, the layout is converted to an image by the image generator. Other kinds of approaches such as HDGAN~\shortcite{zhangp:2018} aim to accommodate hierarchical adversarial objectives inside the network to regularize mid-level representations and assist generator training in order to capture complex image information. This has been shown to generate images with high resolutions. 

Later, instead of dealing with natural-language descriptions, \shortciteA{johnson:2018} used image-specific scene graphs enabling explicitly reasoning about objects and their relationships. Further, for obtaining better high resolution images, coarse-resolution features were taken as input and Perceptual Pyramid Adversarial Network (PPAN) was introduced to directly synthesize multi-scale images conditioned on texts in an adversarial way~\shortcite{gao:2019}. Another approach named MirrorGAN~\shortcite{qiao:2019} targets the main goal of visual realism and semantic consistency for generating images from text. It proposes global-local attention and semantics-preserving framework where the image generated from the text is further used to generate the text back. This has been shown to semantically align with the given text and generated description.

In the following, we explore some of the related ideas which expand the scope of language-to-image generation.

\paragraph{Image Manipulation.} Image manipulation takes a different path from the earlier benchmark approaches about image generation, and so the TAGAN~\shortcite{nam:2018} was introduced to generate semantically manipulated images while preserving text-irrelevant contents. Here, the generator learns to generate images where only regions that correspond to the given text are modified. Another interesting approach is to have an interactive system that generates an image in an iterative manner. Recent approaches~\shortcite{zhuimage:2019} used attention in both the generator and the discriminator, while others~\shortcite{limanigan:2019} have designed error correction modules to rectify mismatched attributes and complete the missing contents in the generated image. There are also other variations where the source image is manipulated via natural language dialogue~\shortcite{cheng:2018}.

\paragraph{Fine-grained Image Generation.} Fine-grained image generation uses a recurrent image generation model~\shortcite{el:2018} to take into account both the generated output up to the current step as well as all past instructions for generation. This has been shown to add new objects, apply simple transformations to existing objects, and correct previous mistakes. Earlier research never concentrated on fine-grained generation of images, i.e., localizing objects. Recently, control of the location of individual objects within an image was made possible~\shortcite{hinz:2019} by adding a pathway in an iterative manner and applying them at different locations specified by the bounding boxes to both the generator and the discriminator. 

\paragraph{Sequential Image Generation.} The sequential image generation approach StoryGAN~\shortcite{listory:2018}, based on the sequential conditional GAN, concentrates on story by generating a sequence of images, when given a multi-sentence paragraph. Termed as story visualization, it behaves exactly opposite to image storytelling and has been shown to generate images with high quality, while also achieving contextual consistency.

\subsubsection{Language-to-Image Generation - Datasets}
\label{sssec:imagegendata}
For image generation, existing image datasets have been modified to accommodate image descriptions. Initially, the Oxford-102\footnote{\url{http://www.robots.ox.ac.uk/~vgg/data/flowers/102} \label{fnote: oxford-102-dataset-url}} and Caltech-UCSD Birds (CUB)\footnote{\url{http://www.vision.caltech.edu/visipedia/CUB-200-2011.html} \label{fnote: cub-200-dataset-url}} datasets consisting of flower and bird images belonging to 102 and 200 classes respectively are expanded with image descriptions~\shortcite{reed:2016}.  Table~\ref{table:oxford-102-dataset} and Table~\ref{table:cub-200-2011-dataset} presents splits of the datasets.
\begin{table}[!ht]
\small
    \centering
    \begin{tabular}{l | c c c c}
    \hline  %\toprule
    \rowcolor{teal!35}
         Split          & Images    & Captions per Image    & Total Captions \\
    \hline \addlinespace[0.3em]    %\midrule
        Training        &5,878      & 10                    & 58,780 \\
        Validation      &1,156      & 10                    & 11,560 \\
        Test            & 1,155     & 10                    & 11,550 \\
    \hline \addlinespace[0.3em]    %\midrule
        Total           & 8,189     &10                     &81,890 \\
    \bottomrule
    \end{tabular}
    \caption{\label{table:oxford-102-dataset} Splits of the Oxford-102 dataset with image descriptions.}
\end{table}

\begin{table}[!ht]
\small
    \centering
    \begin{tabular}{l | c c c}
    \hline  %\toprule
    \rowcolor{teal!35}
        Split           & Images    & Captions per Image    & Total Captions \\
    \hline \addlinespace[0.3em]    %\midrule
        Training        & 8,855     & 10                    & 88,550 \\
        Validation      & - & - & - \\
        Test            & 2,933     & 10                    & 29,330\\
    \hline \addlinespace[0.3em]    %\midrule
        Total           & 11,788    & 10                    &117,880\\
    \bottomrule
    \end{tabular}
    \caption{ \label{table:cub-200-2011-dataset} Splits of the CUB dataset with image descriptions.}
\end{table}

Similarly, the MSCOCO dataset (see Section~\ref{para:mscoco-data}) is also used for the reversed task of description generation, i.e., given a description, generate the image matching the description. We represent this dataset as MSCOCO-Gen. Table~\ref{table:mscocogeneration-dataset} presents the splits of the dataset.
\begin{table}[!ht]
\small
    \centering
    \begin{tabular}{l | c c c}
    \hline %\toprule
    \rowcolor{teal!35}
        Split           &Images         &  Captions per Image   & Total Captions \\
    \hline \addlinespace[0.3em]    %\midrule
        Training        & 82,783        & 5                     & 413,915 \\
        Validation & - & - & - \\
        Test            & 40,504        & 5                     & 202,520 \\
    \hline \addlinespace[0.3em]    %\midrule
        Total           & 123,287       & 5                     & 616,435 \\
    \bottomrule
    \end{tabular}
    \caption{\label{table:mscocogeneration-dataset} Splits of the MSCOCO-Gen dataset. }
\end{table}

\subsubsection{Language-to-Image Generation - Evaluation Measures, Models, and Results}
\label{sssec:ltoiall}
In this section, we review the measures used to evaluate different models of \textit{Language-to-Image Generation} and the results obtained by them.

\paragraph{Evaluation Measures.} There are different evaluation measures which are explicitly used for evaluating Language-to-Image generation models and are discussed below in detail.

\begin{itemize}
    \item \textbf{Inception Score (IS)}~\shortcite{salimans:2016} was initially proposed to compare the quality of images generated by GAN models. A pretrained Inception-v3 model~\shortcite{szegedy:2016} is applied to the generated image to get the conditional label distribution with low entropy. A similar idea is applied for the generated images on the given text descriptions for automatic evaluation. Higher scores are better for IS.
    \item \textbf{Fr{\'e}chet Inception Distance (FID)}~\shortcite{heusel:2017} is supposed to improve on IS by comparing the statistics of generated samples to original samples, instead of evaluating generated samples in an isolated manner. It also depends on the Inception-v3 model. In particular, the pool3 layer of the Inception-v3 is used for generating original samples for comparison. Lower FID is better as it corresponds to more similar generated and original samples.
    \item \textbf{R-precision} is inspired from the ranking retrieval results. It is used as a complementary evaluation metric for the language-to-image generation. Specifically, generated images are used to query their corresponding natural language descriptions to find how many relevant descriptions are retrieved.
\end{itemize}

\paragraph{Models.} Many models have been created to approach the task of \textit{Language-to-Image Generation}. In Table~\ref{arclantoimg}, we present some exemplar architectures (refer to \textit{Combined} column) that integrate both image and language for addressing the task. We also include a column that showcases the optimization techniques used to train those models.

\begin{table*}[!ht]
\small
  \centering
  \begin{tabular}{lccccc}
    \hline  %\toprule
    \rowcolor{teal!35}
    Approach & Image & Language & Combined & Optimizer & RL\\
    \hline \addlinespace[0.3em]     %\midrule
~\shortcite{reed:2016}       & - & char-CNN-RNN & GAN-INT-CLS & ADAM & \xmark \\
~\shortcite{reednips:2016}         & - &  char-CNN-GRU & GAWWN & ADAM & \xmark \\
~\shortcite{zhang:2017}         & - & - & StackGAN & ADAM & \xmark \\
~\shortcite{xu:2018}              & Inception-v3 & BiLSTM & AttGAN & - & \xmark\\
~\shortcite{qiao:2019}         & - & BiLSTM & MirrorGAN & - & \xmark \\
\bottomrule
  \end{tabular}
  \caption{\label{arclantoimg} Exemplar \textit{Language-to-Image Generation} architectures.}
\end{table*}

\paragraph{Results.} In Table~\ref{resltoicub}, Table~\ref{resltoioxford}, and Table~\ref{resltoicoco} we present results obtained with a subset of models built using the CUB, Oxford-102, and COCO datasets presented earlier in Section~\ref{sssec:imagegendata}.

\begin{table*}[!ht]
 \small
 \centering
\begin{tabular}{l|c|ccc}
\hline  %\toprule
\rowcolor{teal!35}
\hline  %\midrule
\rowcolor{teal!35}
 Model         & Resolution              &   IS       &  FID     &   HR    \\
\hline \addlinespace[0.3em]     %\midrule
                                       
GAN-INT-CLS~\shortcite{reed:2016}        &  64x64 & 2.88 $\pm$ .04 & 68.79 & 2.76 $\pm$ .01  \\
\midrule
                                                    &  64x64 & 3.10 $\pm$ .03 & 53.51 & - \\
\multirow{-2}{*}{GAWWN~\shortcite{reednips:2016}}   &  128x128 & 3.62 $\pm$ .07 & 72.65 & 1.95 $\pm$ .02\\
                                
\midrule
                                                    &  64x64 & 3.02 $\pm$ .03 & 35.11 & - \\
\multirow{-2}{*}{StackGAN~\shortcite{zhang:2017}}   &  256x256 & 3.70 $\pm$ .04 & 51.89 & 1.29 $\pm$ .02 \\
\midrule
                                        
StackGAN++~\shortcite{zhang:2018}        &  256x256 & 4.04 $\pm$ .05 & 15.30 & 1.19 $\pm$ .02 \\
\midrule
 %                                       &  64x64 & - & - & - \\
%PPGN~\shortcite{nguyen:2017}            &  128x128 & - & - & - \\
 %                                       &  256x256 & - & - & - \\
AttGAN~\shortcite{xu:2018}              &  256x256 & 4.36 $\pm$ .03 & - & - \\
\midrule
                                        
MirrorGAN~\shortcite{qiao:2019}         &  256x256 & 4.56 $\pm$ .05 & - & - \\
\bottomrule
\end{tabular}
\caption{\label{resltoicub} Comparison of different methods using generated images of different resolutions on the ``CUB'' dataset. R-precision (\%) for 256x256 with AttGAN (53.31) and MirrorGAN (57.67). HR - Human Ranking.}
\end{table*}

\begin{table*}[!ht]
 \small
 \centering
\begin{tabular}{l|c|ccc}
\hline      %\toprule
\rowcolor{teal!35}
\hline  %\midrule
\rowcolor{teal!35}
 Model         & Resolution              &   IS       &  FID     &   HR      \\ 
\hline \addlinespace[0.3em]     %\midrule
                                        
GAN-INT-CLS~\shortcite{reed:2016}      &  64x64 & 2.66 $\pm$ .03 & 79.55 & 1.84 $\pm$ .02 \\

\midrule                                
                                                    &  64x64 & 2.73 $\pm$ .03 & 43.02 & - \\
\multirow{-2}{*}{StackGAN~\shortcite{zhang:2017}}   &  256x256 & 3.20 $\pm$ .01 & 55.28 & 1.16 $\pm$ .02 \\
\midrule
                                        
StackGAN++~\shortcite{zhang:2018}      &  256x256 & 3.26 $\pm$ .01 & 48.68 & 1.30 $\pm$ .03  \\
\bottomrule
\end{tabular}
\caption{\label{resltoioxford} Comparison of different methods using generated images of different resolutions on the ``Oxford-102'' dataset.}
\end{table*}

\begin{table*}[!ht]
 \small
 \centering
\begin{tabular}{l|c|ccc}
\hline  %\toprule
\rowcolor{teal!35}
\hline  %\midrule
\rowcolor{teal!35}
 Model         & Resolution              &   IS       &  FID     &   HR  \\ 
\hline \addlinespace[0.3em]     %\midrule
                                        
GAN-INT-CLS~\shortcite{reed:2016}      &  64x64 & 7.88 $\pm$ .07 & 60.62 & 1.82 $\pm$ .03 \\
                                       
\midrule                                
                                        &  64x64 & 8.35 $\pm$ .11 & 33.88 & - \\
\multirow{-2}{*}{StackGAN~\shortcite{zhang:2017}}   &  256x256 & 8.45 $\pm$ .03 & 74.05 & 1.18 $\pm$ .03 \\
\midrule
                                        
StackGAN++~\shortcite{zhang:2018}       &  256x256 & 8.30 $\pm$ .10 & 81.59 & 1.55 $\pm$ .05 \\
\midrule
                                        
PPGN~\shortcite{nguyen:2017}            &  256x256 & 9.58 $\pm$ .21 & - & - \\
\midrule
                                       
AttGAN~\shortcite{xu:2018}              &  256x256 & 25.89 $\pm$ .47 & - & - \\
\midrule
                                        
MirrorGAN~\shortcite{qiao:2019}         &  256x256 & 26.47 $\pm$ .41 & - & - \\
\bottomrule
\end{tabular}
\caption{\label{resltoicoco} Comparison of different methods using generated images of different resolutions on the ``COCO'' dataset. R-precision (\%) for 256x256 with AttGAN (72.13) and MirrorGAN (74.52).}
\end{table*}

\subsubsection{Language-to-Image Generation - Discussion}
\label{sssec:ltoiopenques}
The \textit{Language-to-Image Generation} task has been evaluated using three different datasets. The CUB and Oxford-102 datasets contain only one visual object per image, while COCO has multiple objects. Several methods based on modified GAN objectives have been proposed for the generation of an image for a given textual description. From Table~\ref{resltoicub}, Table~\ref{resltoioxford}, and Table~\ref{resltoicoco} we observe the recent MirrorGAN~\shortcite{qiao:2019} achieves best results for different image resolution types using task-specific measures on CUB and COCO. It is built on the idea of back-translation of the image to text. However, for Oxford-102, StackGAN++~\shortcite{zhang:2018} achieves the best result.
\subsection{Language-to-Video Generation}
\label{ssec:ltovideogen}
In the following, we discuss the setting of \textit{Language-to-Video Generation} where a video is desired as the visual output from a natural language text description of the scene in video.

\subsubsection{Language-to-Video Generation - Introduction}
\label{sssec:ltovideogenintro}
\begin{figure}[!htb]
    \centering
        \includegraphics[width=\textwidth]{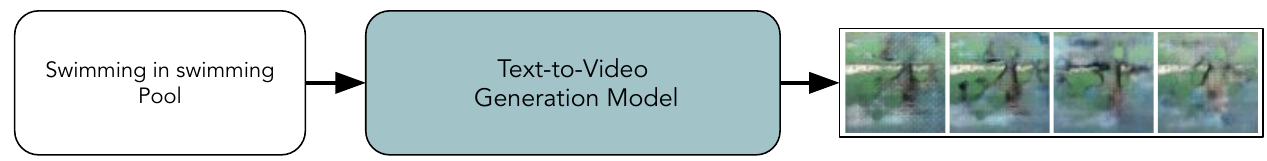}
    \caption{Given a \textit{natural language description}, a Language-to-Video model generates a video (represented as sequence of frames from~\shortciteA{livg:2018}) conditioned on the description.}\label{fig:texttovideo}
\end{figure}

The goal of Language-to-Video generation is to mimic language-to-image generation by considering the temporal aspect. However, language-to-video generation requires a stronger conditional generator than what is generally required for the language-to-image generation. This is because of the temporal nature of the videos. To address this challenge, a conditional generative model is trained~\shortcite{livg:2018} to extract both static and dynamic information from text which combines variational autoencoders (VAE)~\shortcite{kingma:2013} with GANs. Figure~\ref{fig:texttovideo} shows the usage of natural language description to generate a video with a text-to-video generation model. 

Another novel approach is to generate video from script. The composition, retrieval, and fusion network (Craft) model~\shortcite{gupta:2018} is capable of learning knowledge from the video-description data and applying it in generating videos from novel captions. It has been shown that the Craft model performs better than the direct pixel generation approaches and generalizes well to unseen captions and to video databases with no text annotations.

\subsubsection{Language-to-Video Generation - Datasets}
\label{sssec:ltovideogendata}
For video generation there are no publicly available datasets. However,~\shortciteA{livg:2018} have collected the Text2Video dataset belonging to ten different categories of YouTube videos, each ranging between 10-400 seconds for language-to-video generation. The categories of videos are biking in snow, playing hockey, jogging, playing soccer, playing football, kite surfing, playing golf, swimming, sailing and water skiing. For the purposes of model evaluation, the dataset is split into training, validation, and test sets in the ratio of 7:1:2 respectively, the details of which can be found in Table~\ref{table:text2video-dataset}.

\begin{table}[!ht]
\small
    \centering
    \begin{tabular}{l | c}
    \hline  %\toprule
    \rowcolor{teal!35}
         Split              & Videos \\
    \hline \addlinespace[0.3em]    %\midrule
        Training            & 2800 \\
        Validation          & 400 \\
        Test                & 800 \\
    \bottomrule
    \end{tabular}
    \caption{\label{table:text2video-dataset} Splits of Text2Video (Combines all categories).}
\end{table}

\subsubsection{Language-to-Video Generation - Evaluation Measures, Models, and Results}
\label{sssec:ltovideogenall}

In this section, we review the measures used to evaluate different models of \textit{Language-to-Video Generation} and the results obtained by them.

\paragraph{Evaluation Measures.} The \textit{Language-to-Video Generation} models are evaluated based on the Accuracy measure.

\paragraph{Models.} Only a limited set of models have been created so far to handle the task of \textit{Language-to-Video Generation}. In Table~\ref{arclantovideo}, we present an exemplar architecture (refer to \textit{Combined} column) which integrates video and language to address the task. We also include a column that showcases the optimization technique used to train the model.

\begin{table*}[!ht]
\small
  \centering
  \begin{tabular}{lcccccc}
    \hline      %\toprule
    \rowcolor{teal!35}
    {Approach} & Video & Frame & Language & Combined & Optimizer & RL \\ 
    \hline \addlinespace[0.3em]     %\midrule
    ~\shortcite{livg:2018} & MotionFeatures & - & LSTM & T2V & ADAM & \xmark \\
    \bottomrule
  \end{tabular}
  \caption{\label{arclantovideo} Exemplar \textit{Language-to-Video Generation} architectures.}
\end{table*}

\paragraph{Results.} In Table~\ref{resltovideogen} we present results obtained with a subset of models built using the ``TexttoVideo'' dataset presented earlier in Section~\ref{sssec:ltovideogendata}.

\begin{table}[!ht]
\small
  \centering
  \begin{tabular}{lc}
    \hline  %\toprule
    \rowcolor{teal!35}
    {Model} & Accuracy\\
    \hline \addlinespace[0.3em] %\midrule
    DT2V-baseline~\shortcite{livg:2018} & 0.101  \\
    PT2V~\shortcite{reed:2016} & 0.134  \\
    GT2V~\shortcite{livg:2018} & 0.192 \\
    T2V~\shortcite{livg:2018} & \textbf{0.426} \\
    \bottomrule
  \end{tabular}
  \caption{\label{resltovideogen} Comparison of accuracy (\%) scores of different models on Text2Video.}
\end{table}

\subsubsection{Language-to-Video Generation - Discussion}
\label{sssec:ltovideogenopenques}
The task of \textit{Language-to-Video Generation} is not as well-explored as the Language-to-Image generation task due to its complexity. Results presented in Table~\ref{resltovideogen} show that the approach proposed by~\shortciteA{livg:2018} achieves the best accuracy which is calculated using a simple video classifier which is a five-layer neural network model with 3D full convolutions and ReLU nonlinearities as activation functions.

\section{Vision-and-Language Navigation}
\label{sec:vlntask}
In this section, we explore the task of \textit{Vision-and-Language Navigation}. The goal of this task is to carry out navigation in an environment by interpreting natural language instructions.
\subsection{Image-and-Language Navigation}
\label{ssec:imageandlnav}
In the following, we provide a detailed description of the \textit{Image-and-Language Navigation} task in which photorealistic images forming 3D environments are used as visual inputs.

\subsubsection{Image-and-Language Navigation - Introduction}
\label{sssec:imageandlnavintro}
\begin{figure}[!htb]
    \centering
        \includegraphics[width=\textwidth]{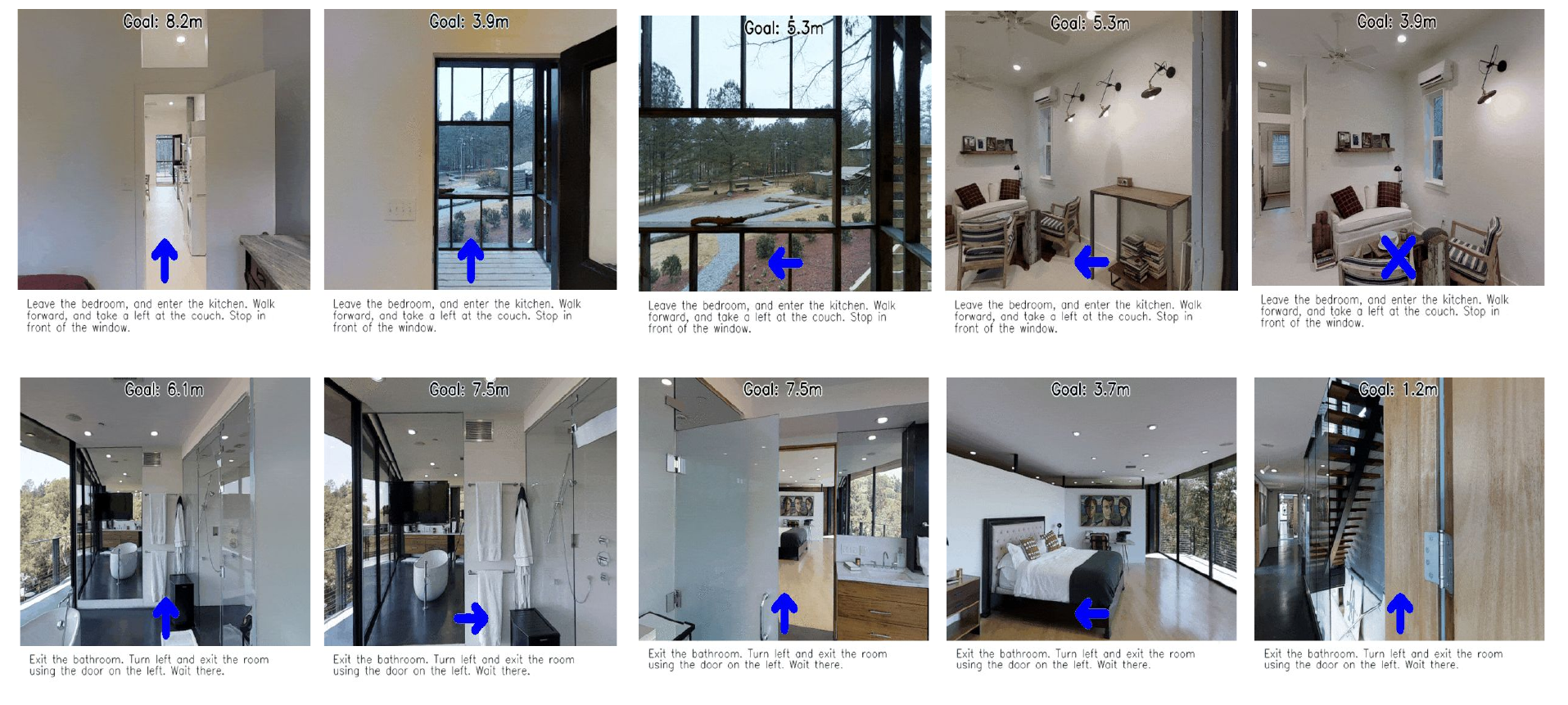}
    \caption{Given a \textit{sequence of images} and few instructions in \textit{textual} format (represented with a sequence of images from~\shortciteA{anderson:2018}), an Image-and-Language Navigation model is expected to carry out the navigation of an agent in an environment (indicated by blue arrows in the pictures).}\label{fig:vlnav}
\end{figure}

Most of the attempts at Vision-and-Language Navigation (VLN) use photorealistic images forming 3D environments. The goal of the Image-and-Language Navigation (ILN) task is to enable an autonomous agent (e.g., robot) to carry out navigation in an environment defined by the photo-realistic image views by means of interpreting natural language instructions~\shortcite{anderson:2018}. This requires the agent/robot to simultaneously process both vision and language inputs and navigate from a source to a target location. Figure~\ref{fig:vlnav} shows a visual depiction of the ILN task.

Initially, sequence-to-sequence models were proposed to address challenges in which the student-forcing approach achieved promising results in previously explored environments. One approach~\shortcite{wangll:2018} integrated a module to combine model-based and model-free reinforcement learning techniques to better generalize to unseen environments. There is also the reinforced cross-modal matching approach~\shortcite{wangrein:2018}, which enforces both local and global cross-modal grounding via reinforcement learning.

ILN can also be viewed as a search on a navigation graph~\shortcite{ma:2019} with a progress monitor as a learnable heuristic for search. It is improved by leveraging a visual-textual co-grounding attention mechanism to better align the instructions and visual scenes, and incorporates a progress monitor to estimate the agent's current progress towards the goal~\shortcite{maself:2019}. Another substantial improvement came from training an action space with an embedded speaker model~\shortcite{fried:2018}.  New instructions are synthesized for data augmentation and pragmatic reasoning was implemented for evaluating how well candidate action sequences explain an instruction. Improving over earlier approaches that make local action decisions or score entire trajectories using beam search, the novel approach of the FAST framework~\shortcite{ke:2019} balances local and global signals when exploring the environment allowing it to act greedily, but use global signals to backtrack when necessary. Also, \shortciteA{tan:2019} explore a generalizable navigational agent by training it in two stages. In the first stage, mixed imitation and reinforcement learning is combined, while in the second stage, fine-tuning is performed via newly-introduced ``unseen'' triplets.

ILN can also be perceived as a form of visual question answering (see Section~\ref{ssec:vqa}) that requires navigation to answer questions. Embodied Question Answering~\shortcite{das:2018,dascorl:2018} is explored with an agent that is spawned at a random location in a 3D environment and asked a question. For answering the question, the agent navigates through the 3D environment, for finding the information observed in the question. Other attempts used interactive question answering~\shortcite{gordon:2018} and grounded dialog~\shortcite{dewalk:2018}. Another set of approaches~\shortcite{misra:2018} aims to map instructions to actions in 3D Environments with visual goal prediction. Recently,~\shortciteA{chijust:2020} also made an interactive learning framework to endow the agent with the ability to ask for users' help in ambiguous situations.

\subsubsection{Image-and-Language Navigation - Datasets}
\label{sssec:imagendlnavdata}
For the image-and-language navigation task, three different datasets have been designed so far. In the following, we present the details of these datasets in separate paragraphs.
\paragraph{Room-2-Room (R2R).}\label{para:r2r-data}The R2R\footnote{\url{https://bringmeaspoon.org} \label{fnote: r2r-dataset-url}}~\shortcite{anderson:2018} dataset consists of real images of previously unseen building-scale 3D environments from Matterport3D~\shortcite{matterport3d-chang:2017}. The navigation instructions have been collected with the help of humans using AMT. Table~\ref{table:r2r-dataset} presents splits of the dataset.
\begin{table}[!ht]
\small
    \centering
    \begin{tabular}{l | c c }
    \hline  %\toprule
    \rowcolor{teal!35}
         Split              & Scenes    & Navigation Instructions  \\
    \hline \addlinespace[0.3em]    %\midrule
        Training            & 61        & 14,025        \\
        Validation (seen)   & 11        & 1,020         \\
        Validation (unseen) & 11        & 2,349          \\
        Test                & 18        & 4,173          \\
    \bottomrule
    \end{tabular}
    \caption{\label{table:r2r-dataset} Splits of the R2R dataset.}
\end{table}

\paragraph{ASKNAV.}\label{para:asknav-data}Similar to R2R, the ASKNAV\footnote{\url{https://github.com/debadeepta/vnla} \label{fnote: asknav-dataset-github-url}}~\shortcite{nguyen:2019} dataset is built on top of Matterport3D\footnote{\url{https://niessner.github.io/Matterport} \label{fnote: matterport-dataset-url}}. However, the objective differs in that the agent queries the advisor when in confusion and makes progress accordingly. It contains 10,800 panoramic views from 194,400 RGB-D images of 90 building-scale scenes. A data point in the dataset consists of a single starting viewpoint, but it has multiple goal viewpoints. Table~\ref{table:asknav-dataset-splits} presents the splits of dataset.

\begin{table}[!ht]
\small
    \centering
    \begin{tabular}{l | c c }
    \hline  %\toprule
    \rowcolor{teal!35}
         Split                  & Data points    & Goals \\
    \hline \addlinespace[0.3em]    %\midrule
        Training                & 94,798         & 139,757        \\
        Validation (seen)       & 4,874          &  7,768         \\
        Validation (unseen)     & 5,005          &  8,245         \\
        Test (seen)             & 4,917          &  7,470         \\
        Test (unseen)           & 5,001          &  7,537         \\
    \bottomrule
    \end{tabular}
    \caption{\label{table:asknav-dataset-splits} Splits of the ASKNAV dataset.}
\end{table}

\paragraph{TOUCHDOWN.}\label{para:touchdown-data}
Extending from building environments, the TOUCHDOWN\footnote{\url{https://github.com/lil-lab/touchdown} \label{fnote: touchdown-dataset-url}}~\shortcite{chentouchdown:2019} dataset is designed for addressing tasks such as executing navigation instructions (Navigation Only) and resolving spatial descriptions (SDR) in real-world environments. SDR is similar to the task of image referring expression (Section~\ref{sssec:iretask}).

The \textit{environment} includes 29,641 panoramas (360$^{\circ}$ Google Street View RGB images) and 61,319 edges from the New York City. Table~\ref{table:touchdown-dataset-stats} has more details about the dataset, while Table~\ref{table:touchdown-dataset-splits} presents its splits.

\begin{table}[!ht]
\small
    \centering
    \begin{tabular}{l | c c c}
    \hline  %\toprule
    \rowcolor{teal!35}
                                    & Dataset   & Vocab.  & Mean Text\\
    \rowcolor{teal!35}  
        \multirow{-2}{*}{Dataset}   & Size      & Size    & Length  \\                 
    \hline \addlinespace[0.3em]    %\midrule
        TOUCHDOWN (Complete task)                  & 9,326     & 5,625   & 108.0 \\
           Navigation Only                 & 9,326     & 4,999   & 89.6 \\
           SDR Only                         & 25,575    & 3,419   & 29.7 \\
    \bottomrule
    \end{tabular}
    \caption{\label{table:touchdown-dataset-stats} Statistics of the TOUCHDOWN dataset. \textit{Vocabulary Size} and \textit{Text Length} are computed by combining the training and validation sets.}
\end{table}

\begin{table}[!ht]
\small
    \centering
    \begin{tabular}{l | c | c }
    \hline  %\toprule
    \rowcolor{teal!35}  
        \multirow{-1}{*}{Task}          &     Split        & Examples  \\
    \hline \addlinespace[0.3em]    %\midrule
                                        & Training      & 6,526     \\
    \multirow{-2}{*}{Complete \&}              & Validation    & 1,391     \\
    \multirow{-2}{*}{Navigation Only}   & Test       & 1,409     \\
    \hline  \addlinespace[0.3em]    %\midrule
                                        & Training      & 17,880    \\
                SDR Only                & Validation    & 3,836     \\
                                        & Test       & 3,859     \\
    \bottomrule
    \end{tabular}
    \caption{\label{table:touchdown-dataset-splits} Splits of the TOUCHDOWN dataset.}
\end{table}

\paragraph{Cooperative Vision-and-Dialog Navigation (CVDN).}\label{para:cvdn-data}CVDN\footnote{\url{https://cvdn.dev} \label{fnote: cvdn-project-url}}~\shortcite{thomason:2019} is a dataset\footnote{\url{https://github.com/mmurray/cvdn/tree/master/tasks/CVDN/data}\label{fnote: cvdn-data-url}} of embodied, human-human dialogs situated in a simulated, photorealistic home environment. Table~\ref{table:cvdn-dataset} presents some statistics about the dataset.

\begin{table}[!htbp]
\small
    \centering
    \begin{tabular}{l c c }
    \hline  %\toprule
    \rowcolor{teal!35}
         Navigation Dialogs                 & Navigation        & Total Scenes  \\
    \rowcolor{teal!35}
    \multirow{1}{*}{(Human-Human)}          &Trajectories       & (MatterPort houses) \\
    \hline \addlinespace[0.3em]    %\midrule
        2,050                               & 7,000              & 83        \\
    \bottomrule
    \end{tabular}
    \caption{\label{table:cvdn-dataset} Statistics of the CVDN dataset.}
\end{table}

\paragraph{Action Learning From Realistic Environments and Directives (ALFRED).}\label{para:alfred-data}ALFRED\footnote{\url{https://askforalfred.com} \label{fnote: alfred-project-url}}~\shortcite{shridhar:2020} is a benchmark and interactive visual dataset for learning a mapping from natural language instructions and egocentric vision to sequences of actions for household tasks.

\begin{table}[!ht]
\small
    \centering
    \begin{tabular}{l | c c c}
    \hline  %\toprule
    \rowcolor{teal!35}
         Data                           &           &Number of          & Number of   \\
    \rowcolor{teal!35}
    \multirow{1}{*}{Split}  &\multirow{-2}{*}{Fold} &Scenes             &Annotations  \\
    \hline \addlinespace[0.3em]    %\midrule
        Training                        &-          & 108               & 21,023   \\\cline{1-4}
                                        &Seen       & 88                & 820   \\  %\cline{2-4}
        \multirow{-2}{*}{Validation}    &Unseen     & 4                 & 821   \\\cline{1-4}
                                        &Seen       & 107               & 1,533   \\  %\cline{2-4}
        \multirow{-2}{*}{Testing}       &Unseen     & 8                 & 1,529   \\    
    \bottomrule
    \end{tabular}
    \caption{\label{table:alfred-dataset-splits} Splits of the ALFRED dataset.}
\end{table}

\subsubsection{Image-and-Language Navigation - Evaluation Measures, Models, and Results}
\label{sssec:imageandlnavall}

In this section, we present the evaluation measures, models, and results achieved with various architectures of \textit{Image-and-Language Navigation}.

\paragraph{Evaluation Measures.} The measures that are designed explicitly for the \textit{Image-and-Language Navigation} system (e.g., R2R) are:

\begin{itemize}
    \item \textbf{Path Length (PL):} PL is a trajectory length where it is the total length of the executed path.
    \item \textbf{Navigation Error (NE):} NE is based on the shortest path distance in the navigation graph, and is calculated by measuring the average distance between the end-location predicted by the follower agent and the true route's end-location.
    \item \textbf{Success Rate (SR):} SR is the percentage of predicted end-locations within 3 meters of the true location.
    \item \textbf{Oracle Success Rate (OSR):} OSR measures the success rate at the closest point to the goal that the agent has visited along the trajectory.
    \item \textbf{Success Path Length (SPL):} SPL is a trade-off between SR and PL, by weighting SR by inverse PL.
\end{itemize}

\paragraph{Models.} Many models have been created to approach the task of \textit{Image-and-Language Navigation}. In Table~\ref{arcimandlannav}, we present some exemplar architectures (refer to \textit{Combined} column) which integrate both image and language to address the task. We also include a column that showcases the optimization techniques used to train those models. 

\begin{table*}[!ht]
\small
  \centering
  \begin{tabular}{lccccc}
    \hline  %\toprule
    \rowcolor{teal!35}
    Approach & Image & Language & Combined & Optimizer & RL\\
    \hline \addlinespace[0.3em]     %\midrule
    ~\shortcite{anderson:2018} & ResNet-152 & LSTM & Seq-to-Seq & ADAM & \xmark \\
    ~\shortcite{wangll:2018} &  ResNet-152 & LSTM & RPA & - & \cmark \\
   ~\shortcite{fried:2018} & ResNet-152 & LSTM &  Speaker-Follower & - & \cmark \\
   ~\shortcite{wangrein:2018} & ResNet-152 & LSTM & RCM  & ADAM & \cmark \\
    ~\shortcite{maself:2019}& ResNet-152 & LSTM & Self-Monitoring & ADAM & \xmark \\
    ~\shortcite{tan:2019} & ResNet-152 & LSTM & BackTranslation & RMSprop & \cmark \\
    ~\shortcite{ke:2019} & - & LSTM & FAST & - & \xmark \\
    \bottomrule
  \end{tabular}
  \caption{\label{arcimandlannav} Exemplar \textit{Image-and-Language Navigation} architectures.}
\end{table*}

\paragraph{Results.} As discussed earlier several models have been created to approach the task of \textit{Image-and-Language Navigation}. Furthermore, many datasets have been created to provide variety in the content so that they improve the generalization ability of the models. In this section, we cover the results obtained by the models from a representative dataset for this task. Table~\ref{resvlntest} -- \ref{resvlnunseenval} present results obtained with a subset of models built and evaluated using the R2R dataset which was introduced in Section~\ref{sssec:imagendlnavdata}.

\begin{table}[!ht]
\small
  \centering
  \begin{tabular}{lccccc}
    \hline      %\toprule
    \rowcolor{teal!35}
    {Model} & PL & NE & OSR & SR & SPL\\
    \hline \addlinespace[0.3em]     %\midrule
    Random  & 9.89 & 9.79 & 18.3 & 13.2 & 12 \\
    Seq-to-Seq~\shortcite{anderson:2018} & 8.13 & 7.85 & 26.6 & 20.4 & 18 \\
    RPA~\shortcite{wangll:2018} & 9.15 & 7.53 & 32.5 & 25.3 & 23 \\
    Speaker-Follower~\shortcite{fried:2018} & 14.82 & 6.62 & 44.0 & 35.0 & 28 \\
    Self-Monitoring~\shortcite{maself:2019} & 18.0 & - & - & 48.0 & 35 \\
    RCM~\shortcite{wangrein:2018} & 15.22 & 6.01 & 50.8 & 43.1 & 35 \\
    BackTranslation-Single~\shortcite{tan:2019} & 11.7 & - & - & \textbf{51.5} & \textbf{47} \\
    TacticalRewind-Greedy~\shortcite{ke:2019} & 22.08 & 5.14 & - & 54 & 41 \\
    \midrule
    BackTranslation-PreExplore~\shortcite{tan:2019} & 9.79 & - & - & 63.9 & 61 \\
    BackTranslation-Beam~\shortcite{tan:2019} & 687 & - & - & 68.9 & 1 \\
    FAST-Beam~\shortcite{ke:2019} & 196.53 & 4.29 & - & 61.0 & 3 \\
    \bottomrule
  \end{tabular}
  \caption{\label{resvlntest} Comparison of different methods on the R2R test set.}
\end{table}

\begin{table}[!ht]
\small
  \centering
  \begin{tabular}{lccccc}
    \hline  %\toprule
    \rowcolor{teal!35}
    {Model} & PL & NE & OSR & SR & SPL\\
    \hline \addlinespace[0.3em]     %\midrule
    Speaker-Follower~\shortcite{fried:2018} & - & 3.36 & 73.8 & 66.4 & - \\
    RCM+SIL~\shortcite{wangrein:2018} & 10.13 & 2.78 & 79.7 & 73.0 & -\\
    BackTranslation-Single~\shortcite{tan:2019} & 11.0 & 3.99 & - & 62.1 & 59 \\
    TacticalRewind-Greedy~\shortcite{ke:2019} & - & - & - & - & - \\
    \midrule
    BackTranslation-PreExplore~\shortcite{tan:2019} & 9.92 & 4.84 & - & 54.7 & 52 \\
    BackTranslation-Beam~\shortcite{tan:2019} & 703 & 2.52 & - & 75.7 & 1 \\
    FAST-Beam~\shortcite{ke:2019} & 188.6 & 3.13 & - & 70.0 & 4 \\
    \bottomrule
  \end{tabular}
  \caption{\label{resvlnseenval}  Comparison of different methods on the \textit{seen} validation set of R2R.}
\end{table}

\begin{table}[!ht]
\small
  \centering
  \begin{tabular}{lccccc}
    \hline      %\toprule
    \rowcolor{teal!35}
    {Model} & PL & NE & OSR & SR & SPL\\
    \hline \addlinespace[0.3em]     %\midrule
    Speaker-Follower~\shortcite{fried:2018} & - & 3.36 & 73.8 & 66.4 & - \\
    RCM+SIL~\shortcite{wangrein:2018} & 10.13 & 2.78 & 79.7 & 73.0 & -\\
    BackTranslation-Single~\shortcite{tan:2019} & 10.7 & 5.22 & - & 52.2 & 48 \\
    TacticalRewind-Greedy~\shortcite{ke:2019} & 21.17 & 4.97 & - & 56.0 & 43 \\
    \midrule
    BackTranslation-PreExplore~\shortcite{tan:2019} & 9.57 & 3.78 & - & 64.5 & 61 \\
    BackTranslation-Beam~\shortcite{tan:2019} & 663 & 3.08 & - & 69.0 & 1 \\
    FAST-Beam~\shortcite{ke:2019} & 224.42 & 4.03 & - & 63.0 & 2 \\
    \bottomrule
  \end{tabular}
  \caption{\label{resvlnunseenval}  Comparison of different methods on the \textit{unseen} validation set of R2R.}
\end{table}

\subsubsection{Image-and-Language Navigation - Discussion}
\label{sssec:imageandlnavopenques}
\textit{Image-and-Language Navigation} is evaluated with different splits of the R2R validation and test datasets. From Table~\ref{resvlntest}, Table~\ref{resvlnseenval}, and Table~\ref{resvlnunseenval} we can observe that Frontier Aware Search with backTracking (FAST)-beam~\shortcite{ke:2019} achieves the best result on the task-specific metrics. This approach balances local and global signals while exploring an unobserved environment. It also helps to act greedily but use global signals to backtrack whenever necessary. 

\section{Vision-and-Language Pretraining}
\label{sec:vlpretrain}

Inspired by the works of pretraining only on vision~\shortcite{he:2016} or solely on language data~\shortcite{devlin:2018,radford:2019,brownlanguage:2020}, the vision-and-language pretraining seeks to jointly learn representations using both visual and textual content for improving the efficiency of previously discussed vision and language integration tasks. Several methods will be discussed for vision-and-language pretraining, the architectures of which can be broadly divided into \textit{Single-stream} and \textit{Two-stream}. In the following, we provide more details on both types of architectures.

\paragraph{Single-stream Architectures.} These neural architectures are based on BERT-like~\shortcite{devlin:2018} models where they incorporate an Image Embedder, a Text Embedder, and a multi-layer Transformer~\shortcite{vaswani:2017}. The proposed models are pretrained on data which in general have parallel multimodal components i.e., videos or images along with captions. Further, the models are optimized with a combination of different objectives such as vision-based and text-based Masked Language Models (MLM), masked visual-feature modeling, and visual-linguistic matching. Learned representations are then used for different downstream tasks such as multimodal understanding or generation. For example, the VideoBERT~\shortcite{sunvideobert:2019} architecture has been designed to learn vision-language representations for a generative downstream task like video description generation (see Section~\ref{sssec:videocaptiongenintro}). While there are several other approaches such as Bounding Boxes in Text Transformer (B2T2)~\shortcite{albertifusion:2019}, Unicoder-VL~\shortcite{liunicoder:2019},  VL-BERT~\shortcite{suvlbert:2019}, UNITER~\shortcite{chenuniter:2020} are all designed for multimodal understanding and facilitate downstream tasks. Works such as VLP~\shortcite{zhouunified:2019}, OSCAR~\shortcite{lioscar:2020} and also its extension VinVL~\shortcite{vinvl-zhang:2021} have built unified models that can jointly understand and generate from cross-modal data. There is also an emergence of interest in probing vision-and-language pretrained models~\shortcite{caobehind:2020} to comprehend the contribution from each modality and also help in designing better model architectures and objectives.

\paragraph{Two-stream Architectures.} In contrast to the single-stream architectures, two-stream architectures adopted two independent encoders for learning visual and text representations. ViLBERT~\shortcite{luvilbert:2019} and LXMERT~\shortcite{tanlxmert:2019} are examples of two-stream architectures which used self-attention principles to jointly learn representations from visual and textual data. ViLBERT builds a co-attentional transformer layer, while LXMERT uses a cross-modality encoder. Similar to single-stream, the two-stream architectures also optimize their models with pretraining tasks, such as MLM and vision-text matching. Sometimes they use additional text-only corpora for achieving better generalization on long and complex sentences.
\newline
\newline
In Table~\ref{vlpretrainsupport}, we summarize both \textit{Single-stream} and \textit{Two-stream} architectures by enumerating the vision and language integration tasks they support. It has to be noted that these architectures only use \textit{subsets of the datasets} from each task. Also, the type of tasks they select are limited and are mostly discriminative. Broadly, we denote with (\cmark) or (\xmark) to indicate whether they support the task in question or not.

\begin{table*}[!ht]
\small
  \centering
  \begin{tabular}{lcccccccccc}
    \hline  %\toprule
    \rowcolor{teal!35}
    Approach & VDG & VS & VRE & VQA & VR & VE & VDiag & MMT & LVG & VLN\\
    \hline \addlinespace[0.3em]     %\midrule
     & &  &  & & Single-stream &  &  &  &  & \\
    \hline
    Unicoder-VL & \xmark &\xmark & \xmark & \xmark & \cmark & \xmark & \xmark & \xmark & \xmark & \xmark \\
    VL-BERT & \xmark & \xmark & \cmark & \cmark & \cmark & \xmark & \xmark & \xmark & \xmark & \xmark \\
    VideoBERT & \cmark & \xmark & \xmark & \xmark & \xmark & \xmark & \xmark & \xmark & \xmark & \xmark \\
    VLP & \cmark & \xmark & \xmark &\cmark & \xmark & \xmark & \xmark & \xmark &\xmark & \xmark \\
    OSCAR & \cmark & \xmark & \xmark &\cmark & \cmark & \xmark & \xmark & \xmark &\xmark & \xmark \\
    B2T2 & \xmark & \xmark &\xmark & \xmark & \cmark & \xmark & \xmark & \xmark & \xmark & \xmark \\
    UNITER & \xmark & \xmark & \cmark & \cmark & \cmark & \cmark & \xmark & \xmark & \xmark & \xmark \\
    VinVL & \cmark & \xmark & \xmark &\cmark & \cmark & \xmark & \xmark & \xmark &\xmark & \xmark \\
    \hline
     & &  &  & & Two-stream &  &  &  &  & \\
    \hline
    ViLBERT & \xmark & \xmark & \cmark & \cmark & \cmark & \xmark & \xmark &\xmark & \xmark & \xmark \\
    LXMERT & \xmark & \xmark & \xmark & \cmark & \cmark & \xmark & \xmark & \xmark & \xmark & \xmark \\
    \bottomrule
  \end{tabular}
  \caption{\label{vlpretrainsupport} Major Vision-and-Language Pretraining Architectures and their support for various Vision and Language Tasks. \textbf{VDG} - Visual Description Generation, \textbf{VS} - Visual Storytelling, \textbf{VRE} - Visual Referring Expression, \textbf{VQA} - Visual Question Answering, \textbf{VR} - Visual Reasoning, \textbf{VE} - Visual Entailment, \textbf{VDiag} - Visual Dialog, \textbf{MMT}- Multimodal Machine Translation, \textbf{LVG} - Language-to-Vision Generation, \textbf{VLN} - Vision-and-Language Navigation.}
\end{table*}

\section{Future Directions}
\label{sec:future}

The integration of vision and language research has come a long way since the pioneering works, particularly after the adoption of deep learning techniques. Although the performance of current state-of-the-art models still needs to catch up with human abilities, the gap is diminishing at a steady rate. However, there is still ample room for theoretical and algorithmic improvements. Here, we enumerate several possible future directions that have the potential to advance the research overall.

\paragraph{Learning Common Sense and World Knowledge.} There is a vast amount of out-of-domain data available which is unpaired with vision and language task-specific corpora. Leveraging such information as factual, hierarchical, or commonsense knowledge can significantly improve the intelligence of vision and language systems. Prior works have been shown to assist independent NLP tasks with pretrained language models such as commonsense reasoning~\shortcite{rajani:2019} and fact predictions~\shortcite{logan:2019}. It has also shown promise for image caption generation~\shortcite{wu:2017,mogadala:2018} and question answering~\shortcite{shah:2019,marino:2019}. Extending such ideas to other tasks would be an interesting research direction to pursue. Another possibility could be to utilize images, videos, and text in a synchronous and synergistic manner as they encode different aspects of the world and implicitly. Here, an open question would be how to extract world and common sense knowledge from these sources.

\paragraph{Addressing Large-scale Data Limitations.} Most approaches designed for tasks that integrate vision and language use large datasets for training. With this trend, it will soon become harder to design new tasks without having a dataset. To avoid such problems, future work will need to be adaptable to datasets of different sizes. Therefore, trade-off approaches are required where we know what amount of data is enough to master a certain task. This requires designing methods which might inspire from neuro-symbolic reasoning systems~\shortcite{yi:2018,vedantam:2019}.

\paragraph{Combining Multiple Tasks.} Some tasks are capable of sharing some ideas or representations of each other. For example, visual referring expression comprehension can be viewed as a visual dialog task~\shortcite{vries:2017} where a sequence of questions is used to refer to an object in the image. Similarly, image caption generation can be viewed as the visual referring expression generation task~\shortcite{maorefer:2016}.

\paragraph{Novel Neural Architectures for Representation.} Up until late 2017, the de facto standard for learning language and visual feature representations were RNNs and CNNs respectively. However, over the last few years, with the introduction of novel ideas that address the limitations of aforementioned neural network types, either theoretically or computationally, there is a growing interest to adopt these new techniques. For instance, the Transformer~\shortcite{vaswani:2017} architecture that is used extensively for pure NLP tasks may see adoption for the integration of vision and language tasks. It has already shown its applicability for image caption generation~\shortcite{sharma:2018}. In a similar manner, graph neural networks~\shortcite{scarselli:2008,kipf:2016,battaglia:2018} that were introduced to tackle graph-structured data, has already shown its promise in visual reasoning~\shortcite{haurilets:2019}. Exploiting the compositionality of visual objects to describe an entire visual scene with neural modular networks is also an interesting direction to explore for many vision and language tasks.

\paragraph{Image vs Video.} Most of the research into integrating vision and language concentrates on static images. This trend is clearly evident from the array of datasets and methods available for image and language integration tasks. Nevertheless, although a complex task, similar attention needs to be embraced for videos for which there is a scarcity of datasets. For instance, there is only one dataset available for tasks such as Video Dialog (Section~\ref{ssec:videodialog}), Video Referring Expression (Section~\ref{ssec:videoretask}), Language-to-Video Generation (Section~\ref{ssec:ltovideogen}), and Machine Translation with Videos (Section~\ref{ssec:mmtvideotask}), while tasks such as Vision-and-Language Navigation (Section~\ref{sec:vlntask}) completely lack video-based datasets.

\paragraph{3D-Vision and Language.} The world that we inhabit is inherently 3D. Thinking from this perspective, restricting vision and language research to just 2D, viz. images and videos, might be a hindrance for real world agents, e.g., humanoid robots, to fully understand the complexities of the 3D world and navigate with ease. To avoid such pitfalls, algorithms and techniques need to be developed for processing 3D inputs such as RGB-D, meshes, and point clouds in conjunction with language. Some pioneering works have already begun in this direction~\shortcite{referit3d:2020,scanrefer:2020,referit-rgbd-liu:2021,langrefer-roh:2021} and we anticipate the trend\footnote{\url{https://language3dscenes.github.io}\label{fnote:lang3d-workshop-url}} to shift more towards developing algorithms for understanding as well as the generation of 3D scenes~\shortcite{text2mesh3d-gan-briq:2021}, while utilizing language as a main or auxiliary modality.

\paragraph{Automatic Evaluation Measures.} Automatic evaluation measures exist for several vision and language tasks. However, most of them are adaptations from standalone NLP tasks such as machine translation. For example, BLEU and METEOR metrics used for evaluating visual caption generation and storytelling models have been found not to correlate well with human judgements~\shortcite{ber:2016}. The SPICE metric designed specifically for visual caption generation is dependent on parsing and is, therefore, not adaptable for other tasks such as storytelling. This kind of shortcoming shows us a promising research direction to pursue in developing evaluation measures applicable for several tasks. Recent attempts in developing BLEURT~\shortcite{bleurt-sellam:2020} and BERTScore~\shortcite{bertscore-zhang:2020} metrics show promising direction towards this goal.
Analogously, language-to-vision generation, although having quantitative measures, is typically dependent on human evaluation. It needs to adopt novel techniques for effective quantitative evaluation. Other tasks such as vision-and-language navigation and visual reasoning have specific measures for evaluation which can be improved further.

\section{Conclusion}
\label{sec:conc}

In undertaking this survey, we provided an overview as well as elaborate details on the recent trends in integration of vision and language research. In the beginning, we started with a background on various tasks in computer vision and NLP. Then, we identified ten distinct prominent tasks that aim to integrate visual and language modalities. To draw connections from traditional research tasks to V\&L integration tasks, we presented information about how each integration task is expanded from the standalone computer vision or NLP tasks on which they are originally based. Following that, we reviewed and analyzed each task separately by presenting a comprehensive introduction on how the tasks are designed in a bottom-up manner. Additionally, we presented different state-of-the-art methods used to address the tasks, along with exemplar architectures that are designed to integrate vision and language representations. We also provided a review on relevant datasets, evaluation measures, and the relative performance obtained by several state-of-the-art methods. Finally, in a separate section, we explored the various ways to pretrain generic models with large-scale multimodal data for supporting downstream vision and language integration tasks with minimal fine-tuning efforts. Moreover, we outlined how much the existing pretraining approaches support the ten prominent integration tasks that we described in earlier sections.

When comparing the standalone research done individually in the fields of computer vision and NLP, the synergy of both, fuelled by advanced machine learning techniques, are expected to yield more intelligent and sustainable systems. Making them easily accessible can, therefore, have direct commercial and societal impact. However, despite the significant progress achieved so far in many integration tasks, large-scale evaluation of those systems show that they still fall behind human performance, by a large margin. This fact confirms that there is still a good deal of room for improvement. In particular, designing novel evaluation measures and architectures that can adequately deal with the complexity of vision and language integration problems has the potential to address some of the challenges. Towards this goal, we outlined a few possible future research directions in the final section.

We believe that our efforts in publishing this survey will help to systematize future research papers and also investigate the unsolved problems that are hindering the progress of effective integration of vision and language modalities.

\acks{This work was supported by the German Research Foundation (DFG) as a part of - Project-ID 232722074 - SFB1102. We extend our special thanks to Matthew Kuhn and Stephanie Lund for painstakingly proofing the whole manuscript. We also acknowledge the insightful comments of Marius Mosbach on the first version of the manuscript.}

\newpage
%\subfile{appendix}
%\input{appendix}
\includepdf[pages=-]{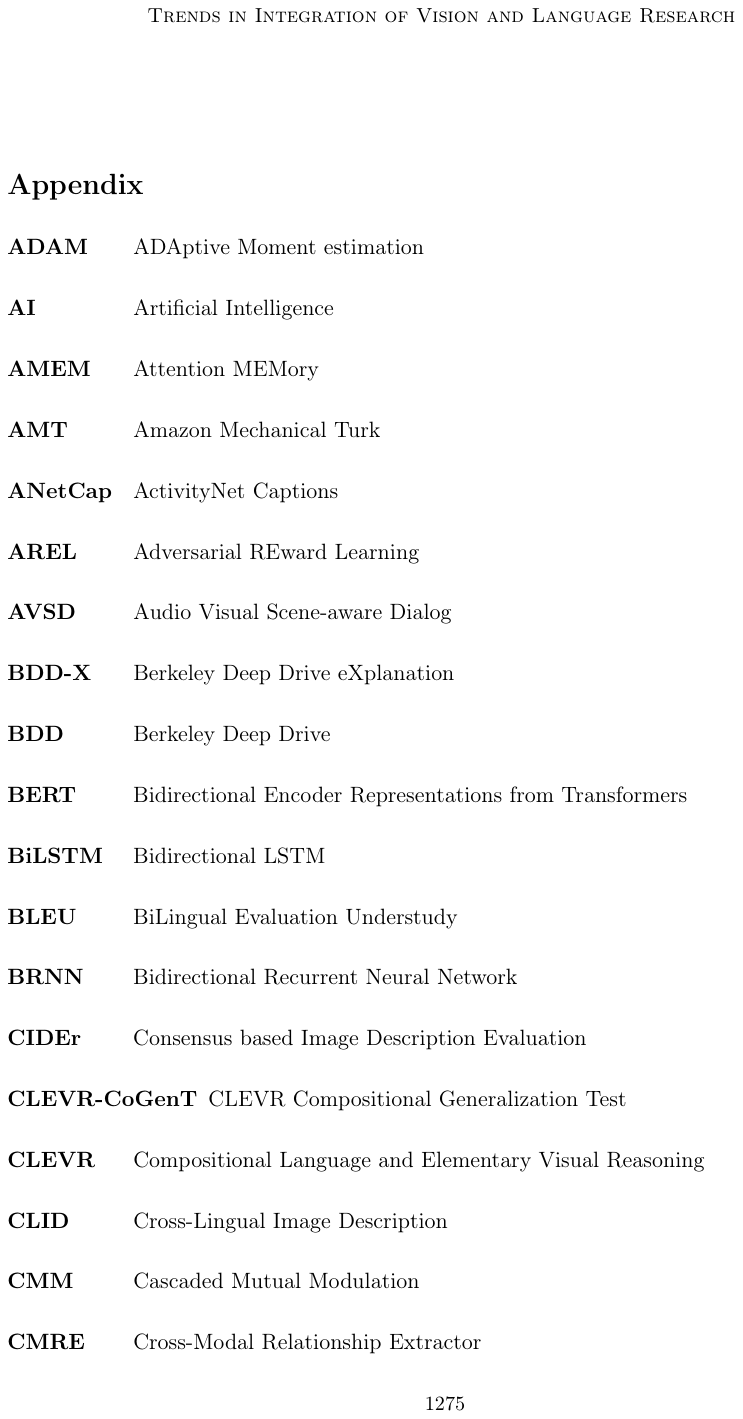}

\newpage % start references in a new page
\bibliography{references}
\bibliographystyle{theapa}

\end{document}